\documentclass[10pt,twocolumn,letterpaper]{article}

\usepackage{wacv} 

\usepackage{graphicx}
\usepackage{amsmath}
\usepackage{amssymb}
\usepackage{booktabs}

\usepackage[utf8]{inputenc} 
\usepackage[T1]{fontenc}    
\usepackage{url}            
\usepackage{booktabs}       
\usepackage{amsfonts}       
\usepackage{nicefrac}       
\usepackage{microtype}      
\usepackage{xcolor}         
\usepackage{bm}             

\usepackage{algorithm}
\usepackage{algpseudocode}

\usepackage{enumitem}       
\usepackage{multirow}       
\usepackage{colortbl}       
\usepackage[pagebackref,breaklinks,colorlinks]{hyperref}
\usepackage[accsupp]{axessibility} 

\usepackage[capitalize]{cleveref}
\crefname{section}{Sec.}{Secs.}
\Crefname{section}{Section}{Sections}
\Crefname{table}{Table}{Tables}
\crefname{table}{Tab.}{Tabs.}
\usepackage{amsmath}
\usepackage{amssymb}
\usepackage{mathtools}
\usepackage{amsthm}
\usepackage{soul, color, xcolor}

\theoremstyle{plain}
\newtheorem{theorem}{Theorem}[section]
\newtheorem{proposition}[theorem]{Proposition}

\theoremstyle{definition}

\theoremstyle{remark}

\def\wacvPaperID{2344} 
\def\confName{WACV}
\def\confYear{2025}

\begin{document}

\title{Elucidating the Solution Space of Extended Reverse-Time SDE for Diffusion Models}
\author{Qinpeng Cui$^{1}$\thanks{Equal contribution.}, Xinyi Zhang$^{1}$\footnotemark[1], Qiqi Bao$^{2}$, Qingmin Liao$^{1}$\thanks{Corresponding Author.}\\
$^1$Tsinghua University, China \\
$^2$Zhejiang University of Science and Technology, China \\
{\tt\small \{cqp22,xinyi-zh22,liaoqm\}@mails.tsinghua.edu.cn, nora919530829@163.com}
}
\maketitle

\begin{abstract}
Sampling from Diffusion Models can alternatively be seen as solving differential equations, where there is a challenge in balancing speed and image visual quality. ODE-based samplers offer rapid sampling time but reach a performance limit, whereas SDE-based samplers achieve superior quality, albeit with longer iterations. In this work, we formulate the sampling process as an Extended Reverse-Time SDE (ER SDE), unifying prior explorations into ODEs and SDEs. Theoretically, leveraging the semi-linear structure of ER SDE solutions, we offer exact solutions and approximate solutions for VP SDE and VE SDE, respectively. Based on the approximate solution space of the ER SDE, referred to as one-step prediction errors, we yield mathematical insights elucidating the rapid sampling capability of ODE solvers and the high-quality sampling ability of SDE solvers. Additionally, we unveil that VP SDE solvers stand on par with their VE SDE counterparts. Based on these findings, leveraging the dual advantages of ODE solvers and SDE solvers, we devise efficient high-quality samplers, namely ER-SDE-Solvers. Experimental results demonstrate that ER-SDE-Solvers achieve state-of-the-art performance across all stochastic samplers while maintaining efficiency of deterministic samplers. Specifically, on the ImageNet $128\times128$ dataset, ER-SDE-Solvers obtain 8.33 FID in only 20 function evaluations. Code is available at \href{https://github.com/QinpengCui/ER-SDE-Solver}{https://github.com/QinpengCui/ER-SDE-Solver}
\end{abstract}

\begin{figure}[ht]
\begin{center}
\centerline{\includegraphics[width=1.1\columnwidth]{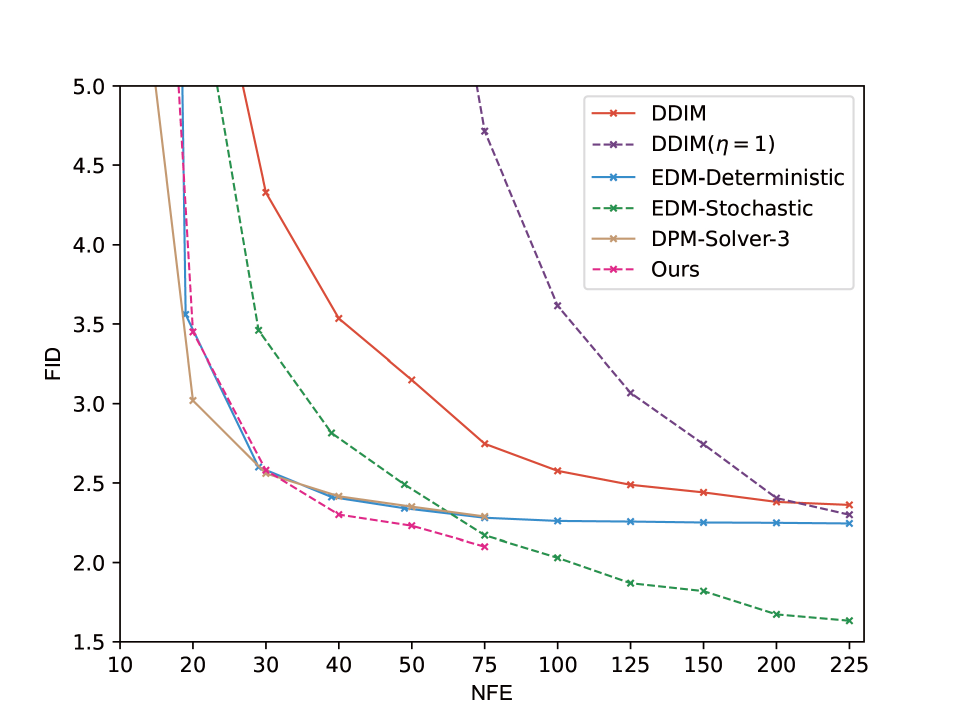}}
\caption{Sample quality (measured by FID$\downarrow$) on ImageNet $64\times64$ versus number of function evaluations (NFE) for deterministic samplers (DDIM \cite{song2021denoising}, EDM-Deterministic \cite{karras2022elucidating}, DPM-Solver-3 \cite{lu2022dpm}) and stochastic samplers (DDIM($\eta=1)$, EDM-Stochastic, Ours). Deterministic samplers excel in achieving rapid sampling but reach a mediocre quality with a large NFE, while stochastic samplers can further enhance image quality with an increase in NFE. Our efficient high-quality samplers demonstrate state-of-the-art performance among all stochastic samplers, simultaneously maintaining sampling efficiency comparable to deterministic samplers.}
\label{FID_NFE_sde_ode}
\end{center}
\end{figure}

\begin{figure}[t]
\centering
\includegraphics[width=0.48\textwidth]{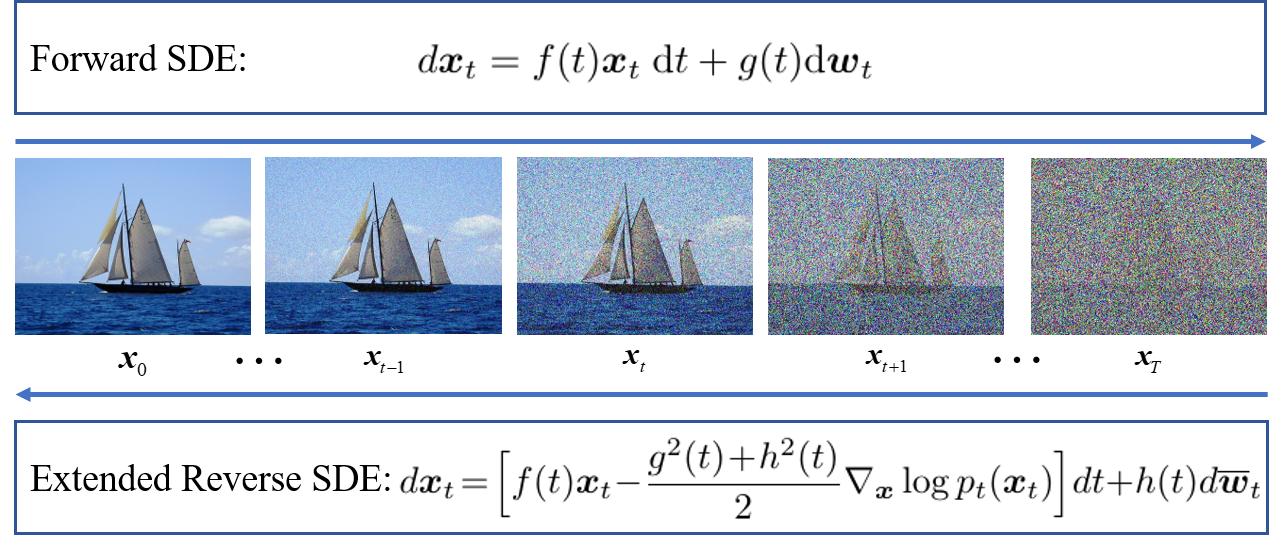}
\caption{A unified framework for DMs: The forward process described by an SDE transforms real data into noise, while the reverse process characterized by an ER SDE generates real data from noise. Once the score function $\nabla_{\mathbf{x}} \log p_t(\mathbf{x}_t)$ is estimated by a neural network, solving the ER SDE enables the generation of high-quality samples.}
\label{er_sde}
\end{figure}

\section{Introduction}
\label{introduction}
Diffusion Models (DMs) demonstrate an aptitude for producing high-quality samples on many tasks, such as image synthesis \cite{ho2020denoising,dhariwal2021diffusion, song2021score}, image super-resolution \cite{saharia2022image,gao2023implicit}, image restoration \cite{chung2022come,luo2023image}, image editing \cite{meng2021sdedit, avrahami2022blended}, image-to-image translation~\cite{zhao2022egsde,su2022dual}, and similar domains. These models define a forward diffusion process by gradually incorporating Gaussian noise to the real data and use iterative backward processes to remove this noise addition. In comparison to alternative generative models like Generative Adversarial Networks (GANs) \cite{goodfellow2014generative}, DMs can lead to higher quality samples for image generation but this often comes with the trade-off of increased time required for sampling. Such inefficiency constrains the applicability of DMs in real-time scenarios.

Prior fast samplers for DMs can be categorized into training-based and training-free methods. While training-based methods can generate high quality samples at 2$\sim$ 4 sampling steps \cite{salimans2021progressive,meng2023distillation,song2023consistency}, the prerequisite for retraining renders them cost-intensive. Conversely, training-free methods directly utilize raw information without retraining, offering broad applicability and high flexibility. \cite{song2021score} have indicated that the image generation process is equivalent to solving ordinary differential equations (ODEs) or stochastic differential equations (SDEs) in reverse time. The essence of training-free methods lies in designing efficient solvers for ODEs \cite{lu2022dpm,zhao2023unipc,zhang2023fast} or SDEs \cite{bao2022analytic,zhang2023gddim}. ODE-based samplers follow a deterministic sampling path, and SDE-based samplers incorporate randomness into the data state at each sampling step, leading to stochastic generation trajectories. Observations \cite{song2021denoising,karras2022elucidating,xue2023sa} (as shown in Fig.\ref{FID_NFE_sde_ode}) indicate that ODE-based samplers exhibit strong rapid sampling capabilities with fewer function evaluations. However, when aiming for even higher-quality images, increasing the number of function evaluations (NFE) leads to limited improvement in image quality. In contrast, SDE-based samplers show promise in producing data of superior quality with a large NFE, but at the cost of increased sampling time. We aim to design samplers that combine the advantages of both ODE-based samplers and SDE-based samplers in different NFE regimes, rendering it highly practical for real-world applications.

In this work, we present a unified framework for DMs, wherein Extended SDE formulation is proposed (see Fig.\ref{er_sde}). Within this framework, we define a solution space and design some highly efficient Extended Reverse-Time (ER)-SDE-Solvers to obtain high quality images. Specifically, we first model the sampling process as an ER SDE, which is an extension of~\cite{song2021score} and~\cite{zhang2023fast}. Inspired by~\cite{lu2022dpm}, we unveil the semi-linear structure inherent in the solutions. This structure consists of linear functions of data variables, nonlinear functions parameterized by neural networks and noise terms. Building on it, we deduce exact solutions for both Variance Exploding (VE) SDE and Variance Preserving (VP) SDE \cite{song2021score}. This is achieved by the analytical computation of the linear portions and noise terms, thereby circumventing associated prediction errors. Furthermore, we offer practical approximations for both VP SDE and VE SDE.

We refer to the errors of the approximate solutions from predicting every data state in the reverse process as one-step prediction errors. Our analysis reveals that varying levels of one-step prediction errors emerge due to the incorporation of different noise scales. This phenomenon gives rise to the solution space inherent in the ER SDE, and also provides a clearer insight into the different performance of ODE-based and SDE-based samplers in different NFE regimes. We ascertain that the minimal one-step prediction errors correspond to ODE solvers within the solution space, which mathematically demonstrates that the rapid sampling performance of ODE solvers when NFE is limited. Additionally, due to the noise injected during the reverse process can gradually corrects the accumulated prediction errors \cite{karras2022elucidating}, SDE solvers have the capability to generate higher-quality images as NFE increases. Moreover, given the consistency of the pretrained models, we theoretically establish that the VP SDE solvers yield image quality equivalent to VE SDE solvers. To take advantages of both ODE and SDE solvers for efficient high-quality sampling, we devise some specialized ER-SDE-Solvers through selecting the noise scale functions carefully.

In summary, we have made several theoretical and practical contributions: 1) We formulate an ER SDE and provide an exact solution as well as approximations for VP SDE and VE SDE, respectively. 2) Through a rigorous analysis of one-step prediction errors in the approximate solutions, we provide a mathematical exposition of the rapid sampling capability of ODE solvers and the high-quality sampling ability of SDE solvers. Moreover, we theoretically demonstrate that VP SDE solvers achieve the same level of image quality compared with VE SDE solvers. 3) By harnessing the dual advantages of ODE and SDE solvers, we present specialized ER-SDE-Solvers for efficient high-quality sampling. Extensive experimentation reveals that ER-SDE-Solvers achieve state-of-the-art performance across all stochastic samplers while maintaining efficiency of deterministic samplers. Additionally, the utilization of classifier guidance further enhances the efficiency of ER-SDE-Solvers.

\section{Diffusion Models}
\label{gen_inst}
Diffusion models (DMs) represent a category of probabilistic generative models encompassing both forward and backward processes. During the forward process, DMs gradually incorporate noise at different scales, while noise is gradually eliminated to yield real samples in the backward process. In the context of continuous time, the forward and backward processes can be described by SDEs or ODEs. In this section, we primarily review the stochastic differential equations (SDEs) and ordinary differential equations (ODEs) pertinent to DMs.
\subsection{Forward Diffusion SDEs}
\label{section2_1}
The forward process can be expressed as a linear SDE \cite{kingma2021variational}:
\begin{equation}\label{1}
d \boldsymbol{x}_{t}=f(t) \boldsymbol{x}_{t} \mathrm{~d} t+g(t) \mathrm{d} \boldsymbol{w}_{t}, \quad \boldsymbol{x}_{0} \sim p_{0}\left(\boldsymbol{x}_{0}\right),
\end{equation}
where $\boldsymbol{x}_{0} \in \mathbb{R}^{D}$ is a D-dimensional random variable following an unknown probability distribution \(p_0(\boldsymbol{x}_{0})\). $\left\{\boldsymbol{x}_{t}\right\}_{t \in[0, T]}$ denotes each state in the forward process, and $\boldsymbol{w}_{t}$ stands for a standard Wiener process. When the coefficients \(f(t)\) and \(g(t)\) are piecewise continuous, a unique solution exists \cite{oksendal2013stochastic}. By judiciously selecting these coefficients, Eq.(\ref{1}) can map the original data distribution to a priory known tractable distribution \(p_T(x_T)\), such as the Gaussian distribution.

The selection of \(f(t)\) and \(g(t)\) in Eq.(\ref{1}) is diverse. Based on the distinct noise employed in SMLD \cite{song2019generative} and DDPM \cite{sohl2015deep, ho2020denoising}, two distinct SDE formulations \cite{song2021score} are presented.

\textbf{Variance Exploding (VE) SDE:} The noise perturbations used in SMLD can be regarded as the discretization of the following SDE:
\begin{equation}\label{2}
d \boldsymbol{x}_{t}=\sqrt{\frac{\mathrm{d}\sigma^{2}_{t}}{\mathrm{d} t}} \mathrm{~d} \boldsymbol{w}_{t},
\end{equation}
where $\sigma_{t}$ is the positive noise scale. As $t \rightarrow \infty$, the variance of this stochastic process also tends to infinity, thus earning the appellation of Variance Exploding (VE) SDE.

\textbf{Variance Preserving (VP) SDE:} The noise perturbations used in DDPM can be considered as the discretization of the following SDE:
\begin{equation}\label{3}
d \boldsymbol{x}_{t}=\frac{d \log \alpha_t}{d t} \boldsymbol{x}_{t} \mathrm{d} t+\sqrt{\frac{d\sigma_t^2}{dt} - 2\frac{d\log \alpha_t}{d t}\sigma_t^2} \mathrm{d} \boldsymbol{w}_{t},
\end{equation}
where $\alpha_t$ is also the positive noise scale. Unlike the VE SDE, the variance of VP SDE remains bounded as $t \rightarrow \infty$. Therefore, it is referred to as Variance Preserving (VP) SDE.

\subsection{Reverse Diffusion SDEs}
\label{section2_2}
The backward process can similarly be described by a reverse-time SDE \cite{song2021score}:
\begin{equation}
\label{4}
d \boldsymbol{x}_{t}=\left[f(t) \boldsymbol{x}_{t}-g^{2}(t) \nabla_{\boldsymbol{x}} \log p_{t}\left(\boldsymbol{x}_{t}\right)\right] \mathrm{d} t+g(t) \mathrm{d} \overline{\boldsymbol{w}}_{t},
\end{equation}
where $\overline{\boldsymbol{w}}_{t}$ is the standard Wiener process in the reverse time. $p_t(\boldsymbol{x}_{t})$ represents the probability distribution of the state $\boldsymbol{x}_{t}$, and its logarithmic gradient $\nabla_{\boldsymbol{x}} \log p_{t}\left(\boldsymbol{x}_{t}\right)$ is referred to as the score function, which is often estimated by a neural network $\boldsymbol{s}_{\theta}(\boldsymbol{x}_{t}, t)$.

There are also studies \cite{zhang2021diffusion,zhang2023fast} that consider the following reverse-time SDE:
\begin{equation}
\begin{aligned}
\label{newsde}
d \boldsymbol{x}_{t}\!&=\!\!\left[f(t) \boldsymbol{x}_{t}\!- \! \frac{1\!+\!\lambda^2}{2}g^{2}(t) \nabla_{\boldsymbol{x}} \log p_{t}\left(\boldsymbol{x}_{t}\right)\right] \mathrm{d}t\!+\!\lambda g(t) \mathrm{d} \overline{\boldsymbol{w}}_{t},
\end{aligned}
\end{equation}
where the parameter \(\lambda \geq 0\). Eq.(\ref{newsde}) similarly shares the same marginal distribution as Eq.(\ref{1}).

Once the score-based network $\boldsymbol{s}_{\theta}(\boldsymbol{x}_{t}, t)$ is trained, generating images only requires solving the reverse-time SDE in Eq.(\ref{4}) or Eq.(\ref{newsde}). The conventional ancestral sampling method \cite{ho2020denoising} can be viewed as a first-order SDE solver \cite{song2021score}, yet it needs thousands of function evaluations for high-quality images. Numerous efforts \cite{jolicoeur2021gotta,bao2022analytic} have aimed to enhance sampling speed by devising highly accurate SDE solvers, but they still require hundreds of function evaluations, presenting a gap compared to ODE solvers.

\subsection{Reverse Diffusion ODEs}
In the backward process, in addition to directly solving the reverse-time SDE in Eq.(\ref{4}), a category of methods \cite{lu2022dpm, zhao2023unipc, zhang2023fast} focuses on solving the probability flow ODE corresponding to Eq.(\ref{4}), expressed specifically as
\begin{equation}
\begin{aligned}
\label{6}
d\boldsymbol{x}_{t}&=\left[f(t) \boldsymbol{x}_{t}-\frac{1}{2}g^{2}(t) \nabla_{\boldsymbol{x}} \log p_{t}\left(\boldsymbol{x}_{t}\right)\right] \mathrm{d} t.
\end{aligned}
\end{equation}
Eq.(\ref{6}) shares the same marginal distribution at each time \(t\) with the SDE in Eq.(\ref{4}), and the score function $\nabla_{\boldsymbol{x}} \log p_{t}\left(\boldsymbol{x}_{t}\right)$ can also be estimated by a neural network. Unlike SDEs introducing stochastic noise at each step, ODEs correspond to a deterministic sampling process. Despite several experiments ~\cite{song2021score} suggesting that ODE solvers outperform SDE solvers in terms of rapid sampling, SDE solvers can generate higher-quality images with an increase in NFE.

In summary, SDE-based methods can generate higher-quality samples, but exhibit slower convergence in high dimensions \cite{kloeden1992stochastic,lu2022dpm}. Conversely, ODE-based methods demonstrate the opposite behavior. To strike a balance between high-quality and efficiency in the generation process, in Sec.\ref{section3}, we model the backward process as an extended SDE, and provide analytical solutions as well as approximations for both VP SDE and VE SDE. Furthermore, we devise some ER-SDE-Solvers in Sec.\ref{section4}, achieving efficient high-quality sampling.

\vspace{3mm}
\section{Extended Reverse-Time SDE Solvers}
\label{section3}
There are three methods for recovering samples from noise in DMs. The first predicts the noise added in the forward process, achieved by a noise prediction model $\boldsymbol{\epsilon}_{\theta}\left(\boldsymbol{x}_{t}, t\right)$ \cite{ho2020denoising}. The second utilizes a score prediction model $\boldsymbol{s}_{\theta}(\boldsymbol{x}_{t}, t)$ to match the score function $\nabla_{\mathbf{x}} \log p_t(\mathbf{x}_t)$ \cite{hyvarinen2005estimation,song2021score}. The last directly restores the original data from the noisy samples, achieved by a data prediction model $\boldsymbol{x}_{\theta}\left(\boldsymbol{x}_{t}, t\right)$. These models can be mutually derived \cite{kingma2021variational}. Previously, most SDE solvers relied on the score-based model \cite{song2021score,jolicoeur2021gotta,bao2022analytic}. Based on modeling the backward process as an extended SDE in Sec.\ref{section3_1}, we proceed to solve the VE SDE and VP SDE for the data prediction model in Sec.\ref{section3_2} and Sec.\ref{section3_3}, respectively. Compared with the other two types of models, data prediction model can be synergistically combined with thresholding methods \cite{ho2020denoising,saharia2022photorealistic} to mitigate the adverse impact of large guiding scales, thereby finding broad application in guided image generation \cite{lu2022dpm-solver}.

\subsection{Extended Reverse-Time SDE}
\label{section3_1}
Besides Eq.(\ref{4}) and Eq.(\ref{newsde}), an infinite variety of diffusion processes can be employed. In this paper, we consider the following family of SDEs (referred to as Extended Reverse-Time SDE (ER SDE)):
\begin{equation}
\label{ersde}
d \boldsymbol{x}_{t}\!=\!\Big[f(t)\boldsymbol{x}_{t}-\frac{g^2(t)\!+\!h^2(t)}{2}\nabla_{\boldsymbol{x}} \log p_t(\boldsymbol{x}_{t})\Big]d t + h(t) d\overline{\boldsymbol{w}}_{t}.
\end{equation}
The score function $\nabla_{\boldsymbol{x}} \log p_t(\boldsymbol{x}_{t})$ can be estimated using the pretrained neural network. Hence, generating samples only requires solving Eq.(\ref{ersde}), guaranteed by Proposition \ref{prop0}.

\begin{proposition}[The validity of the ER SDE, proof in Supp.1.1]
\label{prop0}
When $\boldsymbol{s}_{\theta}(\boldsymbol{x}_{t}, t)=\nabla_{\boldsymbol{x}} \log p_{t}\left(\boldsymbol{x}_{t}\right)$ for all $\boldsymbol{x}_{t}$, $\overline p_{T}\left(\boldsymbol{x}_{T}\right)=p_{T}\left(\boldsymbol{x}_{T}\right)$, the marginal distribution $\overline p_t(\boldsymbol{x}_{t})$ of Eq.(\ref{ersde}) matches $p_t(\boldsymbol{x}_{t})$ of the forward diffusion Eq.(\ref{1}) for all $0 \leq t \leq T$.
\end{proposition}

Eq.(\ref{ersde}) extends the reverse-time SDE proposed in~\cite{song2021score,karras2022elucidating,xue2023sa,zhang2023fast}. Specifically, in~\cite{song2021score}, the noise scale \(g(t)\) added at each time step $t$ of the reverse process is the same as that of the corresponding moment in the forward process. ~\cite{karras2022elucidating,xue2023sa,zhang2023fast} introduce a non-negative parameter to control the extent of noise added during the reverse process. However, the form of the noise scale is relevant to \(g(t)\). In contrast, our ER SDE introduces a completely new noise scale \(h(t)\) for the reverse process. This implies that the noise scale \(h(t)\) added during the reverse process may not necessarily be correlated with the scale \(g(t)\) of the forward process. Particularly, the ER SDE reduces to the reverse-time SDE in Eq.(\ref{4}), Eq.(\ref{newsde}) and the ODE depicted in Eq.(\ref{6}) respectively  when the specific values of $h(t)$ are chosen.

By expanding the reverse-time SDE, we not only unify ODEs and SDEs under a single framework, facilitating the comparative analysis of these two methods, but also lay the groundwork for designing more efficient samplers. Further details are discussed in Sec.\ref{section4}.

\subsection{VE ER-SDE-Solvers}
\label{section3_2}
For the VE SDE, $f(t)\!=\!0$ and $g(t)\!=\!\sqrt{\frac{\mathrm{d}\sigma^{2}_{t}}{\mathrm{d}t}}$ \cite{kingma2021variational}. The relationship between the score prediction model $\boldsymbol{s}_{\theta}(\boldsymbol{x}_{t}, t)$ and the data prediction model $\boldsymbol{x}_{\theta}\left(\boldsymbol{x}_{t}, t\right)$ is $-[\boldsymbol{x}_{t}-\boldsymbol{x}_{\theta}\left(\boldsymbol{x}_{t}, t\right)]/{\sigma_{t}^{2}}\!=\!\boldsymbol{s}_{\theta}\left(\boldsymbol{x}_{t}, t\right)$. By replacing the score function with the data prediction model, Eq.(\ref{ersde}) can be expressed as
\begin{equation}
\label{12}
d \boldsymbol{x}_{t} \!=\! \frac{1}{2\sigma_t^2}\Big[\frac{d \sigma_t^2}{dt}\! + \! h^2(t)\Big][\boldsymbol{x}_{t}\! -\! \boldsymbol{x}_\theta(\boldsymbol{x}_{t}, t)] d t \!+ \!h(t) d\overline{\boldsymbol{w}}_{t}.
\end{equation}

Denote $d \boldsymbol{w}_{\sigma} := \sqrt{\frac{d \sigma_t}{d t}}d \overline{\boldsymbol{w}}_{t} $, $h^2(t) = \xi(t)\frac{d\sigma_t}{dt}$, we can rewrite Eq.(\ref{12}) w.r.t $\sigma$ as
\begin{equation}
\label{13}
	d \boldsymbol{x}_\sigma = \Big[\frac{1}{\sigma} + \frac{\xi(\sigma)}{2\sigma^2}\Big][\boldsymbol{x}_\sigma -\boldsymbol{x}_\theta(\boldsymbol{x}_\sigma, \sigma)]d \sigma + \sqrt{\xi(\sigma)}d \boldsymbol{w}_\sigma.
\end{equation}

We propose the exact solution for Eq.(\ref{13}) using \textit{variation-of-constants} formula \cite{lu2022dpm-solver}.
\begin{proposition}[Exact solution of the VE SDE, proof in Supp.1.2]
\label{prop1}
Given an initial value $\boldsymbol x_{s}$ at time $s > 0$, the solution $\boldsymbol x_{t}$ at time $t \in [0, s]$ of VE SDE in Eq.(\ref{13}) is:
\begin{equation}
\begin{aligned}
\label{15}
\boldsymbol x_t &=  \underbrace{\frac{\phi(\sigma_t)}{\phi(\sigma_s)}\boldsymbol x_s}_\text{(a) Linear term} + \underbrace{\phi(\sigma_t) \int_{\sigma_t}^{\sigma_s}\frac{\phi^{(1)}(\sigma)}{\phi^2(\sigma)}\boldsymbol x_\theta(\boldsymbol x_\sigma, \sigma)d \sigma}_\text{(b) Nonlinear term}\\
&+ \underbrace{ \sqrt{\sigma_t^2 - \sigma_s^2\Big[\frac{\phi(\sigma_t)}{\phi(\sigma_s)}\Big]^2 }\boldsymbol{z}_{s}}_\text{(c) Noise term},
\end{aligned}
\end{equation}
where $\boldsymbol{z}_{s} \sim \mathcal{N}(\mathbf{0}, \boldsymbol{I})$. $\phi(x)$ is derivable and $\int \frac{1}{\sigma} + \frac{\xi(\sigma)}{2\sigma^2}d \sigma = \ln \phi(\sigma)$.
\end{proposition}

Notably, the nonlinear term in Eq(\ref{15}) involves the integration of a non-analytical neural network $\boldsymbol x_\theta(\boldsymbol x_\sigma, \sigma)$, which can be challenging to compute. For practical applicability, Proposition \ref{prop3} furnishes high-stage solvers (followed by~\cite{gonzalez2023seeds}) for Eq.(\ref{15}).
\begin{proposition}[High-stage approximations of the VE SDE, proof in Supp.1.3]
\label{prop3} 
Given an initial value $\boldsymbol{x}_{T}$ and $M + 1$ time steps $\{{t_i}\}^M_{i=0}$ decreasing from $t_0 = T$ to $t_M = 0$. Starting with $ \tilde{\boldsymbol{x}}_{t_{0}}=\boldsymbol{x}_{T}$ , the sequence $\left\{\tilde{\boldsymbol{x}}_{t_{i}}\right\}_{i=1}^{M}$ is computed iteratively as follows:
\begin{equation}
	\begin{aligned}
\label{app_ve_sde}
		\tilde{\boldsymbol{x}}_{t_{i}} &= \frac{\phi(\sigma_{t_i})}{\phi(\sigma_{t_{i-1}})}\tilde{\boldsymbol{x}}_{t_{i-1}} + \Big{[} 1 - \frac{\phi(\sigma_{t_{i}})}{\phi(\sigma_{t_{i-1}})} \Big{]} \boldsymbol x_\theta(\tilde{\boldsymbol x}_{\sigma_{t_{i-1}}}, \sigma_{t_{i-1}})\\
		 &+ \sum_{n=1}^{k-1}\boldsymbol x_{\theta}^{(n)}(\tilde{\boldsymbol x}_{\sigma_{t_{i-1}}}, \sigma_{t_{i-1}})\Big[\frac{(\sigma_{t_i} - \sigma_{t_{i-1}})^n}{n!}+\!\phi(\sigma_{t_i})\\
&\!\!\!\!\!\!\!\int_{\sigma_{t_{i}}}^{\sigma_{t_{i-1}}}\!\!\frac{(\sigma \!-\! \sigma_{t_{i-1}})^{n-1}}{(n\!-\!1)!\phi(\sigma)} d \sigma\Big]\!\!+\!\!\sqrt{\sigma_{t_i}^2 \!\!-\! \sigma_{t_{i-1}}^2\Big[\frac{\phi(\sigma_{t_i})}{\phi(\sigma_{t_{i-1}})}\Big]^2 }\boldsymbol z_{t_{i\!-\!1}},
	\end{aligned}
\end{equation}
where $k \geq 1$. $\boldsymbol x_\theta^{(n)}(\boldsymbol x_\sigma, \sigma) := \frac{d^n \boldsymbol x_\theta(\boldsymbol x_\sigma, \sigma)}{d \sigma^n}$ is the $n$-th order total derivative of $\boldsymbol x_\theta(\boldsymbol x_\sigma, \sigma)$ w.r.t $\sigma$. 
\end{proposition}

$\int_{\sigma_{t_{i}}}^{\sigma_{t_{i-1}}} \frac{(\sigma - \sigma_{t_{i-1}})^{n\!-\!1}}{(n-1)!\phi(\sigma)} d \sigma$ in Eq.(\ref{app_ve_sde}) lacks an analytical expression, and we resort to \(N\)-point numerical integration for estimation. The detailed algorithms refer to Supp.2.

\subsection{VP ER-SDE-Solvers}
\label{section3_3}
For the VP SDE, $f(t) \!=\! \frac{d \log \alpha_t}{d t}$, $g(t)\! =\! \sqrt{\frac{d\sigma_t^2}{dt} - 2\frac{d\log \alpha_t}{d t}\sigma_t^2}$ \cite{kingma2021variational}. The relationship between the score prediction model $\boldsymbol{s}_{\theta}(\boldsymbol{x}_{t}, t)$ and the data prediction model $\boldsymbol{x}_{\theta}\left(\boldsymbol{x}_{t}, t\right)$ is $-[\boldsymbol x_t - \alpha_t \boldsymbol{x}_{\theta}\left(\boldsymbol{x}_{t}, t\right)]/{\sigma_t^2} = \boldsymbol{s}_{\theta}(\boldsymbol x_t , t)$. By replacing the score function with the data prediction model, Eq.(\ref{ersde}) can be written as:
\begin{equation}
\begin{aligned}
\label{vp12}
	d \boldsymbol x_t = & \Big{\{}\Big[\frac{1}{\sigma_t} \frac{d \sigma_t}{d t} + \frac{h^2(t)}{2 \sigma_t^2}\Big]\boldsymbol x_t\! - \!\Big[\frac{1}{\sigma_t}\frac{d\sigma_t}{dt}\! -\! \frac{1}{\alpha_t}\frac{d \alpha_t}{dt} + \frac{h^2(t)}{2\sigma_t^2}\Big]\\
&\alpha_t \boldsymbol x_{\theta}(\boldsymbol x_t, t)  \Big{\}}dt + h(t)d \overline{\boldsymbol{w}}_{t}.
\end{aligned}
\end{equation}

Let $h(t) = \eta(t)\alpha_t$, $\boldsymbol y_t = \frac{\boldsymbol x_t}{\alpha_t}$ and $\lambda_t = \frac{\sigma_t}{\alpha_t}$. Denote $d \boldsymbol w_\lambda := \sqrt{\frac{d\lambda_t}{dt}}d \overline{\boldsymbol w}_t$, $\eta^2(t)\!=\!\xi(t)\frac{d\lambda_t}{dt}$,and rewrite Eq.(\ref{vp12}) w.r.t $\lambda$ as
\begin{equation}
\label{vp13}
	d \boldsymbol y_\lambda\! = \!\Big[\frac{1}{\lambda}\! +\! \frac{\xi(\lambda)}{2\lambda^2}\Big][\boldsymbol y_\lambda\! - \! \boldsymbol x_\theta(\boldsymbol x_\lambda, \lambda)]d \lambda \!+\! \sqrt{\xi(\lambda)}d \boldsymbol w_\lambda. 
\end{equation}
Following \cite{lu2022dpm-solver}, we propose the exact solution for Eq.(\ref{vp13}) using the \textit{variation-of-constants} formula.

\begin{proposition}[Exact solution of the VP SDE, proof in Supp.1.4]
\label{prop4}
Given an initial value $\boldsymbol x_{s}$ at time $s > 0$, the solution $\boldsymbol x_{t}$ at time $t \in [0, s]$ of VP SDE in Eq.(\ref{vp13}) is:
\begin{equation}
\begin{aligned}
\label{vp14}
\boldsymbol x_t &=  \underbrace{\frac{\alpha_t}{\alpha_s} \frac{\phi(\lambda_t)}{\phi(\lambda_s)}\boldsymbol x_s}_\text{(a) Linear term} + \underbrace{\alpha_t\phi(\lambda_t) \int_{\lambda_t}^{\lambda_s}\frac{\phi^{(1)}(\lambda)}{\phi^2(\lambda)}\boldsymbol x_\theta(\boldsymbol x_\lambda, \lambda)d \lambda}_\text{(b) Nonlinear term} \\
& + \underbrace{\alpha_t \sqrt{\lambda_t^2 - \lambda_s^2\Big[\frac{\phi(\lambda_t)}{\phi(\lambda_s)}\Big]^2 }\boldsymbol{z}_{s}}_\text{(c) Noise term},
\end{aligned}
\end{equation}
where $\boldsymbol{z}_{s} \sim \mathcal{N}(\mathbf{0}, \! \boldsymbol{I})$. $\phi(x)$ is derivable and $\int \frac{1}{\lambda} + \frac{\xi(\lambda)}{2\sigma^2}d \lambda = \ln \phi(\lambda)$.
\end{proposition}

The solution of the VP SDE also involves integrating a non-analytical and nonlinear neural network. Proposition \ref{prop5} furnishes high-stage solvers (followed by~\cite{gonzalez2023seeds}) for Eq.(\ref{vp14}).

\begin{proposition}[High-stage approximations of the VP SDE, proof in Supp.1.5]
\label{prop5} 
Given an initial value $\boldsymbol{x}_{T}$ and $M + 1$ time steps $\{{t_i}\}^M_{i=0}$ decreasing from $t_0 = T$ to $t_M = 0$. Starting with $ \tilde{\boldsymbol{x}}_{t_{0}}=\boldsymbol{x}_{T}$ , the sequence $\left\{\tilde{\boldsymbol{x}}_{t_{i}}\right\}_{i=1}^{M}$ is computed iteratively as follows:
\begin{equation}
	\begin{aligned}
\label{app_vp_sde}
	\tilde{\boldsymbol{x}}_{t_{i}}\!\!&=\!\!\frac{\alpha_{t_i}}{\alpha_{t_{i-1}}} \frac{\phi(\lambda_{t_i})}{\phi(\lambda_{t_{i-1}})}\tilde{\boldsymbol{x}}_{t_{i-1}}\!\!+\!\alpha_{t_i}\!\Big{[}\!1 \!\!-\!\! \frac{\phi(\lambda_{t_{i}})}{\phi(\lambda_{t_{i-1}}\!)}\! \Big{]} \boldsymbol x_\theta(\tilde{\boldsymbol x}_{\lambda_{t_{i-1}}}\!,\! \lambda_{t_{i-1}}\!)\\
       &+\alpha_{t_i}\sum_{n=1}^{k-1}\boldsymbol x_{\theta}^{(n)}(\tilde{\boldsymbol x}_{\lambda_{t_{i-1}}}, \lambda_{t_{i-1}})\Big[\frac{(\lambda_{t_i} - \lambda_{t_{i-1}})^n}{n!}+\phi(\lambda_{t_i})\\
&\!\!\!\!\int_{\lambda_{t_{i}}}^{\lambda_{t_{i-1}}}\!\! \frac{(\lambda \!\!-\!\! \lambda_{t_{i-1}})^{n\!-\!1}}{(n\!\!-\!\!1)!\phi(\lambda)} d \lambda \Big]\!\!+\!\!\alpha_{t_i}\sqrt{\lambda_{t_i}^2\!\! -\!\! \lambda_{t_{i\!-\!1}}^2\!\Big[\frac{\phi(\lambda_{t_i})}{\phi(\lambda_{t_{i\!-\!1}})}\Big]^2 }\ \boldsymbol z_{t_{i\!-\!1}},
	\end{aligned}
\end{equation}
where $k \geq 1$. $\boldsymbol x_\theta^{(n)}(\boldsymbol x_\lambda, \lambda) := \frac{d^n \boldsymbol x_\theta(\boldsymbol x_\lambda, \lambda)}{d \lambda^n}$ is the $n$-th order total derivative of $\boldsymbol x_\theta(\boldsymbol x_\lambda, \lambda)$ w.r.t $\lambda$.
\end{proposition}

Similarly, we employ \(N\)-point numerical integration to estimate $\int_{\lambda_{t_{i}}}^{\lambda_{t_{i-1}}} \frac{(\lambda - \lambda_{t_{i-1}})^{n-1}}{(n-1)!\phi(\lambda)} d \lambda$ in Eq.(\ref{app_vp_sde}). The detailed algorithms are proposed in Supp.2.

\begin{figure*}[t]
    \begin{minipage}[t]{0.5\linewidth}
        \centering
        \includegraphics[width=1.0\textwidth]{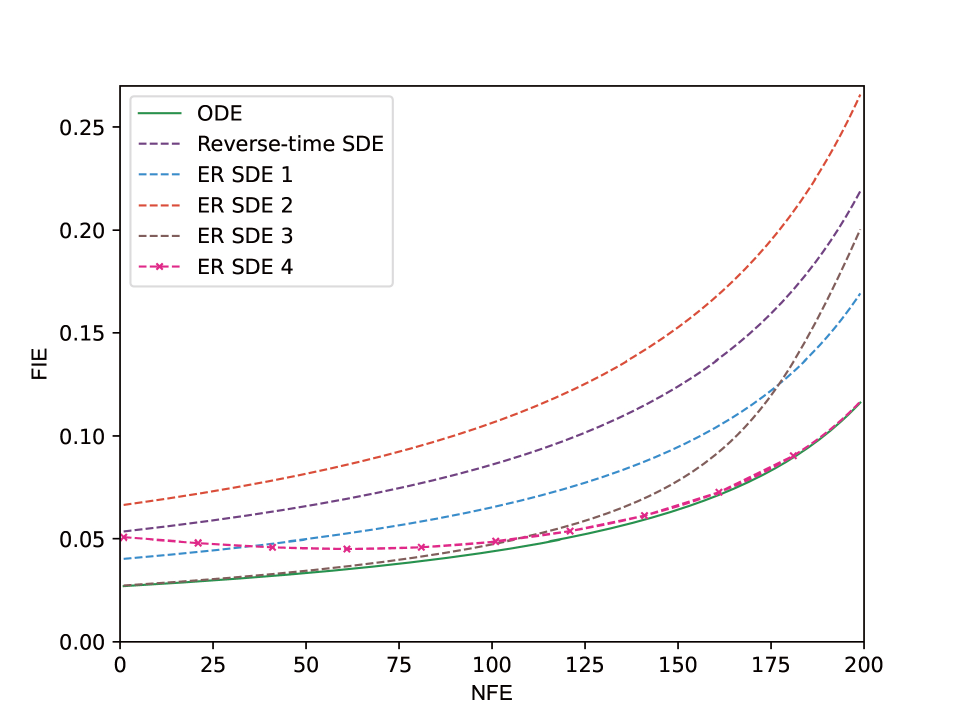}
        \centerline{(a) FIE coefficients}
    \end{minipage}%
    \begin{minipage}[t]{0.5\linewidth}
        \centering
        \includegraphics[width=1.0\textwidth]{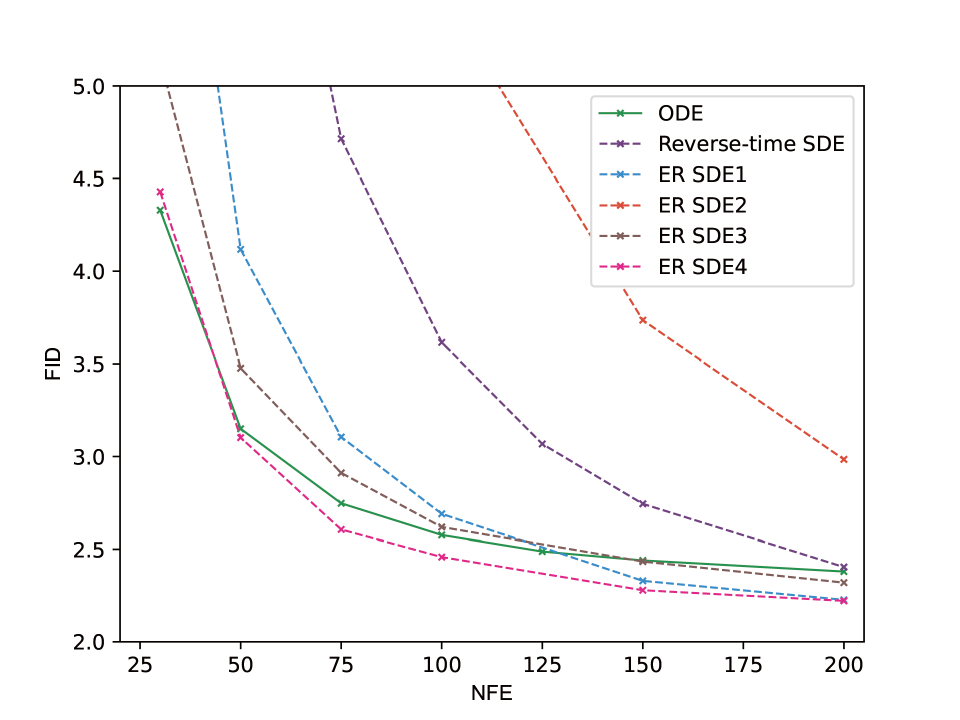}
        \centerline{(b) FID scores($\downarrow$) on ImageNet $64\times64$}
    \end{minipage}
    \caption{FIE coefficients (a) and FID scores (b) versus NFE for distinct noise scale functions. 1st-order solver is used here with the pretrained EDM. In the solution space of ER SDE, ODE solver shows minimal one-step prediction errors. ER SDE 4 demonstrates elevated error in the initial 100 NFE and gradually converges to the ODE's error profile. Thus, ER SDE 4 exhibit comparable efficiency to ODE solver but can further generate high-quality images. Image quality deteriorates for ill-suited noise scale functions (like ER SDE 2).}
\label{FEI_FID}
\end{figure*}

\vspace{3mm}
\section{Elucidating the Solution Space of ER SDE}
\label{section4}
This section primarily focuses on the solution space of ER SDE. Specifically, in Sec.{\ref{section4_1}}, we provide a mathematical explanation for experimental observations made in previous research. Furthermore, we introduce various specialized Extended Reverse-Time SDE Solvers (ER-SDE-Solvers) in Sec.{\ref{section4_2}}, which achieve efficient high-quality sampling.

\subsection{Insights about the Solution Space of ER SDE}
\label{section4_1}
Sec.{\ref{section3}} demonstrates that the exact solution of ER SDE comprises three components: a linear function of the data variables, a non-linear function parameterized by neural networks and a noise term. The linear and noise terms can be precisely computed, while errors arising from predicting the data state in the reverse process are present in the non-linear term. Due to the decreasing error as the stage increases (see Table\ref{table1}), the first-order error predominantly influences the overall error. Therefore, we exemplify the case with order $k=1$ for error analysis. Specifically, the first-order approximation for VE SDE is given by
\begin{equation}
\begin{aligned}
\label{app_ve_sde_1}
		\tilde{\boldsymbol{x}}_{t_{i}} &=\frac{\phi(\sigma_{t_i})}{\phi(\sigma_{t_{i\!-1}})}\tilde{\boldsymbol{x}}_{t_{i-1}}+ \Big{[\!} 1 - \frac{\phi(\sigma_{t_{i}})}{\phi(\sigma_{t_{i-1}})} \Big{]} \boldsymbol{x}_\theta(\tilde{\boldsymbol{x}}_{\sigma_{t_{i-1}}}, \sigma_{t_{i\!-1}})\\
 &+\sqrt{\sigma_{t_i}^2 - \sigma_{t_{i-1}}^2\Big[\frac{\phi(\sigma_{t_i})}{\phi(\sigma_{t_{i-1}})}\Big]^2 }\ \boldsymbol z_{t_{i-1}},
\end{aligned}
\end{equation}
and the first-order approximation for VP SDE is
\begin{equation}
\begin{aligned}
\label{app_vp_sde_1}
\frac{\tilde{\boldsymbol{x}}_{t_{i}}}{\alpha_{t_i}}&=\frac{\phi(\lambda_{t_i})}{\phi(\lambda_{t_{i-1}})}\frac{\tilde{\boldsymbol{x}}_{t_{i-1}}}{\alpha_{t_{i-1}}}+ \Big{[} 1 - \frac{\phi(\lambda_{t_{i}})}{\phi(\lambda_{t_{i-1}})} \Big{]} \boldsymbol{x}_\theta(\tilde{\boldsymbol{x}}_{\lambda_{t_{i-1}}},\lambda_{t_{i-1}})\\
&+\sqrt{\lambda_{t_i}^2 - \lambda_{t_{i-1}}^2\Big[\frac{\phi(\lambda_{t_i})}{\phi(\lambda_{t_{i-1}})}\Big]^2 }\ \boldsymbol z_{t_{i-1}}.
\end{aligned}
\end{equation}

To elaborate more clearly, we refer to the errors arising from predicting every data state in the reverse process as one-step prediction errors. For the VE SDE, one-step prediction errors can be expressed as (derivation in Supp.1.6)
\begin{equation}
\label{prediction_error}
|\boldsymbol{x}_t - \tilde{\boldsymbol{x}}_t| = \Big[1-\frac{\phi(\sigma_t)}{\phi(\sigma_s)}\Big] |\boldsymbol{x}_0(\boldsymbol{x}_{\sigma_s},\sigma_s)-\boldsymbol{x}_\theta(\boldsymbol{x}_{\sigma_s},\sigma_s)| + \check{\mathcal{R}}_{1},
\end{equation}
and for VP SDE is (derivation in Supp.1.7)
\begin{equation}
\label{prediction_error_vp}
|\boldsymbol{x}_t - \tilde{\boldsymbol{x}}_t| = \alpha_t \Big[1-\frac{\phi(\lambda_t)}{\phi(\lambda_s)}\Big] |\boldsymbol{x}_0(\boldsymbol{x}_{\lambda_s},\lambda_s)-\boldsymbol{x}_\theta(\boldsymbol{x}_{\lambda_s},\lambda_s)| + \check{\mathcal{R}}_{1}.
\end{equation}

We observe that one-step prediction errors of both VE SDE and VP SDE are influenced by the First-order It\^{o}-Taylor Expansion (FIE) coefficient $1 \!- \! \frac{\phi(x_t)}{\phi(x_s)}$, which is only determined by the noise scale function $\phi(x)$ introduced in the reverse process. As $\phi(x)$ is arbitrary, different noise scale functions correspond to different solutions, collectively forming the solution space of ER SDE (here we borrow the concept of solution space from linear algebra \cite{leon2006linear}).

\textbf{ODE Solvers exhibit rapid sampling capability while SDE Solvers demonstrate high-quality sampling ability:} Taking the first order approximation of VE SDE as an example, an intuitive strategy for reducing one-step prediction errors is to decrease the FIE coefficient. Due to $\frac{\phi\left(\sigma_{t}\right)}{\phi\left(\sigma_{s}\right)} \leq \frac{\sigma_{t}}{\sigma_{s}}$ (see Supp.1.9), the minimum value for the FIE coefficient is $1- \frac{\sigma_{t}}{\sigma_{s}}$. Interestingly, when the FIE coefficient reaches its minimum value, the ER SDE precisely reduces to ODE (in this case, $\phi(\sigma)=\sigma$). This implies that ODE solvers possess the minimal one-step prediction error, theoretically explaining the strong rapid sampling capabilities observed in Fig.\ref{FID_NFE_sde_ode}. Further analysis in Supp.1.8 reveals that ER SDE reduces to the reverse-time SDE when $\phi(\sigma)=\sigma^2$. In theory, smaller one-step prediction errors lead to higher image generation quality in the small NFE regime. However, why do SDE solvers produce higher-quality images compared with ODE solvers when increasing NFE further? This is because the larger FIE coefficient of SDE solvers corresponds to more noise during the reverse process, which gradually corrects the accumulated prediction errors \cite{karras2022elucidating}. As shown in Fig.{\ref{FEI_FID}}(a), ODE Solvers have the minimal FIE coefficient, thus demonstrating rapid sampling capability in Fig.{\ref{FEI_FID}}(b) in the small NFE regime. However, once the noise gradually corrects the accumulated prediction errors, SDE Solvers exhibits greater potential for more efficient high-quality sampling.

\textbf{VP SDE Solvers achieve parity with VE SDE Solvers:} The only difference between Eq.(\ref{app_ve_sde_1}) and Eq.(\ref{app_vp_sde_1}) lies in the latter being scaled by $1/\alpha_t$, but their relative errors remain the same. In other words, the performance of the VP SDE and the VE SDE solver is equivalent under the same NFE and pretrained model. Directly comparing them by experiments has been challenging in prior research due to the absence of a generative model simultaneously supporting both types of SDEs. This has led to divergent conclusions, with some studies \cite{song2021score} finding that VE SDE provides better sample quality than VP SDE, while others \cite{jolicoeur2021gotta} reaching the opposite conclusion. Fortunately, EDM \cite{karras2022elucidating} allows us for a fair comparison between VP SDE and VE SDE, as elaborated in Supp.3.3.

\subsection{Customized Efficient High-Quality ER-SDE-Solvers}
\label{section4_2}
In order to combine the rapid sampling performance of ODE solvers with the high-quality sampling capability of SDE solvers, we devise specialized ER-SDE-Solvers by carefully selecting noise scale functions.

To further demonstrate how the noise scale function \(\phi(x)\) directly impacts the efficiency of the sampling process, we initially provide three different forms of \(\phi(x)\):
$$\text{ER SDE 1:}\phi(x)=x^{1.5},\quad\text{ER SDE 2:}\phi(x)=x^{2.5}$$
$$\text{ER SDE 3:}\phi(x)=x^{0.9} \log _{10}\left(1+100 x\right).$$
Fig.\ref{FEI_FID} illustrates that unfavorable choices of \(\phi(x)\) (such as ER SDE 2) lead to significant one-step prediction errors and inefficient sampling. With inappropriate \(\phi(x)\) and limited NFE, the role of noise correction is not prominent. Thus, it is crucial to carefully select the noise scale function \(\phi(x)\) to achieve high-quality sampling with fewer NFE.

In order to achieve rapid sampling, we make the FIE coefficient as close as possible to the ODE case, which has the minimum one-step prediction errors. To further enhance the quality of generated images, we introduce a moderate amount of noise during the sampling process, i.e., allowing for a controlled amplification of one-step prediction errors when NFE is relatively small. Consequently, we propose a customized ER SDE solver where
\begin{equation}
\label{er_sde_5}
\text{ER SDE 4:}\quad \phi(x)=x(e^{x^{0.3}}+10).
\end{equation}
Although ER SDE 4 exhibits more significant one-step prediction errors in the early stages ($\sim$100 NFE) in Fig.\ref{FEI_FID}(a), its later-stage errors closely approach the minimum error (i.e., the error of the ODE). As a result, ER SDE 4 has the potential to generate high-quality images while maintaining efficiency of ODE solvers (see Fig.\ref{FEI_FID}(b)). 

We select ER SDE 4 as the noise scale function by default in the subsequent experiments. This strategy not only facilitates rapid sampling but also contributes to preserving the stochastic noise introduced by the reverse process, thereby enhancing higher-quality in the generated images. In fact, there are countless possible choices for $\phi(x)$, and we have only provided a few examples here. Researchers should select the suitable one based on specific application scenarios, as detailed in Supp.1.10.

\vspace{3mm}
\section{Experiments}
\label{section5}
In this section, we demonstrate that ER-SDE-Solvers can significantly accelerate the sampling process of pretrained DMs. We vary the number of function evaluations (NFE), i.e., the invocation number of the data prediction model, and compare the sample quality between ER-SDE-Solvers of different stages and other training-free samplers. For each experiment, we draw 50K samples and employ the widely-used FID score \cite{heusel2017gans} and sFID \cite{nash2021generating} to evaluate sample quality, where a lower FID/sFID typically signifies better sample quality. Detailed implementation and experimental settings refer to Supp.3.

\begin{table}[t]
\begin{small}
\begin{center}
\centering
\caption{Sample quality measured by FID$\downarrow$ on ImageNet $64\times 64$ for different stages of VE(P) ER-SDE-Solvers with EDM, varying the NFE. VE(P)-x denotes the x-th stage VE(P) ER-SDE-Solver.}
\vskip 0.10in
\label{table1}
\begin{tabular}{@{}lllllllllll@{}}
\toprule
\multicolumn{2}{l}{Method\textbackslash{}NFE} & 10    & 20   & 30   & 50     \\ \midrule
\multicolumn{2}{l}{VE-2}                      & $11.81$ & $3.67$ & $2.67$ & $2.31$ \\
\multicolumn{2}{l}{VP-2}                      & $11.94$ & $3.73$ & $2.67$ & $2.27$  \\ 
\multicolumn{2}{l}{VE-3}                      & $11.46$ & $3.45$ & $2.58$ & $2.24$ \\
\multicolumn{2}{l}{VP-3}                      & $11.32$ & $3.48$ & $2.58$ & $2.28$\\
\bottomrule
\end{tabular}
\end{center}
\end{small}
\end{table}

\begin{table}[t]
\begin{small}
\begin{center}
\caption{Sample quality measured by FID$\downarrow$ on ImageNet $64\times64$ with the pretrained model EDM, varying the NFE. The upper right $-$ indicates a reduction of NFE by one, and $+$ signifies an increase in NFE by one.}
\vskip 0.10in
\label{exp2}
\begin{tabular}{@{}lllllll@{}}
\toprule
\multicolumn{3}{l}{Sampling method\textbackslash{}NFE}    & \multicolumn{1}{l}{35} & \multicolumn{1}{l}{40} & \multicolumn{1}{l}{45} & \multicolumn{1}{l}{50}\\
\\[-2ex]
\multicolumn{3}{l}{\textbf{Stochastic Sampling}}       \\ \midrule
\multirow{2}{*}                            & \multicolumn{2}{l}{\!\!\!DDIM($\eta=1$)}               & 11.4   & 9.49 & 8.37      & 7.35 \\
                                           & \multicolumn{2}{l}{\!\!\!SDE-DPM-Solver++(2M)}  & 3.02   & 2.73 & 2.50             & 2.29    \\
                                           & \multicolumn{2}{l}{\!\!\!EDM-Stochastic}     & 2.97    & 2.82$^{-}$ & 2.57         & 2.49$^{-}$        \\
                                           & \multicolumn{2}{l}{\!\!\!SEEDS(ETD-SERK)}       & 53.7$^{+}$   & 46.9$^{-}$ & 34.0  & 25.9$^{+}$        \\
                                           & \multicolumn{2}{l}{\!\!\!Ours(ER-SDE-Solver-3)}      & \bf{2.46}   & \bf{2.30} & \bf{2.25}    & \bf{2.22}          \\ 
\\[-2ex]
\multicolumn{2}{l}{\textbf{Deterministic Sampling}}       \\ \midrule
 & \multicolumn{2}{l}{\!\!\!DDIM}                & 3.85   & 3.54 & 3.31   & 3.15          \\
                                                                                  & \multicolumn{2}{l}{\!\!\!EDM-Deterministic}      & \bf{2.46}  & 2.41$^{-}$ & 2.37       & 2.34$^{-}$         \\
                                                                                  & \multicolumn{2}{l}{\!\!\!DPM-Solver-3} & 2.48   & 2.42 & 2.38  & 2.35 \\
                                                                                  & \multicolumn{2}{l}{\!\!\!DPM-Solver++(2M)}  &2.47   &2.42 & 2.39   & 2.35 \\ 
                                                                                  & \multicolumn{2}{l}{\!\!\!SEEDS(ETD-ERK)}   & \bf{2.46}   & 2.39 & 2.37$^{-}$         & 2.34  \\\bottomrule
\end{tabular}
\end{center}
\end{small}
\end{table}

\begin{table*}[t]
  \centering
  \scriptsize
\caption{Sample quality measured by FID$\downarrow$ and sFID$\downarrow$ on class-conditional ImageNet $128\times128$ with the pretrained model Guided-diffusion (without classifier guidance, linear noise schedule), varying the NFE.}
\vskip 0.12in
\label{image_gui}
\scalebox{1.2}{
      \begin{tabular}{@{}lcccccccccccccccccc@{}}
        \toprule
        \multirow{2}{*}{\textbf{Sampling method\textbackslash{}NFE}} & \multicolumn{5}{c}{\textbf{FID$\downarrow$}} & \phantom{c} & \multicolumn{5}{c}{\textbf{sFID$\downarrow$}} \\ \cmidrule{2-6} \cmidrule{8-12}
    & \bf{20} & \bf{25} & \bf{30} & \bf{40} & \bf{50}& \phantom{c} & \bf{20} & \bf{25} & \bf{30} & \bf{40} & \bf{50} \\ \midrule
        \textbf{Stochastic Sampling}      \\ \midrule
        DDIM($\eta=1$)        &21.23  &17.31   & 14.95  &12.05 &10.42 & \phantom{c}&27.77  &22.66   &19.62   &15.67 &13.30\\
        SDE-DPM-Solver++(2M) &9.61  &8.75   & 8.38  &7.85 &7.83  & \phantom{c} &7.89  &7.03   &6.62   &5.90 &5.57 \\
        Ours(ER-SDE-Solver-3)  &\bf{8.33}  & \bf{7.87}  & \bf{7.84}   &\bf{7.78} &\bf{7.72}  & \phantom{c} &5.90  &\bf{5.39}   &\bf{5.13} &\bf{5.00} &\bf{4.88} \\
        \\[-2ex]
        \textbf{Deterministic Sampling}      \\ \midrule
        DDIM  &11.49  & 10.59  & 9.79  &8.99  &8.65 & \phantom{c} &7.88  &6.87   &6.24 &5.51  &5.27 \\
        DPM-Solver-3  &9.55  &9.45   & 9.39  &9.13 &9.02 & \phantom{c} &\bf{5.75}  &5.52   &5.30   &5.16  &5.06 \\
        DPM-Solver++(2M)   &10.24  &9.99   & 9.93   &9.76 &9.61 & \phantom{c} &6.52  &6.21   &6.01  &5.70 &5.55 \\ \bottomrule      
\end{tabular}}
\end{table*}

\begin{table}[t]
\begin{small}
\begin{center}
\caption{Sample quality measured by FID$\downarrow$ on class-conditional ImageNet $256\times256$ with the pretrained Guided-diffusion (classifier guidance scale = 2.0, linear noise schedule), varying the NFE.}
\label{image256}
\begin{tabular}{@{}lllllll@{}}
\toprule
\multicolumn{3}{l}{Sampling method\textbackslash{}NFE}    & \multicolumn{1}{l}{10} & \multicolumn{1}{l}{20} & \multicolumn{1}{l}{30} & \multicolumn{1}{l}{50}\\
\\[-2ex]
\multicolumn{3}{l}{\textbf{Stochastic Sampling}}       \\ \midrule
\multirow{2}{*}  & \multicolumn{2}{l}{\!\!\!DDIM($\eta=1$)}                 & 17.97                  & 10.23                & 8.19                  & 6.85 \\
                                           & \multicolumn{2}{l}{\!\!\!SDE-DPM-Solver++(2M)} & 9.21         & 6.01       &5.47         &5.19               \\
                                           & \multicolumn{2}{l}{\!\!\!Ours(ER-SDE-Solver-3)}    & \bf{6.24}     & \bf{4.76}       &\bf{4.62}    &\bf{4.57}          \\
\\[-2ex]
\multicolumn{2}{l}{\textbf{Deterministic Sampling}}       \\ \midrule
\multirow{2}{*} & \multicolumn{2}{l}{\!\!\!DDIM}                 & 8.63          &5.60         &5.00          &4.59          \\
                                                                                  & \multicolumn{2}{l}{\!\!\!DPM-Solver-3}       &6.45      &5.03          & 4.94 & 4.92 \\
                                                                                  & \multicolumn{2}{l}{\!\!\!DPM-Solver++(2M)}     &7.19  & 5.54 & 5.32 & 5.16 \\ \bottomrule
\end{tabular}
\end{center}
\end{small}
\end{table}

\subsection{Different Stages of VE and VP ER-SDE-Solvers}
To ensure a fair comparison between VP ER-SDE-Solvers and VE ER-SDE-Solvers, we opt for EDM \cite{karras2022elucidating} as the pretrained model, as detailed in Supp.3.3. It can be observed from Table \ref{table1} that the image generation quality produced by both of them is similar, consistent with the findings in Sec.\ref{section4_1}. Additionally, the high-stage ER-SDE-Solver-3 converges faster than ER-SDE-Solver-2, particularly in the few-step regime under 20 NFE. This is because higher stages result in more minor discretization errors, which constitute one of the components of one-step prediction errors (see Supp.1.6 and Supp.1.7). We also arrive at the same conclusions on CIFAR-10 dataset \cite{krizhevsky2009learning} as can be found in supplementary material Table.1.

\subsection{Comparisons with Other Training-Free Methods}
We compare ER-SDE-Solvers with other training-free sampling methods, including stochastic samplers such as SDE-DPM-Solver++ \cite{lu2022dpm-solver} and SEEDS \cite{gonzalez2023seeds}, as well as deterministic samplers like DDIM \cite{song2021denoising}, DPM-Solver \cite{lu2022dpm} and DPM-Solver++ \cite{lu2022dpm-solver}. Table \ref{image_gui} presents experimental results on the ImageNet $128\times 128$ dataset \cite{deng2009imagenet} using the same pretrained Guided-diffusion model \cite{dhariwal2021diffusion}. It is evident that ER-SDE-Solvers emerge as the most efficient stochastic samplers, achieving a remarkable $2\sim8\times$ speedup compared with previously state-of-the-art stochastic sampling methods. Specifically, ER-SDE-Solvers achieve high-quality sampling with FID = 7.84 requiring only 30 NFE, while SDE-DPM-Solver++ needs 50 NFE. Additionally, compared to deterministic samplers, ER-SDE-Solvers can generate higher-quality images when NFE is fixed. As illustrated in Fig.\ref{FID_NFE_sde_ode}, increasing NFE further enhances the image quality produced by ER-SDE-Solvers, whereas the improvement in image quality for deterministic samplers is limited. In summary, ER-SDE-Solvers achieve state-of-the-art performance among all stochastic samplers by enhancing image generation quality while maintaining efficiency. We also provide comparisons on ImageNet $64\times64$ using EDM \cite{karras2022elucidating} as the pretrained model in Table \ref{exp2}, yielding consistent conclusions.

Particularly, we combine ER-SDE-Solvers with \textit{classifier guidance} to generate high-resolution images. Table \ref{image256} provides comparative results on ImageNet $256\times256$ \cite{deng2009imagenet} using Guided-diffusion \cite{dhariwal2021diffusion} as the pretrained model. We surprisingly find that ER-SDE-Solvers with \textit{classifier guidance} exhibit high image generation quality even with very low NFE. This may be attributed to the customized noise injected into the sampling process, which mitigates the inaccuracies in data estimation introduced by classifier gradient guidance. Further investigation is left for future work.

\vspace{3mm}
\section{Conclusion}
We address the challenges of sampling speed and image visual quality in DMs. Initially, we formulate the sampling process as an ER SDE, which unifies ODEs and SDEs in previous studies. Leveraging the semi-linear structure of ER SDE solutions, we provide exact solutions and high-stage approximations for both VP SDE and VE SDE. Building upon it, we introduce one-step prediction errors and establish two crucial findings from mathematical standpoints: the superior performance of ODE solvers for rapid sampling and the high-quality sampling ability of SDE solvers, and the comparable performance of VP SDE solvers with VE SDE solvers. Finally, leveraging the advantages of both ODE solvers and SDE solvers, we introduce state-of-the-art efficient high-quality samplers, known as ER-SDE-Solvers.

\section*{Impact Statement}
In line with other advanced deep generative models like GANs, DMs can be harnessed to produce deceptive or misleading content, particularly in manipulated images. The efficient solvers we propose herein offer the capability to expedite the sampling process of DMs, thereby enabling faster image generation and manipulation, potentially leading to the creation of convincing but fabricated visuals. As with any technology, this acceleration could accentuate the potential ethical concerns associated with DMs, particularly their susceptibility to misuse or malicious applications. For instance, more frequent image generation might elevate the likelihood of unauthorized exposure of personal information, facilitate content forgery and dissemination of false information, and potentially infringe upon intellectual property rights.

{\small
\bibliographystyle{ieee_fullname}
\bibliography{example_paper}
}

\clearpage
\appendix

\section{Additional Proofs}
\subsection{Proof of Proposition 4.1}
\label{appendix0}
In this section, we provide the derivation process of Eq.(7) in Main Text (MT), with the key insight being that the forward and backward SDEs share the same marginal distribution.
We begin by considering the forward process. As outlined in MT Sec.3.1, the forward process can be expressed as the SDE shown in MT Eq.(1). In accordance with the Fokker-Plank Equation (also known as the Forward Kolmogorov Equation) \cite{risken1996fokker}, we obtain:
\begin{equation}\label{7}
\begin{aligned}
&\frac{\partial{p_t(\boldsymbol{x}_t)}}{\partial t}= -\nabla_{\boldsymbol{x}} [f(t)\boldsymbol{x}_t p_t(\boldsymbol{x}_t)] + \frac{\partial}{\partial x_i \partial x_j}\Big[\frac{1}{2}g^2(t)p_t(\boldsymbol{x}_t)\Big]\\
                                      &= -\nabla_{\boldsymbol{x}} [f(t)\boldsymbol{x}_t p_t(\boldsymbol{x}_t)] +  \nabla_{\boldsymbol{x}}\Big[\frac{1}{2}g^2(t)\nabla_{\boldsymbol{x}} p_t(\boldsymbol{x}_t)\Big] \\
                                      &= -\nabla_{\boldsymbol{x}} [f(t)\boldsymbol{x}_t p_t(\boldsymbol{x}_t)]+\nabla_{\boldsymbol{x}}\Big[\frac{1}{2}g^2(t)p_t(\boldsymbol{x}_t)\nabla_{\boldsymbol{x}} \log p_t(\boldsymbol{x}_t)\Big]\\
                                      &= -\nabla_{\boldsymbol{x}}\Big{\{}\Big[f(t)\boldsymbol{x}_t  - \frac{1}{2}g^2(t)\nabla_{\boldsymbol{x}} \log p_t(\boldsymbol{x}_t)\Big]p_t(\boldsymbol{x}_t)\Big{\}},
\end{aligned}
\end{equation}
where \( p(\boldsymbol x_t) \) denotes the probability density function of state \(\boldsymbol x_t \).

Most processes defined by a forward-time or conventional diffusion equation model possess a corresponding reverse-time model \cite{anderson1982reverse,song2021score}, which can be formulated as:
\begin{equation}
\label{8}
d \boldsymbol{x}_t = \mu(t, \boldsymbol{x}_t)dt + \sigma(t, \boldsymbol{x}_t)d \overline{\boldsymbol{w}}_{t}.
\end{equation}
According to the Backward Kolmogorov Equation \cite{risken1996fokker}, we have:
\begin{equation}\label{9}
\begin{aligned}
&\frac{\partial{p_t(\boldsymbol{x}_t)}}{\partial t} = -\nabla_{\boldsymbol{x}} [\mu(t, \boldsymbol{x}_t) p_t(\boldsymbol{x}_t)]- \frac{\partial}{\partial x_i \partial x_j}\Big[\frac{1}{2}\sigma^2(t, \boldsymbol{x}_t)p_t(\boldsymbol{x}_t)\Big]\\
&= -\nabla_{\boldsymbol{x}} [\mu(t, \boldsymbol{x}_t) p_t(\boldsymbol{x}_t)] - \nabla_{\boldsymbol{x}}\Big[\frac{1}{2}\sigma^2(t, \boldsymbol{x}_t)\nabla_{\boldsymbol{x}} p_t(\boldsymbol{x}_t)\Big]\\
&=-\nabla_{\boldsymbol{x}} [\mu(t, \boldsymbol{x}_t) p_t(\boldsymbol{x}_t)] - \nabla_{\boldsymbol{x}}\Big[\frac{1}{2}\sigma^2(t, \boldsymbol{x}_t)p(\boldsymbol{x}_t)\nabla \log p_t(\boldsymbol{x}_t)\Big]\\
&=-\nabla_{\boldsymbol{x}}\Big{\{}\Big{[}\mu(t, \boldsymbol{x}_t) + \frac{1}{2}\sigma^2(t, x_t)\nabla_{\boldsymbol{x}} \log p_t(\boldsymbol{x}_t)\Big{]}p_t(\boldsymbol{x}_t)\Big{\}}.
\end{aligned}
\end{equation}

We aim for the forward process and the backward process to share the same distribution, namely:
\begin{equation}
\begin{aligned}
\label{10}
&\mu(t, \boldsymbol{x}_t) + \frac{1}{2}\sigma^2(t, \boldsymbol{x}_t)\nabla_{\boldsymbol{x}} \log p_t(\boldsymbol{x}_t) = f(t)\boldsymbol{x}_t  - \frac{1}{2}g^2(t)\nabla_{\boldsymbol{x}} \log p_t(\boldsymbol{x}_t)\\
&\mu(t, \boldsymbol{x}_t) = f(t)\boldsymbol{x}_t - \frac{g^2(t) + \sigma^2(t, \boldsymbol{x}_t)}{2}\nabla_{\boldsymbol{x}} \log p(\boldsymbol{x}_t).
\end{aligned}
\end{equation}
Let $\sigma(t, \boldsymbol{x}_t) = h(t)$, yielding the Extended Reverse-Time SDE (ER-SDE):
\begin{equation}
\label{11}
d \boldsymbol{x}_t = \Big[f(t)\boldsymbol{x}_t - \frac{g^2(t) + h^2(t)}{2}\nabla_{\boldsymbol{x}} \log p_t(\boldsymbol{x}_t)\Big]d t + h(t) d\overline{\boldsymbol{w}}_{t}.
\end{equation}

\subsection{Proof of Proposition 4.2}

\label{appendix1}
Eq.(9) in MT has the following analytical solution \cite{kloeden1992numerical}:
\begin{equation}
\label{14}
\begin{aligned}  
\boldsymbol{x}_t &= e^{\int_{\sigma_s}^{\sigma_t} \frac{1}{\sigma} + \frac{\xi(\sigma)}{2\sigma^2} d \sigma}\boldsymbol{x}_s - 
\int_{\sigma_s}^{\sigma_t}e^{\int_{\sigma}^{\sigma_t}\frac{1}{\tau} + \frac{\xi(\tau)}{2\tau^2}d \tau}\Big[\frac{1}{\sigma} + \frac{\xi(\sigma)}{2\sigma^2}\Big]\boldsymbol{x}_\theta(\boldsymbol{x}_\sigma, \sigma)d \sigma \\
&+\int_{\sigma_s}^{\sigma_t}\!\!e^{\int_{\sigma}^{\sigma_t}\frac{1}{\tau} + \frac{\xi(\tau)}{2\tau^2}d \tau}\sqrt{\xi(\sigma)}d\boldsymbol{w}_\sigma.
\end{aligned}
\end{equation}
Let $\int \frac{1}{\sigma} + \frac{\xi(\sigma)}{2\sigma^2}d \sigma = \ln \phi(\sigma)$ and suppose $\phi(x)$ is derivable, then
\begin{equation}
\label{A1_1}
	\frac{1}{\sigma} + \frac{\xi(\sigma)}{2\sigma^2} = \frac{\phi^{(1)}(\sigma)}{\phi(\sigma)},
\end{equation}
where $\phi^{(1)}(x)$ is the first derivative of $\phi(x)$.

Substituting Eq.(\ref{A1_1}) into  Eq.(\ref{14}), we obtain
\begin{equation}
\label{A1_2}
\begin{aligned}
&\boldsymbol{x}_t =\frac{\phi(\sigma_t)}{\phi(\sigma_s)}\boldsymbol{x}_s\!-\!\!\int_{\sigma_s}^{\sigma_t}\!\!\frac{\phi(\sigma_t)}{\phi(\sigma)}\frac{\phi^{(1)}(\sigma)}{\phi(\sigma)}\boldsymbol{x}_\theta(\boldsymbol{x}_\sigma, \sigma)d \sigma\!+\!\!\int_{\sigma_s}^{\sigma_t}\!\!\frac{\phi(\sigma_t)}{\phi(\sigma)}\sqrt{\xi(\sigma)}d\boldsymbol{w}_\sigma\\
&=\!\!\!\underbrace{\frac{\phi(\sigma_t)}{\phi(\sigma_s)}\boldsymbol{x}_s}_\text{(a) Linear term}\!\!\!\!+\!\underbrace{\phi(\sigma_t) \!\!\int_{\sigma_t}^{\sigma_s}\!\!\frac{\phi^{(1)}(\sigma)}{\phi^2(\sigma)}\boldsymbol{x}_\theta(\boldsymbol{x}_\sigma, \sigma)d \sigma}_\text{(b) Nonlinear term}+\!\!\!\underbrace{ \int_{\sigma_s}^{\sigma_t}\!\!\frac{\phi(\sigma_t)}{\phi(\sigma)}\sqrt{\xi(\sigma)}d\boldsymbol{w}_\sigma}_\text{(c) Noise term}\!.
\end{aligned}
\end{equation}
Inspired by~\cite{lu2022dpm-solver}, the noise term $(c)$ can be computed as follows:
\begin{equation}
\label{A1_3}
(c) = -\phi(\sigma_t)\int_{\sigma_t}^{\sigma_s}\frac{\sqrt{\xi(\sigma)}}{\phi(\sigma)}d\boldsymbol{w}_\sigma = -\phi(\sigma_t) \sqrt{\int_{\sigma_t}^{\sigma_s}\frac{\xi(\sigma)}{\phi^2(\sigma)}d \sigma}\boldsymbol{z}_{s},
\end{equation}
where $\boldsymbol{z}_{s} \sim \mathcal{N}(\mathbf{0}, \boldsymbol{I})$.
Substituting Eq.(\ref{A1_1}) into Eq.(\ref{A1_3}), we have
\begin{equation}
\label{A1_4}
\begin{aligned}
(c) &= -\phi(\sigma_t) \sqrt{\int_{\sigma_t}^{\sigma_s}{\Big[2\sigma^2\frac{\phi^{(1)}(\sigma)}{\phi^3(\sigma)} - 2\sigma\frac{1}{\phi^2(\sigma)}}\Big]d \sigma}\boldsymbol{z}_{s}\\
&= -\phi(\sigma_t) \sqrt{-\frac{\sigma^2}{\phi^2(\sigma)}\Big|_{\sigma_t}^{\sigma_s}}\boldsymbol{z}_{s}\\
&=-\sqrt{\sigma_t^2 - \sigma_s^2\Big[\frac{\phi(\sigma_t)}{\phi(\sigma_s)}\Big]^2 }\boldsymbol{z}_{s}.
\end{aligned}
\end{equation}
Considering that adding Gaussian noise is equivalent to subtracting Gaussian noise, we can rewrite Eq.(\ref{A1_2}) as
\begin{equation}
\label{A1_5}
\boldsymbol{x}_t = \frac{\phi(\sigma_t)}{\phi(\sigma_s)}\boldsymbol{x}_s + \phi(\sigma_t)\!\!\!\int_{\sigma_t}^{\sigma_s}\!\!\frac{\phi^{(1)}(\sigma)}{\phi^2(\sigma)}\boldsymbol{x}_\theta(\boldsymbol{x}_\sigma,\sigma)d \sigma + \sqrt{\sigma_t^2\!-\!\sigma_s^2\Big[\frac{\phi(\sigma_t)}{\phi(\sigma_s)}\Big]^2 }\boldsymbol{z}_{s}.
\end{equation}

\subsection{Proof of Proposition 4.3}
\label{appendix3}
In order to minimize the approximation error between $\tilde{\boldsymbol{x}}_{M}$ and the true solution at time $0$, it is essential to progressively reduce the approximation error for each $\tilde{\boldsymbol{x}}_{t_{i}}$ during each step. Beginning from the preceding value $\tilde{\boldsymbol{x}}_{t_{i-1}}$ at the time $t_{i-1}$, as outlined in Eq.(\ref{A1_5}), the exact solution $\boldsymbol{x}_{{t_{i}}}$ at time $t_i$ can be determined as follows:
\begin{equation}
\label{veti}
\begin{aligned}
\boldsymbol{x}_{{t_{i}}} &= \frac{\phi(\sigma_{t_i})}{\phi(\sigma_{t_{i-1}})}\tilde{\boldsymbol{x}}_{t_{i-1}} + \phi(\sigma_{t_i})\int_{\sigma_{t_i}}^{\sigma_{t_{i-1}}}\frac{\phi^{(1)}(\sigma)}{\phi^2(\sigma)}\boldsymbol{x}_\theta(\boldsymbol{x}_\sigma,\sigma)d \sigma \\
&+ \sqrt{\sigma_{t_i}^2 - \sigma_{t_{i-1}}^2\Big[\frac{\phi(\sigma_{t_i})}{\phi(\sigma_{t_{i-1}})}\Big]^2 }\boldsymbol{z}_{{t_{i-1}}}.
\end{aligned}
\end{equation}

Inspired by~\cite{lu2022dpm}, we approximate the integral of $\boldsymbol{x}_\theta$ from $\sigma_{t_{i-1}}$ to $\sigma_{t_{i}}$ in order to compute $\tilde{\boldsymbol{x}}_{t_{i}}$ for approximating $\boldsymbol{x}_{{t_{i}}}$. Denote $\boldsymbol{x}_\theta^{(n)}(\boldsymbol{x}_\sigma, \sigma) := \frac{d^n \boldsymbol{x}_\theta(\boldsymbol{x}_\sigma, \sigma)}{d \sigma^n}$ as the $n$-th order total derivative of $\boldsymbol{x}_\theta(\boldsymbol{x}_\sigma, \sigma)$ w.r.t $\sigma$. For $k \geq 1$, the $k-1$-th order It\^{o}-Taylor expansion of $\boldsymbol{x}_\theta(\boldsymbol{x}_\sigma, \sigma)$ w.r.t $\sigma$ at $\sigma_{t_{i-1}}$ is \cite{gonzalez2023seeds}
\begin{equation}
\boldsymbol{x}_\theta(\boldsymbol{x}_\sigma, \sigma) = \sum_{n=0}^{k-1} \frac{(\sigma - \sigma_{t_{i-1}})^n}{n!} \boldsymbol{x}^{(n)}(\boldsymbol{x}_{\sigma_{t_{i-1}}}, \sigma_{t_{i-1}}) + \mathcal{R}_{k},
\end{equation}
where the residual $\mathcal{R}_{k}$ comprises of deterministic iterated integrals of length greater than $k$ and all iterated integrals with at least one stochastic component.

Substituting the above It\^{o}-Taylor expansion into Eq.(\ref{veti}), we obtain
\begin{equation}
\begin{aligned}
	&\boldsymbol{x}_{t_i} = \frac{\phi(\sigma_{t_i})}{\phi(\sigma_{t_{i-1}})}\boldsymbol{x}_{t_{i-1}} \\
 &+ \phi(\sigma_{t_{i}}) \sum_{n=0}^{k-1} \boldsymbol{x}^{(n)}(\boldsymbol{x}_{\sigma_{t_{i-1}}}, \sigma_{t_{i-1}}) 
	\int_{\sigma_{t_{i}}}^{\sigma_{t_{i-1}}} \frac{\phi^{(1)}(\sigma)}{\phi^2(\sigma)} \frac{(\sigma - \sigma_{t_{i-1}})^n}{n!} d\sigma \\
	 &+ \sqrt{\sigma_{t_i}^2 - \sigma_{t_{i-1}}^2\Big[\frac{\phi(\sigma_{t_i})}{\phi(\sigma_{t_{i-1}})}\Big]^2 }\ \boldsymbol z_{t_{i-1}}+ \tilde{\mathcal{R}}_{k},	
\end{aligned}
\end{equation}
where $\tilde{\mathcal{R}}_{k}$ can be easily obtained from $\mathcal{R}_{k}$ and $\phi(\sigma_{t_i})\int_{\sigma_{t_i}}^{\sigma_{t_{i-1}}}\frac{\phi^{(1)}(\sigma)}{\phi^2(\sigma)}d \sigma$.

When $n \geq 1$,
\begin{equation}
	\begin{aligned}
  &\quad \int_{\sigma_{t_{i}}}^{\sigma_{t_{i-1}}} \frac{\phi^{(1)}(\sigma)}{\phi^2(\sigma)} \frac{(\sigma - \sigma_{t_{i-1}})^n}{n!} d\sigma \\
&= \int_{\sigma_{t_{i}}}^{\sigma_{t_{i-1}}}  \frac{(\sigma - \sigma_{t_{i-1}})^n}{n!} d \Big{[}-\frac{1}{\phi(\sigma)}\Big{]} \\
&=  -\frac{(\sigma - \sigma_{t_{i-1}})^n}{n!}\frac{1}{\phi(\sigma)} \Big{|}_{\sigma_{t_{i}}}^{\sigma_{t_{i-1}}} - \int_{\sigma_{t_{i}}}^{\sigma_{t_{i-1}}} -\frac{1}{\phi(\sigma)} d \Big{[}  \frac{(\sigma - \sigma_{t_{i-1}})^n}{n!} \Big{]}   \\	
&= \frac{(\sigma_{t_i} - \sigma_{t_{i-1}})^n}{n! \phi(\sigma_{t_i})} + \int_{\sigma_{t_{i}}}^{\sigma_{t_{i-1}}} \frac{(\sigma - \sigma_{t_{i-1}})^{n-1}}{(n-1)!\phi(\sigma)} d \sigma.
	\end{aligned}
\end{equation}
Thus, by dropping the $\tilde{\mathcal{R}}_{k}$ contribution as in ~\cite{gonzalez2023seeds}, we have
\begin{equation}
	\begin{aligned}
\label{ve_app_ap}
		\tilde{\boldsymbol{x}}_{t_{i}} &= \frac{\phi(\sigma_{t_i})}{\phi(\sigma_{t_{i-1}})}\tilde{\boldsymbol{x}}_{t_{i-1}} + \Big{[} 1 - \frac{\phi(\sigma_{t_{i}})}{\phi(\sigma_{t_{i-1}})} \Big{]} \boldsymbol{x}_\theta(\tilde{\boldsymbol{x}}_{\sigma_{t_{i-1}}}, \sigma_{t_{i-1}})\\&+\sqrt{\sigma_{t_i}^2 - \sigma_{t_{i-1}}^2\Big[\frac{\phi(\sigma_{t_i})}{\phi(\sigma_{t_{i-1}})}\Big]^2 }\ \boldsymbol z_{t_{i-1}}\\
		 &+ \sum_{n=1}^{k-1} \boldsymbol{x}_{\theta}^{(n)}(\tilde{\boldsymbol{x}}_{\sigma_{t_{i-1}}}, \sigma_{t_{i-1}}) \frac{(\sigma_{t_i} - \sigma_{t_{i-1}})^n}{n!} \\
         &+ \sum_{n=1}^{k-1} \boldsymbol{x}_{\theta}^{(n)}(\tilde{\boldsymbol{x}}_{\sigma_{t_{i-1}}}, \sigma_{t_{i-1}}) \phi(\sigma_{t_i}) \int_{\sigma_{t_{i}}}^{\sigma_{t_{i-1}}} \frac{(\sigma - \sigma_{t_{i-1}})^{n-1}}{(n-1)!\phi(\sigma)} d \sigma,
	\end{aligned}
\end{equation}
where $k \ge 2$. Notably, when $k=1$, the summation term in Eq.(\ref{ve_app_ap}) with the upper index smaller than the lower index can be defined as an empty sum, and its value is $0$ \cite{treeby2014modeling}. In conclusion, Eq.(\ref{ve_app_ap}) holds when $k \ge 1$.

\textbf{Note:} The proposed algorithm has a global error of at least $\mathcal{O}((\sigma_{t_i} - \sigma_{t_{i-1}}))$ \cite{gonzalez2023seeds}. Therefore, when $k=1$, we refer to it as a first-order solver with a strong convergence guarantee. When $k\ge2$, we designate it as a $k$th-stage solver in accordance with the statement in ~\cite{gonzalez2023seeds}.

\subsection{Proof of Proposition 4.4}
\label{appendix4}
Eq.(13) in MT has the following analytical solution \cite{kloeden1992numerical}:
\begin{equation}
\begin{aligned}
\boldsymbol y_t &= e^{\int_{\lambda_s}^{\lambda_t} \frac{1}{\lambda} + \frac{\xi(\lambda)}{2\lambda^2} d \lambda}\boldsymbol y_s \\
& - 
\int_{\lambda_s}^{\lambda_t}e^{\int_{\lambda}^{\lambda_t}\frac{1}{\tau} + \frac{\xi(\tau)}{2\tau^2}d \tau}\Big[\frac{1}{\lambda} + \frac{\xi(\lambda)}{2\lambda^2}\Big{]}\boldsymbol x_\theta(\boldsymbol x_\lambda, \lambda)d \lambda \\
&+ \int_{\lambda_s}^{\lambda_t}e^{\int_{\lambda}^{\lambda_t}\frac{1}{\tau} + \frac{\xi(\tau)}{2\tau^2}d \tau}\sqrt{\xi(\lambda)}d\boldsymbol w_\lambda.
\end{aligned}
\end{equation}
Similar to VE SDE, let $\int \frac{1}{\lambda} + \frac{\xi(\lambda)}{2\lambda^2} d\lambda = \ln \phi(\lambda)$. Substituting $\boldsymbol y_t = \frac{\boldsymbol x_t}{\alpha_t}$, we have
\begin{equation}
\begin{aligned}
\boldsymbol x_t &= \frac{\alpha_t}{\alpha_s}\frac{\phi(\lambda_t)}{\phi(\lambda_s)}\boldsymbol x_s + \alpha_t \phi(\lambda_t)\int_{\lambda_t}^{\lambda_s} \frac{\phi^{(1)}(\lambda)}{\phi^2(\lambda)} \boldsymbol x_\theta(\boldsymbol x_\lambda, \lambda)d\lambda \\
&+ \alpha_t\int_{\lambda_s}^{\lambda_t} \frac{\phi(\lambda_t)}{\phi(\lambda)}\sqrt{\xi(\lambda)}d\boldsymbol w_\lambda.
\end{aligned}
\end{equation}
Similarly, the noise term can be computed as follows:
\begin{equation}
\begin{aligned}
	 &\quad \enspace \alpha_t\int_{\lambda_s}^{\lambda_t} \frac{\phi(\lambda_t)}{\phi(\lambda)}\sqrt{\xi(\lambda)}d\boldsymbol w_\lambda \\
	&= -\alpha_t \phi(\lambda_t) \int_{\lambda_t}^{\lambda_s} \frac{\sqrt{\xi(\lambda)}}{\phi(\lambda)}d\boldsymbol w_\lambda \\
	&= -\alpha_t \phi(\lambda_t) \sqrt{\int_{\lambda_t}^{\lambda_s} \frac{\xi(\lambda)}{\phi^2(\lambda)}d\lambda} \boldsymbol{z}_{s}  \\
	&= -\alpha_t \phi(\lambda_t) \sqrt{\int_{\lambda_t}^{\lambda_s}\Big[2\lambda^2\frac{\phi^{(1)}(\lambda)}{\phi^3(\lambda)} - 2\lambda\frac{1}{\phi^2(\lambda)}\Big]d\lambda}\boldsymbol{z}_{s} \\
	&= -\alpha_t \phi(\lambda_t) \sqrt{-\frac{\lambda^2}{\phi^2(\lambda)}\Big|_{\lambda_t}^{\lambda_s}}\boldsymbol{z}_{s} \\
	&=- \alpha_t \sqrt{\lambda_t^2 - \lambda_s^2\Big[\frac{\phi(\lambda_t)}{\phi(\lambda_s)}\Big]^2 }\boldsymbol{z}_{s}.
\end{aligned}
\end{equation}
Considering that adding Gaussian noise is equivalent to subtracting Gaussian noise. Above all, we have exact solution
\begin{equation}
\label{vp_solve}
\begin{aligned}
\boldsymbol x_t &= \frac{\alpha_t}{\alpha_s}\frac{\phi(\lambda_t)}{\phi(\lambda_s)}\boldsymbol x_s + \alpha_t \phi(\lambda_t)\int_{\lambda_t}^{\lambda_s} \frac{\phi^{(1)}(\lambda)}{\phi^2(\lambda)} \boldsymbol x_\theta(\boldsymbol x_\lambda, \lambda)d\lambda \\
&+ \alpha_t \sqrt{\lambda_t^2 - \lambda_s^2\Big[\frac{\phi(\lambda_t)}{\phi(\lambda_s)}\Big]^2 }\boldsymbol  z_s.
\end{aligned}
\end{equation}

\subsection{Proof of Proposition 4.5}
\label{appendix5}
Similarly, to diminish the approximation error between $\tilde{\boldsymbol{x}}_{M}$ and the true solution at time $0$, it is necessary to iteratively decrease the approximation error for each $\tilde{\boldsymbol{x}}_{t_{i}}$ at each step. Starting from the preceding value  $\tilde{\boldsymbol{x}}_{t_{i-1}}$ at the time $t_{i-1}$, following Eq.(\ref{vp_solve}), the precise solution $\boldsymbol{x}_{{t_{i}}}$ at time $t_i$ is derived as:
\begin{equation}
\begin{aligned}
\label{vpti}
&\boldsymbol x_{t_{i}}= \frac{\alpha_{t_{i}}}{\alpha_{t_{i-1}}}\frac{\phi(\lambda_{t_{i}})}{\phi(\lambda_{t_{i-1}})}\boldsymbol x_{t_{i-1}} + \alpha_{t_{i}} \phi(\lambda_{t_{i}})\int_{\lambda_{t_{i}}}^{\lambda_{t_{i-1}}} \frac{\phi^{(1)}(\lambda)}{\phi^2(\lambda)} \boldsymbol x_\theta(\boldsymbol x_\lambda, \lambda)d\lambda \\
&+ \alpha_{t_{i}} \sqrt{\lambda_{t_{i}}^2 -\lambda_{t_{i-1}}^2\Big[\frac{\phi(\lambda_{t_{i}})}{\phi(\lambda_s)}\Big]^2 }\boldsymbol  z_{t_{i-1}}.
\end{aligned}
\end{equation}
We also approximate integrals of $\boldsymbol x_\theta$ from $\sigma_{t_{i-1}}$ to $\sigma_{t_{i}}$ to compute $\tilde{\boldsymbol{x}}_{t_{i}}$ for $\boldsymbol{x}_{{t_{i}}}$ \cite{lu2022dpm}. Denote $\boldsymbol x_\theta^{(n)}(\boldsymbol x_\lambda, \lambda):= \frac{d^n\boldsymbol x_\theta(\boldsymbol x_\lambda, \lambda)}{d \lambda^{n}}$ as the $n$-th order total derivative of $\boldsymbol x_\theta(\boldsymbol x_\lambda, \lambda)$ w.r.t $\lambda$. For $k \geq 1$, the $k-1$-th order It\^{o}-Taylor expansion of $\boldsymbol x_\theta(\boldsymbol x_\lambda, \lambda)$ w.r.t $\lambda$ at $\lambda_{t_{i-1}}$ is \cite{gonzalez2023seeds}
\begin{equation}
	\boldsymbol x_\theta(\boldsymbol x_\lambda, \lambda) = \sum_{n=0}^{k-1} \frac{(\lambda - \lambda_{t_{i-1}})^n}{n!}\boldsymbol x^{(n)}(\boldsymbol x_{\lambda_{t_{i-1}}}, \lambda_{t_{i-1}}) + \mathcal{R}_{k},
\end{equation}
where the residual $\mathcal{R}_{k}$ comprises of deterministic iterated integrals of length greater than $k$ and all iterated integrals with at least one stochastic component.

Substituting the above It\^{o}-Taylor expansion into Eq.(\ref{vpti}), we obtain
\begin{equation}
\begin{aligned}
&\boldsymbol x_{t_i} = \frac{\alpha_{t_i}}{\alpha_{t_{i-1}}}\frac{\phi(\lambda_{t_i})}{\phi(\lambda_{t_{i-1}})}\boldsymbol x_{t_{i-1}} \\
&+ \alpha_{t_i}\phi(\lambda_{t_i})\sum_{n=0}^{k-1}\boldsymbol x^{(n)}(\boldsymbol x_{\lambda_{t_{i-1}}}, \lambda_{t_{i-1}})\int_{\lambda_{t_i}}^{\lambda_{t_{i-1}}}\frac{\phi^{(1)}(\lambda)}{\phi^2(\lambda)}\frac{(\lambda - \lambda_{t_{i-1}})^n}{n!}d\lambda
\\ &+ \alpha_{t_i}\sqrt{\lambda_{t_i}^2 - \lambda_{t_{i-1}}^2 \Big[\frac{\phi(\lambda_{t_i})}{\phi(\lambda_{t_{i-1}})}\Big]^2 }\boldsymbol z_{t_{i-1}} + \tilde{\mathcal{R}}_{k},
\end{aligned}
\end{equation}
where $\tilde{\mathcal{R}}_{k}$ can be easily obtained from $\mathcal{R}_{k}$ and $\alpha_{t_{i}} \phi(\lambda_{t_{i}})\int_{\lambda_{t_{i}}}^{\lambda_{t_{i-1}}} \frac{\phi^{(1)}(\lambda)}{\phi^2(\lambda)}d\lambda$.
When $n \geq 1$,
\begin{equation}
\begin{aligned}
&\quad \int_{\lambda_{t_{i}}}^{\lambda_{t_{i-1}}} \frac{\phi^{(1)}(\lambda)}{\phi^2(\lambda)} \frac{(\lambda - \lambda_{t_{i-1}})^n}{n!} d\lambda \\
&= \int_{\lambda_{t_{i}}}^{\lambda_{t_{i-1}}}  \frac{(\lambda - \lambda_{t_{i-1}})^n}{n!} d \Big{[}-\frac{1}{\phi(\lambda)}\Big{]} \\
&=  -\frac{(\lambda - \lambda_{t_{i-1}})^n}{n!}\frac{1}{\phi(\lambda)} \Big{|}_{\lambda_{t_{i}}}^{\lambda_{t_{i-1}}} \!-\! \int_{\lambda_{t_{i}}}^{\lambda_{t_{i-1}}} -\frac{1}{\phi(\lambda)} d \Big{[}  \frac{(\lambda - \lambda_{t_{i-1}})^n}{n!} \Big{]}   \\	
&= \frac{(\lambda_{t_i} - \lambda_{t_{i-1}})^n}{n! \phi(\lambda_{t_i})} + \int_{\lambda_{t_{i}}}^{\lambda_{t_{i-1}}} \frac{(\lambda - \lambda_{t_{i-1}})^{n-1}}{(n-1)!\phi(\lambda)} d \lambda.
\end{aligned}
\end{equation}
Thus, by dropping the $\tilde{\mathcal{R}}_{k}$ contribution as in \cite{gonzalez2023seeds}, we have
\begin{equation}
\begin{aligned}
\label{vp_app_ap}
\tilde{\boldsymbol{x}}_{t_{i}}&=\frac{\alpha_{t_i}}{\alpha_{t_{i-1}}} \frac{\phi(\lambda_{t_i})}{\phi(\lambda_{t_{i-1}})}\tilde{\boldsymbol{x}}_{t_{i-1}}+\alpha_{t_i} \Big{[} 1-\frac{\phi(\lambda_{t_{i}})}{\phi(\lambda_{t_{i-1}})} \Big{]}\boldsymbol x_\theta(\tilde{\boldsymbol x}_{\lambda_{t_{i-1}}},\lambda_{t_{i-1}})\\
&+\alpha_{t_i}\sqrt{\lambda_{t_i}^2 - \lambda_{t_{i-1}}^2\Big[\frac{\phi(\lambda_{t_i})}{\phi(\lambda_{t_{i-1}})}\Big]^2 }\ \boldsymbol z_{t_{i-1}}\\
&+ \alpha_{t_i}\sum_{n=1}^{k-1} \boldsymbol x_{\theta}^{(n)}(\tilde{\boldsymbol x}_{\lambda_{t_{i-1}}}, \lambda_{t_{i-1}})\frac{(\lambda_{t_i} - \lambda_{t_{i-1}})^n}{n!} \\
&+ \alpha_{t_i}\sum_{n=1}^{k-1} \boldsymbol x_{\theta}^{(n)}(\tilde{\boldsymbol x}_{\lambda_{t_{i-1}}}, \lambda_{t_{i-1}}) \phi(\lambda_{t_i}) \int_{\lambda_{t_{i}}}^{\lambda_{t_{i-1}}} \frac{(\lambda - \lambda_{t_{i-1}})^{n-1}}{(n-1)!\phi(\lambda)} d \lambda,
\end{aligned}
\end{equation}
where $k \ge 2$. Notably, when $k=1$, the summation term in Eq.(\ref{vp_app_ap}) with the upper index smaller than the lower index can be defined as an empty sum, and its value is $0$ \cite{treeby2014modeling}. In conclusion, Eq.(\ref{vp_app_ap}) holds when $k \ge 1$.

\textbf{Note:} The proposed algorithm has a global error of at least $\mathcal{O}((\lambda_{t_i} - \lambda_{t_{i-1}}))$ \cite{gonzalez2023seeds}. Therefore, when $k=1$, we refer to it as a first-order solver with a strong convergence guarantee. When $k\ge2$, we designate it as a $k$th-stage solver in accordance with the statement in ~\cite{gonzalez2023seeds}.

\subsection{Derivation of Eq.(18) in Main Text}
\label{appendix_new_ve}
Diffusion models gradually predicts every data state in the reverse process until obtaining the final data state $\hat{\boldsymbol{x}}_0$. We denote the errors between the final predicted data state $\hat{\boldsymbol{x}}_0$ and the true data state $\boldsymbol{x}_0$ as $|\boldsymbol{x}_0-\hat{\boldsymbol{x}}_0|$, termed as cumulative errors. Since cumulative errors are determined by the errors arising from predicting every data state in the reverse process (one-step prediction errors), the following derivation outlines the one-step prediction errors for the VE SDE.

According to Proposition 4.2, the exact solution of the VE SDE is 
\begin{equation}
\begin{aligned}
\label{ex_ve}
\boldsymbol{x}_t &= \frac{\phi(\sigma_t)}{\phi(\sigma_s)}\boldsymbol{x}_s + \phi(\sigma_t)\!\!\int_{\sigma_t}^{\sigma_s}\!\!\frac{\phi^{(1)}(\sigma)}{\phi^2(\sigma)}\boldsymbol{x}_\theta(\boldsymbol{x}_\sigma,\sigma)d \sigma \\
&+ \sqrt{\sigma_t^2\!-\!\sigma_s^2\Big[\frac{\phi(\sigma_t)}{\phi(\sigma_s)}\Big]^2 }\boldsymbol{z}_{s}.
\end{aligned}
\end{equation}
Using the first-order It\^{o}-Taylor expansion, the first-order VE ER-SDE-Solvers is
\begin{equation}
\begin{aligned}
\tilde{\boldsymbol{x}}_t\!&=\!\frac{\phi(\sigma_t)}{\phi(\sigma_s)}\boldsymbol{x}_s\!\!+\!\phi(\sigma_t)\!\!\!\!\int_{\!\sigma_t}^{\!\sigma_s}\!\!\frac{\phi^{(\!1\!)}\!(\sigma)}{\phi^2\!(\sigma)}d \sigma \boldsymbol{x}_\theta(\boldsymbol{x}_{\sigma_s},\!\sigma_s)\! \\
& +\!\!\sqrt{\sigma_t^2\!-\!\sigma_s^2\Big[\frac{\phi(\sigma_t)}{\phi(\sigma_s)}\Big]^2 }\boldsymbol{z}_{s}\!\!+\!\tilde{\mathcal{R}}_{1}\\
&= \!\frac{\phi(\sigma_t)}{\phi(\sigma_s)}\boldsymbol{x}_s\!\!+\!\phi(\sigma_t)\!\!\!\!\int_{\!\sigma_t}^{\!\sigma_s}\!\!\!d\Big[\!\!-\!\!\frac{1}{\phi(\sigma)}\Big] \boldsymbol{x}_\theta(\boldsymbol{x}_{\sigma_s},\!\sigma_s)\!\\
&+\!\sqrt{\sigma_t^2\!-\!\sigma_s^2\Big[\frac{\phi(\sigma_t)}{\phi(\sigma_s)}\Big]^2 }\boldsymbol{z}_{s}\!\!+\!\tilde{\mathcal{R}}_{1}\\
&= \frac{\phi(\sigma_t)}{\phi(\sigma_s)}\boldsymbol{x}_s + \Big[1-\frac{\phi(\sigma_t)}{\phi(\sigma_s)}\Big] \boldsymbol{x}_\theta(\boldsymbol{x}_{\sigma_s},\sigma_s) \\
&+ \sqrt{\sigma_t^2 - \sigma_s^2\Big[\frac{\phi(\sigma_t)}{\phi(\sigma_s)}\Big]^2 }\boldsymbol{z}_{s}	+ \tilde{\mathcal{R}}_{1}.
\end{aligned}
\end{equation}

When the neural network can accurately estimate the data state, i.e., $\boldsymbol{x}_\theta(\boldsymbol{x}_\sigma,\sigma) = \boldsymbol{x}_0(\boldsymbol{x}_\sigma,\sigma)$, Eq.(\ref{ex_ve}) is rewritten as
\begin{equation}
\begin{aligned}
\boldsymbol{x}_t &= \frac{\phi(\sigma_t)}{\phi(\sigma_s)}\boldsymbol{x}_s + \phi(\sigma_t)\!\!\int_{\sigma_t}^{\sigma_s}\!\!\frac{\phi^{(1)}(\sigma)}{\phi^2(\sigma)}\boldsymbol{x}_0(\boldsymbol{x}_\sigma,\sigma)d \sigma \\
&+ \sqrt{\sigma_t^2 - \sigma_s^2\Big[\frac{\phi(\sigma_t)}{\phi(\sigma_s)}\Big]^2 }\boldsymbol{z}_{s}.
\end{aligned}
\end{equation}

Similarity, using the first-order It\^{o}-Taylor expansion, we can obtain
\begin{equation}
\begin{aligned}
\boldsymbol{x}_t\!&=\!\frac{\phi(\sigma_t)}{\phi(\sigma_s)}\boldsymbol{x}_s\!\!+\!\phi(\sigma_t)\!\!\!\!\int_{\!\sigma_t}^{\!\sigma_s}\!\!\frac{\phi^{(\!1\!)}(\sigma)}{\phi^2(\sigma)}d\sigma \boldsymbol{x}_0(\boldsymbol{x}_{\sigma_s},\!\sigma_s)\!\\
&+\!\!\sqrt{\sigma_t^2\!-\!\sigma_s^2\Big[\frac{\phi(\sigma_t)}{\phi(\sigma_s)}\Big]^2 }\boldsymbol{z}_{s}\!\!+\!\mathcal{R}_{1}\\
&= \!\frac{\phi(\sigma_t)}{\phi(\sigma_s)}\boldsymbol{x}_s\!\!+\!\phi(\sigma_t)\!\!\!\!\int_{\!\sigma_t}^{\!\sigma_s}\!\!\!d\Big[\!\!-\!\!\frac{1}{\phi(\sigma)}\Big] \boldsymbol{x}_0(\boldsymbol{x}_{\sigma_s},\!\sigma_s)\!\\
&+\!\sqrt{\sigma_t^2\!-\!\sigma_s^2\Big[\frac{\phi(\sigma_t)}{\phi(\sigma_s)}\Big]^2 }\boldsymbol{z}_{s}\!\!+\!\mathcal{R}_{1}\\
&= \frac{\phi(\sigma_t)}{\phi(\sigma_s)}\boldsymbol{x}_s + \Big[1-\frac{\phi(\sigma_t)}{\phi(\sigma_s)}\Big] \boldsymbol{x}_0(\boldsymbol{x}_{\sigma_s},\sigma_s) \\
&+ \sqrt{\sigma_t^2 - \sigma_s^2\Big[\frac{\phi(\sigma_t)}{\phi(\sigma_s)}\Big]^2 }\boldsymbol{z}_{s}	+ \mathcal{R}_{1}.
\end{aligned}
\end{equation}

Hence,
\begin{equation}
\begin{aligned}
& \quad \enspace one\!-\!step\ prediction\ error \\
&= |\boldsymbol{x}_t - \tilde{\boldsymbol{x}}_t| \\
&= \Big[1-\frac{\phi(\sigma_t)}{\phi(\sigma_s)}\Big] |\boldsymbol{x}_0(\boldsymbol{x}_{\sigma_s},\sigma_s)-\boldsymbol{x}_\theta(\boldsymbol{x}_{\sigma_s},\sigma_s)| + \mathcal{R}_{1} - \tilde{\mathcal{R}}_{1}.
\end{aligned}
\end{equation}
Denote $\check{\mathcal{R}}_{1} = \mathcal{R}_{1} - \tilde{\mathcal{R}}_{1}$, $FIE = 1-\frac{\phi(\sigma_t)}{\phi(\sigma_s)}$, we have
\begin{equation}
\begin{aligned}
\label{pre_ve}
&one\!-\!step\ prediction\ error \!\\
&=\! FIE |\boldsymbol{x}_0(\boldsymbol{x}_{\sigma_s},\sigma_s)-\boldsymbol{x}_\theta(\boldsymbol{x}_{\sigma_s},\sigma_s)| \!+\! \check{\mathcal{R}}_{1}.
\end{aligned}
\end{equation}

It can be observed from Eq.(\ref{pre_ve}) that one-step prediction errors consist of estimation errors $|\boldsymbol{x}_0(\boldsymbol{x}_{\sigma_s},\sigma_s)-\boldsymbol{x}_\theta(\boldsymbol{x}_{\sigma_s},\sigma_s)|$ and discretization errors $\check{\mathcal{R}}_{1}$. The estimation errors are jointly determined by the estimation accuracy of the neural network $\boldsymbol{x}_\theta(\boldsymbol{x}_{\sigma_s},\sigma_s)$ and the FIE coefficient, while the discretization errors are influenced by the stage of the Ito-Taylor expansion. Since the estimation accuracy of the pretrained neural network $\boldsymbol{x}_\theta(\boldsymbol{x}_{\sigma_s},\sigma_s)$ is already established, controlling the FIE coefficient is necessary to reduce one-step prediction errors under given stage conditions.

\subsection{Derivation of Eq.(19) in Main Text}
\label{appendix_new_vp}
Diffusion models gradually predicts every data state in the reverse process until obtaining the final data state $\hat{\boldsymbol{x}}_0$. We denote the errors between the final predicted data state $\hat{\boldsymbol{x}}_0$ and the true data state $\boldsymbol{x}_0$ as $|\boldsymbol{x}_0-\hat{\boldsymbol{x}}_0|$, termed as cumulative errors. Since cumulative errors are determined by the errors arising from predicting every data state in the reverse process (one-step prediction errors), the following derivation outlines the one-step prediction errors for the VP SDE.

According to Proposition 4.4, the exact solution of the VP SDE is 
\begin{equation}
\begin{aligned}
\boldsymbol x_t &= \frac{\alpha_t}{\alpha_s}\frac{\phi(\lambda_t)}{\phi(\lambda_s)}\boldsymbol x_s + \alpha_t \phi(\lambda_t)\int_{\lambda_t}^{\lambda_s} \frac{\phi^{(1)}(\lambda)}{\phi^2(\lambda)} \boldsymbol x_\theta(\boldsymbol x_\lambda, \lambda)d\lambda \\
& + \alpha_t \sqrt{\lambda_t^2 - \lambda_s^2\Big[\frac{\phi(\lambda_t)}{\phi(\lambda_s)}\Big]^2 }\boldsymbol  z_s.
\end{aligned}
\end{equation}
Using the first-order It\^{o}-Taylor expansion, the first-order VP ER-SDE-Solvers is
\begin{equation}
\label{ex_vp}
\begin{aligned}
\tilde{\boldsymbol{x}}_t &= \frac{\alpha_t}{\alpha_s}\frac{\phi(\lambda_t)}{\phi(\lambda_s)}\boldsymbol{x}_s + \alpha_t \phi(\lambda_t)\int_{\lambda_t}^{\lambda_s}\frac{\phi^{(1)}(\lambda)}{\phi^2(\lambda)}d \lambda \boldsymbol{x}_\theta(\boldsymbol{x}_{\lambda_s},\lambda_s) \\
& + \alpha_t \sqrt{\lambda_t^2 - \lambda_s^2\Big[\frac{\phi(\lambda_t)}{\phi(\lambda_s)}\Big]^2 }\boldsymbol{z}_{s}	+ \tilde{\mathcal{R}}_{1}\\
&=  \frac{\alpha_t}{\alpha_s}\frac{\phi(\lambda_t)}{\phi(\lambda_s)}\boldsymbol{x}_s + \alpha_t\phi(\lambda_t)\int_{\lambda_t}^{\lambda_s} d\Big[-\frac{1}{\phi(\lambda)}\Big] \boldsymbol{x}_\theta(\boldsymbol{x}_{\lambda_s},\lambda_s) \\
& + \alpha_t\sqrt{\lambda_t^2 - \lambda_s^2\Big[\frac{\phi(\lambda_t)}{\phi(\lambda_s)}\Big]^2 }\boldsymbol{z}_{s}	+ \tilde{\mathcal{R}}_{1}\\
&= \frac{\alpha_t}{\alpha_s}\frac{\phi(\lambda_t)}{\phi(\lambda_s)}\boldsymbol{x}_s + \alpha_t\Big[1-\frac{\phi(\lambda_t)}{\phi(\lambda_s)}\Big] \boldsymbol{x}_\theta(\boldsymbol{x}_{\lambda_s},\lambda_s) \\
& + \alpha_t\sqrt{\lambda_t^2 - \lambda_s^2\Big[\frac{\phi(\lambda_t)}{\phi(\lambda_s)}\Big]^2 }\boldsymbol{z}_{s}	+ \tilde{\mathcal{R}}_{1}.
\end{aligned}
\end{equation}

When the neural network can accurately estimate the data state, i.e., $\boldsymbol{x}_\theta(\boldsymbol{x}_\lambda,\lambda) = \boldsymbol{x}_0(\boldsymbol{x}_\lambda,\lambda)$. Eq.(\ref{ex_vp}) is rewritten as
\begin{equation}
\begin{aligned}
\boldsymbol{x}_t &= \frac{\alpha_t}{\alpha_s}\frac{\phi(\lambda_t)}{\phi(\lambda_s)}\boldsymbol{x}_s + \alpha_t \phi(\lambda_t)\int_{\lambda_t}^{\lambda_s}\frac{\phi^{(1)}(\lambda)}{\phi^2(\lambda)}\boldsymbol{x}_0(\boldsymbol{x}_\lambda,\lambda)d \lambda \\
&+ \alpha_t \sqrt{\lambda_t^2 - \lambda_s^2\Big[\frac{\phi(\lambda_t)}{\phi(\lambda_s)}\Big]^2 }\boldsymbol{z}_{s}	
\end{aligned}
\end{equation}
Similarity, using the first-order It\^{o}-Taylor expansion, we can obtain
\begin{equation}
\begin{aligned}
\boldsymbol{x}_t &= \frac{\alpha_t}{\alpha_s}\frac{\phi(\lambda_t)}{\phi(\lambda_s)}\boldsymbol{x}_s +\alpha_t\phi(\lambda_t)\int_{\lambda_t}^{\lambda_s}\frac{\phi^{(1)}(\lambda)}{\phi^2(\lambda)}d \lambda \boldsymbol{x}_0(\boldsymbol{x}_{\lambda_s},\lambda_s) \\
&+ \alpha_t\sqrt{\lambda_t^2 - \lambda_s^2\Big[\frac{\phi(\lambda_t)}{\phi(\lambda_s)}\Big]^2 }\boldsymbol{z}_{s}	+ \mathcal{R}_{1}\\
&=  \frac{\alpha_t}{\alpha_s}\frac{\phi(\lambda_t)}{\phi(\lambda_s)}\boldsymbol{x}_s + \alpha_t\phi(\lambda_t)\int_{\lambda_t}^{\lambda_s} d\Big[-\frac{1}{\phi(\lambda)}\Big] \boldsymbol{x}_0(\boldsymbol{x}_{\lambda_s},\lambda_s) \\
&+ \alpha_t\sqrt{\lambda_t^2 - \lambda_s^2\Big[\frac{\phi(\lambda_t)}{\phi(\lambda_s)}\Big]^2 }\boldsymbol{z}_{s}	+ \mathcal{R}_{1}\\
&= \frac{\alpha_t}{\alpha_s}\frac{\phi(\lambda_t)}{\phi(\lambda_s)}\boldsymbol{x}_s + \alpha_t\Big[1-\frac{\phi(\lambda_t)}{\phi(\lambda_s)}\Big] \boldsymbol{x}_0(\boldsymbol{x}_{\lambda_s},\lambda_s) \\
&+ \alpha_t\sqrt{\lambda_t^2 - \lambda_s^2\Big[\frac{\phi(\lambda_t)}{\phi(\lambda_s)}\Big]^2 }\boldsymbol{z}_{s}	+ \mathcal{R}_{1}.
\end{aligned}
\end{equation}

Hence,
\begin{equation}
\begin{aligned}
& \quad \enspace one\!-\!step\ prediction\ error  \\
& = |\boldsymbol{x}_t - \tilde{\boldsymbol{x}}_t| \\
& = \alpha_t \Big[1-\frac{\phi(\lambda_t)}{\phi(\lambda_s)}\Big] |\boldsymbol{x}_0(\boldsymbol{x}_{\lambda_s},\lambda_s)-\boldsymbol{x}_\theta(\boldsymbol{x}_{\lambda_s},\lambda_s)| + \mathcal{R}_{1} - \tilde{\mathcal{R}}_{1}.
\end{aligned}
\end{equation}
Denote $\check{\mathcal{R}}_{1} = \mathcal{R}_{1} - \tilde{\mathcal{R}}_{1}$, $FIE = 1-\frac{\phi(\lambda_t)}{\phi(\lambda_s)}$, we have
\begin{equation}
\label{pre_vp}
one\!-\!step\ prediction\ error \!=\! \alpha_t FIE |\boldsymbol{x}_0(\boldsymbol{x}_{\lambda_s},\lambda_s)-\boldsymbol{x}_\theta(\boldsymbol{x}_{\lambda_s},\lambda_s)| + \check{\mathcal{R}}_{1}.
\end{equation}

It can be observed from Eq.(\ref{pre_vp}) that one-step prediction errors consist of estimation errors $|\boldsymbol{x}_0(\boldsymbol{x}_{\lambda_s},\lambda_s)-\boldsymbol{x}_\theta(\boldsymbol{x}_{\lambda_s},\lambda_s)| $ and discretization errors $\check{\mathcal{R}}_{1}$. The estimation errors are jointly determined by the estimation accuracy of the neural network $\boldsymbol{x}_\theta(\boldsymbol{x}_{\lambda_s},\lambda_s)$, $\alpha_t$ and the FIE coefficient, while the discretization errors are influenced by the stage of the Ito-Taylor expansion. Since the estimation accuracy of the pretrained neural network $\boldsymbol{x}_\theta(\boldsymbol{x}_{\lambda_s},\lambda_s)$ and $\alpha_t$ are already established, controlling the FIE coefficient is necessary to reduce one-step prediction errors under given stage conditions.

\subsection{Relationship with SDE and ODE}
\label{appendixa6}
We derive conditions under which ER SDE reduces to SDE \cite{song2021score} and ODE \cite{song2021score} from the perspective of noise scale function selection.

\textbf{Related to SDE.} When $h(t) = g(t)$, ER SDE reduces to SDE.

For the VE SDE, we have
\begin{equation}
 h^2(t) = g^2(t) = \frac{d\sigma_t^2}{dt} = 2\sigma_t \frac{d\sigma_t}{dt}.
\end{equation}
Since $h^2(t) = \xi(t)\frac{d\sigma_t}{dt}$, we further obtain $\xi(t) = 2\sigma_t$ and $\xi(\sigma) = 2\sigma$. Substituting it into $\int \frac{1}{\sigma} + \frac{\xi(\sigma)}{2\sigma^2}d \sigma = \ln \phi(\sigma)$, we have
\begin{equation}
\begin{aligned}
	\int \frac{1}{\sigma} + \frac{2\sigma}{2\sigma^2}d\sigma &= \ln\phi(\sigma)\\
\phi(\sigma)&=\sigma^2.
\end{aligned}
\end{equation}
For the VP SDE, we also have
\begin{equation}
h(t) = g(t) = \sqrt{\frac{d\sigma_t^2}{dt} - 2\frac{d\log\alpha_t}{dt}\sigma_t^2}.
\end{equation}
Thus,
\begin{equation}
\begin{aligned}
\label{eta}
\eta(t) &= \frac{h(t)}{\alpha_t} = \frac{1}{\alpha_t}\sqrt{\frac{d\sigma_t^2}{dt} - 2\frac{d\log\alpha_t}{dt}\sigma_t^2}\\
\eta^2(t) &= \frac{1}{\alpha_t^2}(\frac{d\sigma_t^2}{dt} - 2\frac{d\log\alpha_t}{dt}\sigma_t^2) 
= 2 \frac{\sigma_t}{\alpha_t}\Big(\frac{1}{\alpha_t}\frac{d\sigma_t}{dt}-\frac{\sigma_t}{\alpha_t^2}\frac{d\alpha_t}{dt}\Big).
\end{aligned}
\end{equation}
By $\lambda_t = \frac{\sigma_t}{\alpha_t}$, it holds that
\begin{equation}
\label{lamb}
d\lambda_t = \frac{1}{\alpha_t}\frac{d\sigma_t}{dt} - \frac{\sigma_t}{\alpha_t^2} \frac{d \alpha_t}{dt}.
\end{equation}
Substituting Eq.(\ref{lamb}) into Eq.(\ref{eta}), we have
\begin{equation}
\eta^2(t) = 2\frac{\sigma_t}{\alpha_t}\frac{d\lambda_t}{dt} = 2\lambda_t \frac{d\lambda_t}{dt}.
\end{equation}
Since $\eta^2(t)=\xi(t)\frac{d\lambda_t}{dt}$, we further obtain $\xi(t) = 2\lambda_t$ and $\xi(\lambda) = 2\lambda$. Substituting it into $\int \frac{1}{\lambda} + \frac{\xi(\lambda)}{2\sigma^2}d \lambda = \ln \phi(\lambda)$, we have
\begin{equation}
\begin{aligned}
\int \frac{1}{\lambda} + \frac{2\lambda}{2\lambda^2}d\lambda &= \ln\phi(\lambda)\\
\phi(\lambda) &= \lambda^2.
\end{aligned}
\end{equation}
Therefore, when $\phi(x)=x^2$, ER SDE reduces to SDE.

\textbf{Related to ODE.} When $h(t) = 0$, ER SDE reduces to ODE.

For the VE SDE, we have $\xi(t) = 0$ and $\xi(\sigma) = 0$. Substituting it into $\int \frac{1}{\sigma} + \frac{\xi(\sigma)}{2\sigma^2}d \sigma = \ln \phi(\sigma)$, we have
\begin{equation}
\begin{aligned}
	\int \frac{1}{\sigma}d\sigma &= \ln\phi(\sigma)\\
\phi(\sigma)&=\sigma.
\end{aligned}
\end{equation}
For the VP SDE, we also have $\eta(t)=0,\xi(t) = 0$ and $\xi(\sigma) = 0$. Substituting it into $\int \frac{1}{\lambda} + \frac{\xi(\lambda)}{2\sigma^2}d \lambda = \ln \phi(\lambda)$, we have
\begin{equation}
\begin{aligned}
	\int \frac{1}{\lambda}d\lambda &= \ln\phi(\lambda)\\
\phi(\lambda)&=\lambda.
\end{aligned}
\end{equation}
Therefore, when $\phi(x)=x$, ER SDE reduces to ODE.

\subsection{Restriction of $\phi(x)$}
\label{res_phi}
Due to the fact that the constraint equation $\int \frac{1}{\sigma} + \frac{\xi(\sigma)}{2\sigma^2}d \sigma = \ln \phi(\sigma)$ for VE SDE and VP SDE have the same expression form (see Eq.(10) and Eq.(14) in MT), VE SDE is used as an example for derivation here. 

Since $\phi(x)$ satisfies $\frac{1}{\sigma} + \frac{\xi(\sigma)}{2\sigma^2} = \frac{\phi^{(1)}(\sigma)}{\phi(\sigma)}$ and $\xi(\sigma)\geq 0$, we have
\begin{equation}
\label{cons_eq}
	\frac{\phi^{(1)}(\sigma)}{\phi(\sigma)} \geq \frac{1}{\sigma}.
\end{equation}

Suppose $t < s$ and $\sigma(t)$ is monotonically increasing, we have
\begin{equation}
	\phi^{(1)}(\sigma_s) = \lim_{t \to s} \frac{\phi(\sigma_s) - \phi(\sigma_t)}{\sigma_s - \sigma_t}.
\end{equation}
Combining with Eq.\ref{cons_eq}, we obtain
\begin{equation}
	\lim_{t \to s} \frac{\phi(\sigma_s) - \phi(\sigma_t)}{\sigma_s - \sigma_t} \geq \frac{\phi(\sigma_s)}{\sigma_s},
\end{equation}
which means that when $t$ is in the left neighboring domain of $s$, it satisfies
\begin{equation}
\label{restiction}
	\frac{\phi(\sigma_t)}{\phi(\sigma_s)} \leq \frac{\sigma_t}{\sigma_s}.
\end{equation}
In fact, Eq.(\ref{restiction}) is consistent with the limitation of noise term in MT Eq.(10), which is 
\begin{equation}
\begin{aligned}
\sqrt{\sigma_t^2 - \sigma_s^2\Big[\frac{\phi(\sigma_t)}{\phi(\sigma_s)}\Big]^2 } &\geq 0  \\
\Big[\frac{\phi(\sigma_t)}{\phi(\sigma_s)}\Big]^2 &\leq \frac{\sigma_t^2}{\sigma_s^2}  \\
\frac{\phi(\sigma_t)}{\phi(\sigma_s)} &\leq \frac{\sigma_t}{\sigma_s}.
\end{aligned}
\end{equation}
\subsection{Customize the Noise Scale Function $\phi(x)$}
\label{customize_nosie_scale}
The noise scale function $\phi(x)$ constitutes the solution space of ER SDE. In this section, we elaborate on two ways to customize $\phi(x)$.

\textbf{Indirect determination method via the intermediate variable $\xi(x)$:} For the ER SDE in Main Text Eq.(7), $h(t)$ can take any function as long as $h(t)\geq0$. For the VE SDE, $h^2(t) = \xi(t)\frac{d\sigma_t}{dt}$ and $\frac{d\sigma_t}{dt} \geq 0$. Thus, $\xi(t)$ can take on any arbitrary function, provided that $\xi(t)\geq0$ and $\xi(\sigma)$ also meets this condition. For the VP SDE, $h(t) = \eta(t)\alpha_t$, $\eta^2(t)=\xi(t)\frac{d\lambda_t}{dt}$ and $\frac{d\lambda_t}{dt} \geq 0$. Similarly, $\xi(\lambda)$ is also can take any function as long as $\xi(\lambda)\geq0$. In general, as long as $\xi(x)$ takes a function which satisfies $\xi(x)\geq 0$, and then based on the constraint equation $\int \frac{1}{\sigma} + \frac{\xi(\sigma)}{2\sigma^2}d \sigma = \ln \phi(\sigma)$, we can obtain the corresponding $\phi(x)$. Then, by plotting the FIE-step curve (just like MT Fig.3(a)), we can observe the discretization errors caused by the determined $\phi(x)$.

\textbf{Direct determination method:} The first method involves determining an initial function $\xi(x)$ and then obtaining the corresponding $\phi(x)$ by computing an indefinite integral $\int \frac{1}{\sigma} + \frac{\xi(\sigma)}{2\sigma^2}d \sigma$. However, this integral is not easy to calculate in many cases. Sec.\ref{res_phi} mentions that this indefinite integral is equivalent to Eq.(\ref{restiction}) and the right-hand side of the inequality corresponds to the ODE case (see Sec.\ref{appendixa6}). Therefore, $\phi(x)$ can be directly written without considering the specific expression of $\xi(x)$. The FIE-step curve can be plotted to check whether $\phi(x)$ is valid. More specifically, only if the FIE-step curve drawn based on $\phi(x)$ lies above the ODE curve, $\phi(x)$ is valid.

\section{Algorithms}
\label{appendixb}
The first to third-stage solvers for VE SDE are provided in Algorithm \ref{alg:ersde_solver_1_ve}, \ref{alg:ersde_solver_2_ve} and \ref{alg:ersde_solver_3_ve}. The detailed VP ER-SDE-Solver-1, 2, 3 are listed in Algorithms \ref{alg:ersde_solver_1_vp}, \ref{alg:ersde_solver_2_vp}, \ref{alg:ersde_solver_3_vp} respectively.

\begin{algorithm}
\centering
\caption{VE ER-SDE-Solver-1(Euler).}
\label{alg:ersde_solver_1_ve}
\begin{algorithmic}
\Statex {\bfseries Input:} initial value $\boldsymbol x_T$, time steps $\{t_i\}_{i=0}^M$, customized noise scale function $\phi$, data prediction model $\boldsymbol x_\theta$.
\Statex $\boldsymbol x_{t_0} \gets \boldsymbol x_T$ 

\For{$i\gets 1$ to $M$}
\Statex $r_i \gets \frac{\phi(\sigma_{t_{i}})}{\phi(\sigma_{t_{i-1}})}$
\Statex $\boldsymbol z \sim \mathcal{N}(\mathbf{0},\mathbf{I})$
\Statex $\boldsymbol n_{t_i} \gets \sqrt{\sigma_{t_{i}}^2 - r_i^2 \sigma_{t_{i-1}}^2 } \boldsymbol z$
\Statex $\tilde{\boldsymbol x}_0 \gets \boldsymbol x_\theta(\boldsymbol x_{t_{i-1}}, t_{i-1})$
\Statex $\boldsymbol x_{t_i} \gets r_i \boldsymbol x_{t_{i-1}} + (1 - r_i)\tilde{\boldsymbol x}_0 + \boldsymbol n_{t_i}$ 
\EndFor
\Statex {\bfseries Return:} $\boldsymbol x_{t_M}$
\end{algorithmic}
\end{algorithm}

\begin{algorithm}
\centering
\caption{VP ER-SDE-Solver-1(Euler).}\label{alg:ersde_solver_1_vp}
\begin{algorithmic}
\Statex {\bfseries Input:} initial value $\boldsymbol x_T$, time steps $\{t_i\}_{i=0}^M$, customized noise scale function $\phi$, data prediction model $\boldsymbol x_\theta$.
\Statex $\boldsymbol x_{t_0} \gets \boldsymbol x_T$
\For{$i\gets 1$ to $M$}
\Statex $r_i \gets \frac{\phi(\lambda_{t_{i}})}{\phi(\lambda_{t_{i-1}})}$
\Statex $r_{\alpha_i} \gets \frac{\alpha_{t_i}}{\alpha_{t_{i-1}}}$
\Statex $\boldsymbol z \sim \mathcal{N}(\mathbf{0},\mathbf{I})$
\Statex $\boldsymbol n_{t_i} \gets \alpha_{t_i}\sqrt{\lambda_{t_{i}}^2 - r_i^2 \lambda_{t_{i-1}}^2 } \boldsymbol z$
\Statex $\tilde{\boldsymbol x}_0 \gets \boldsymbol x_\theta(\boldsymbol x_{t_{i-1}}, t_{i-1})$
\Statex $\boldsymbol x_{t_i} \gets r_{\alpha_i} r_i \boldsymbol x_{t_{i-1}} + \alpha_{t_i}(1 - r_i)\tilde{\boldsymbol x}_0 + \boldsymbol n_{t_i}$ 
\EndFor
\Statex {\bfseries Return:} $\boldsymbol x_{t_M}$
\end{algorithmic}
\end{algorithm}

\begin{algorithm}
\centering
\caption{VP ER-SDE-Solver-2.}\label{alg:ersde_solver_2_vp}
\begin{algorithmic}
\Statex {\bfseries Input:} initial value $\boldsymbol x_T$, time steps $\{t_i\}_{i=0}^M$, customized noise scale function $\phi$, 
data prediction model $\boldsymbol x_\theta$, number of numerical integration points $N$.
\Statex $\boldsymbol x_{t_0} \gets \boldsymbol x_T$, $Q \gets None$
\For{$i\gets 1$ to $M$}
\Statex $r_i \gets \frac{\phi(\lambda_{t_{i}})}{\phi(\lambda_{t_{i-1}})}$, $r_{\alpha_i} \gets \frac{\alpha_{t_i}}{\alpha_{t_{i-1}}}$, $\boldsymbol z \sim \mathcal{N}(\mathbf{0},\mathbf{I})$
\Statex $\boldsymbol n_{t_i} \gets \alpha_{t_i}\sqrt{\lambda_{t_{i}}^2 - r_i^2 \lambda_{t_{i-1}}^2 } \boldsymbol z$
\If{$Q = None$}  
\Statex $\boldsymbol x_{t_i} \gets r_{\alpha_i} r_i \boldsymbol x_{t_{i-1}} + \alpha_{t_i}(1 - r_i)\boldsymbol x_\theta(\boldsymbol x_{t_{i-1}}, t_{i-1}) + \boldsymbol n_{t_i}$ 
\Else
\Statex $\Delta_{\lambda} \gets \frac{\lambda_{t_{i-1}} - \lambda_{t_{i}}}{N}$
\Statex $S_i \gets \sum_{k = 0}^{N-1} \frac{\Delta_{\lambda}}{\phi(\lambda_{t_{i}}+k\Delta_{\lambda})}$    
\Comment{Numerical integration}	
\Statex $\mathbf{D}_i \gets \frac{\boldsymbol x_\theta(\boldsymbol x_{t_{i-1}}, t_{i-1}) - \boldsymbol x_\theta(\boldsymbol x_{t_{i-2}}, t_{i-2})}{\lambda_{t_{i-1}} - \lambda_{t_{i-2}}}$
\Statex $\boldsymbol{\delta}_{t_i} \gets \alpha_{t_i}[\lambda_{t_{i}} - \lambda_{t_{i-1}} + S_i\phi(\lambda_{t_{i}})]\mathbf{D}_i$
\Statex $\boldsymbol x_{t_i} \gets r_{\alpha_i} r_i \boldsymbol x_{t_{i-1}} + \alpha_{t_i}(1 - r_i)\boldsymbol x_\theta(\boldsymbol x_{t_{i-1}}, t_{i-1}) +\boldsymbol{\delta}_{t_i}+\boldsymbol n_{t_i}$
\EndIf
\Statex $Q \xleftarrow[]{\text{buffer}} \boldsymbol x_\theta(\boldsymbol x_{t_{i-1}}, t_{i-1})$ 
\EndFor 
\Statex {\bfseries Return:} $\boldsymbol x_{t_M}$
\end{algorithmic}
\end{algorithm}

\begin{algorithm}[H]
\centering
\caption{VE ER-SDE-Solver-2.}
\label{alg:ersde_solver_2_ve}
\begin{algorithmic}
\Statex {\bfseries Input:} initial value $\boldsymbol x_T$, time steps $\{t_i\}_{i=0}^M$, customized noise scale function $\phi$, data prediction model $\boldsymbol x_\theta$, number of numerical integration points $N$.
\Statex $\boldsymbol x_{t_0} \gets \boldsymbol x_T$
\Statex $Q \gets None$
\For{$i\gets 1$ to $M$}
\Statex $r_i \gets \frac{\phi(\sigma_{t_{i}})}{\phi(\sigma_{t_{i-1}})}$
\Statex $\boldsymbol z \sim \mathcal{N}(\mathbf{0},\mathbf{I})$
\Statex $\boldsymbol n_{t_i} \gets \sqrt{\sigma_{t_{i}}^2 - r_i^2 \sigma_{t_{i-1}}^2 } \boldsymbol z$
\If{$Q = None$}
\Statex $\boldsymbol x_{t_i} \gets r_i \boldsymbol x_{t_{i-1}} + (1 - r_i)\boldsymbol x_\theta(\boldsymbol x_{t_{i-1}}, t_{i-1}) + \boldsymbol n_{t_i}$ 
\Else
\Statex $\Delta_{\sigma} \gets \frac{\sigma_{t_{i-1}} - \sigma_{t_{i}}}{N}$
\Statex $S_i \gets \sum_{k = 0}^{N-1} \frac{\Delta_{\sigma}}{\phi(\sigma_{t_{i}}+k\Delta_{\sigma})}$
\Comment {Numerical integration}	
\Statex $\mathbf{D}_i \gets \frac{\boldsymbol x_\theta(\boldsymbol x_{t_{i-1}}, t_{i-1}) - \boldsymbol x_\theta(\boldsymbol x_{t_{i-2}}, t_{i-2})}{\sigma_{t_{i-1}} - \sigma_{t_{i-2}}}$
\Statex $\boldsymbol{\delta}_{t_i} \gets [\sigma_{t_{i}} - \sigma_{t_{i-1}} + S_i\phi(\sigma_{t_{i}})]\mathbf{D}_i$
\Statex $\boldsymbol x_{t_i} \gets r_i \boldsymbol x_{t_{i-1}} + (1 - r_i)\boldsymbol x_\theta(\boldsymbol x_{t_{i-1}}, t_{i-1})+\boldsymbol{\delta}_{t_i}+\boldsymbol n_{t_i}$
\EndIf	
\Statex $Q \xleftarrow[]{\text{buffer}} \boldsymbol x_\theta(\boldsymbol x_{t_{i-1}}, t_{i-1})$ 
\EndFor 
\Statex {\bfseries Return:} $\boldsymbol x_{t_M}$
\end{algorithmic}
\end{algorithm}

\begin{algorithm}[ht]
\centering
\caption{VE ER-SDE-Solver-3.}\label{alg:ersde_solver_3_ve}
\begin{algorithmic}
\Statex {\bfseries Input:} initial value $\boldsymbol x_T$, time steps $\{t_i\}_{i=0}^M$, customized noise scale function $\phi$, data prediction model $\boldsymbol x_\theta$, number of numerical integration points $N$.
\Statex $\boldsymbol x_{t_0} \gets \boldsymbol x_T$, $Q \gets None$, $Q_d \gets None$
\For{$i\gets 1$ to $M$}
\Statex $r_i \gets \frac{\phi(\sigma_{t_{i}})}{\phi(\sigma_{t_{i-1}})}$, $\boldsymbol z \sim \mathcal{N}(\mathbf{0},\mathbf{I})$
\Statex $\boldsymbol n_{t_i} \gets \sqrt{\sigma_{t_{i}}^2 - r_i^2 \sigma_{t_{i-1}}^2 } \boldsymbol z$
\If{$Q = None$ and $Q_d = None$}  
\Statex $\boldsymbol x_{t_i} \gets r_i \boldsymbol x_{t_{i-1}} + (1 - r_i)\boldsymbol x_\theta(\boldsymbol x_{t_{i-1}}, t_{i-1}) + \boldsymbol n_{t_i}$ 
\ElsIf {$Q \neq None$ and $Q_d = None$}
\Statex $\Delta_{\sigma} \gets \frac{\sigma_{t_{i-1}} - \sigma_{t_{i}}}{N}$, $S_i \gets \sum_{k = 0}^{N-1} \frac{\Delta_{\sigma}}{\phi(\sigma_{t_{i}}+k\Delta_{\sigma})}$ 
\Comment{Numerical integration}	
\Statex $\mathbf{D}_i \gets \frac{\boldsymbol x_\theta(\boldsymbol x_{t_{i-1}}, t_{i-1}) - \boldsymbol x_\theta(\boldsymbol x_{t_{i-2}}, t_{i-2})}{\sigma_{t_{i-1}} - \sigma_{t_{i-2}}}$
\Statex $\boldsymbol{\delta}_{t_i} \gets [\sigma_{t_{i}} - \sigma_{t_{i-1}} + S_i\phi(\sigma_{t_{i}})]\mathbf{D}_i$
\Statex $\boldsymbol x_{t_i} \gets r_i \boldsymbol x_{t_{i-1}} + (1 - r_i)\boldsymbol x_\theta(\boldsymbol x_{t_{i-1}}, t_{i-1})+\boldsymbol{\delta}_{t_i}+\boldsymbol n_{t_i}$
\Statex $Q_d \xleftarrow[]{\text{buffer}} \mathbf{D}_i$
\Else
\Statex $\Delta_{\sigma} \gets \frac{\sigma_{t_{i-1}} - \sigma_{t_{i}}}{N}$, $S_i \gets \sum_{k = 0}^{N-1} \frac{\Delta_{\sigma}}{\phi(\sigma_{t_{i}}+k\Delta_{\sigma})}$, $S_{d_i} \gets \sum_{k = 0}^{N-1} \frac{\sigma_{t_{i}} + k\Delta_{\sigma}-\sigma_{t_{i-1}}}{\phi(\sigma_{t_{i}}+k\Delta_{\sigma})}\Delta_{\sigma}$
\Comment{Numerical integration}	
\Statex $\mathbf{D}_i \gets \frac{\boldsymbol x_\theta(\boldsymbol x_{t_{i-1}}, t_{i-1}) - \boldsymbol x_\theta(\boldsymbol x_{t_{i-2}}, t_{i-2})}{\sigma_{t_{i-1}} - \sigma_{t_{i-2}}}$, $\mathbf{U}_{i} \gets \frac{\mathbf{D}_i - \mathbf{D}_{i-1}}{\frac{\sigma_{t_{i-1}} - \sigma_{t_{i-3}}}{2}}$
\Statex $\boldsymbol{\delta}_{t_i} \gets [\sigma_{t_{i}} - \sigma_{t_{i-1}} + S_i\phi(\sigma_{t_{i}})]\mathbf{D}_i$
\Statex $\boldsymbol{\delta}_d{_{t_i}} \gets \Big[\frac{(\sigma_{t_{i}} - \sigma_{t_{i-1}})^2}{2} + S_{d_i}\phi(\sigma_{t_{i}})\Big]\mathbf{U}_{i}$
\Statex $\boldsymbol x_{t_i} \gets r_i \boldsymbol x_{t_{i-1}} + (1 - r_i)\boldsymbol x_\theta(\boldsymbol x_{t_{i-1}}, t_{i-1}) +
\boldsymbol{\delta}_{t_i} + \boldsymbol{\delta}_d{_{t_i}} + \boldsymbol n_{t_i}$
\Statex $Q_d \xleftarrow[]{\text{buffer}} \mathbf{D}_i$
\EndIf
\Statex $Q \xleftarrow[]{\text{buffer}} \boldsymbol x_\theta(\boldsymbol x_{t_{i-1}}, t_{i-1})$ 
\EndFor
\Statex {\bfseries Return:} $\boldsymbol x_{t_M}$
\end{algorithmic}
\end{algorithm}
\vspace{-4mm}

\begin{algorithm}[ht]
\centering
\caption{VP ER-SDE-Solver-3.}\label{alg:ersde_solver_3_vp}
\begin{algorithmic}
\Statex {\bfseries Input:} initial value $\boldsymbol x_T$, time steps $\{t_i\}_{i=0}^M$, customized noise scale function $\phi$, data prediction model $\boldsymbol x_\theta$, number of numerical integration points $N$.
\Statex $\boldsymbol x_{t_0} \gets \boldsymbol x_T$, $Q \gets None$, $Q_d \gets None$
\For{$i\gets 1$ to $M$}
\Statex $r_i \gets \frac{\phi(\lambda_{t_{i}})}{\phi(\lambda_{t_{i-1}})}$, $r_{\alpha_i} \gets \frac{\alpha_{t_i}}{\alpha_{t_{i-1}}}$, $\boldsymbol z \sim \mathcal{N}(\mathbf{0},\mathbf{I})$
\Statex $\boldsymbol n_{t_i} \gets \alpha_{t_i}\sqrt{\lambda_{t_{i}}^2 - r_i^2 \lambda_{t_{i-1}}^2 } \boldsymbol z$
\If {$Q = None$ and $Q_d = None$}  
\Statex $\boldsymbol x_{t_i} \gets r_{\alpha_i} r_i \boldsymbol x_{t_{i-1}} + \alpha_{t_i}(1 - r_i)\boldsymbol x_\theta(\boldsymbol x_{t_{i-1}}, t_{i-1}) + \boldsymbol n_{t_i}$  
\ElsIf{$Q \neq None$ and $Q_d = None$}
\Statex $\Delta_{\lambda} \gets \frac{\lambda_{t_{i-1}} - \lambda_{t_{i}}}{N}$, $S_i \gets \sum_{k = 0}^{N-1} \frac{\Delta_{\lambda}}{\phi(\lambda_{t_{i}}+k\Delta_{\lambda})}$
\Comment{Numerical integration}
\Statex $\mathbf{D}_i \gets \frac{\boldsymbol x_\theta(\boldsymbol x_{t_{i-1}}, t_{i-1}) - \boldsymbol x_\theta(\boldsymbol x_{t_{i-2}}, t_{i-2})}{\lambda_{t_{i-1}} - \lambda_{t_{i-2}}}$
\Statex $\boldsymbol{\delta}_{t_i} \gets \alpha_{t_i}[\lambda_{t_{i}} - \lambda_{t_{i-1}} + S_i\phi(\lambda_{t_{i}})]\mathbf{D}_i$
\Statex $\boldsymbol x_{t_i} \gets r_{\alpha_i} r_i \boldsymbol x_{t_{i-1}} + \alpha_{t_i}(1 - r_i)\boldsymbol x_\theta(\boldsymbol x_{t_{i-1}}, t_{i-1}) +\boldsymbol{\delta}_{t_i}+\boldsymbol n_{t_i}$
\Statex $Q_d \xleftarrow[]{\text{buffer}} \mathbf{D}_i$
\Else
\Statex $\Delta_{\lambda} \gets \frac{\lambda_{t_{i-1}} - \lambda_{t_{i}}}{N}$, $S_i \gets \sum_{k = 0}^{N-1} \frac{\Delta_{\lambda}}{\phi(\lambda_{t_{i}}+k\Delta_{\lambda})}$, $S_{d_i} \gets \sum_{k = 0}^{N-1} \frac{\lambda_{t_{i}} + k\Delta_{\lambda}-\lambda_{t_{i-1}}}{\phi(\lambda_{t_{i}}+k\Delta_{\lambda})}\Delta_{\lambda}$
\Comment{Numerical integration}
\Statex $\mathbf{D}_i \gets \frac{\boldsymbol x_\theta(\boldsymbol x_{t_{i-1}}, t_{i-1}) - \boldsymbol x_\theta(\boldsymbol x_{t_{i-2}}, t_{i-2})}{\lambda_{t_{i-1}} -\lambda_{ t_{i-2}}}$
\Statex $\mathbf{U}_{i} \gets \frac{\mathbf{D}_i - \mathbf{D}_{i-1}}{\frac{\lambda_{t_{i-1}} - \lambda_{t_{i-3}}}{2}}$
\Statex $\boldsymbol{\delta}_{t_i} \gets \alpha_{t_i}[\lambda_{t_{i}} - \lambda_{t_{i-1}} + S_i\phi(\lambda_{t_{i}})]\mathbf{D}_i$, $\boldsymbol{\delta}_d{_{t_i}} \gets \alpha_{t_i}\Big[\frac{(\lambda_{t_{i}} - \lambda_{t_{i-1}})^2}{2} + S_{d_i}\phi(\lambda_{t_{i}})\Big]\mathbf{U}_{i}$
\Statex $\boldsymbol x_{t_i} \gets r_{\alpha_i} r_i \boldsymbol x_{t_{i-1}} + \alpha_{t_i}(1 - r_i)\boldsymbol x_\theta(\boldsymbol x_{t_{i-1}}, t_{i-1}) +\boldsymbol{\delta}_{t_i} + \boldsymbol{\delta}_d{_{t_i}} + \boldsymbol n_{t_i}$
\Statex $Q_d \xleftarrow[]{\text{buffer}} \mathbf{D}_i$
\EndIf
\Statex $Q \xleftarrow[]{\text{buffer}} \boldsymbol x_\theta(\boldsymbol x_{t_{i-1}}, t_{i-1})$ 
\EndFor
 \Statex {\bfseries Return:} $\boldsymbol x_{t_M}$
\end{algorithmic}
\end{algorithm}

\vspace{-4mm}

\section{Implementation Details}
\label{appendixc}
We test our sampling method on VE-type and VP-type pretrained diffusion models. For VE-type, we select EDM \cite{karras2022elucidating} as the pretrained diffusion model. Although EDM differs slightly from the commonly used VE-type diffusion model \cite{song2021score}, its forward process still follows the VE SDE described in MT Eq.(2). Therefore, it can be regarded as a generalized VE-type diffusion model, which is proved in \ref{EDM_VE_VP}. Furthermore, EDM provides a method that can be converted into VP-type diffusion models, facilitating a fair comparison between VE SDE solvers and VP SDE solvers using the same model weights. For VP-type pretrained diffusion models, we choose widely-used Guided-diffusion \cite{dhariwal2021diffusion}. Different from EDM, Guided-diffusion provides pretrained models on high-resolution datasets, such as ImageNet $128\times128$ \cite{deng2009imagenet}, LSUN $256\times256$ \cite{yu2015lsun} and so on. For all experiments, we evaluate our method on NVIDIA GeForce RTX 3090 GPUs.

\subsection{Step Size Schedule}
\label{stepsize}
With EDM pretrained model, the time step size of ER-SDE-Solvers and other comparison methods are aligned with EDM \cite{karras2022elucidating}. The specific time step size is
\begin{equation}
t_{i< M}=\Big[\sigma_{\max}^{\frac{1}{\rho}}+\frac{i}{M-1}(\sigma_{\min}^{\frac{1}{\rho}}-\sigma_{\max}^{\frac{1}{\rho}})\Big]^{\rho},
\end{equation}
where $\sigma_{min}=0.002,\sigma_{max}=80,\rho=7$. 

With Guided-diffusion pretrained model \cite{dhariwal2021diffusion}, we use the uniform time steps for our method and other comparison methods, i.e.,
\begin{equation}
	\label{uniform_time_step}
	t_{i< M} = 1 + \frac{i}{M-1}(\epsilon - 1).
\end{equation}
In theory, we would need to solve the ER SDE from time $T$ to time $0$ to generate samples. However, taking inspiration from~\cite{lu2022dpm}, to circumvent numerical issues near $t = 0$, the training and evaluation of the data prediction model $\boldsymbol{x}_{\theta}\left(\boldsymbol{x}_{t}, t\right)$ typically span from time $T$ to a small positive value $\epsilon$, where $\epsilon$ is a hyperparameter \cite{song2021score}.

\subsection{Noise Schedule}
In EDM, the noise schedule is equal to the time step size schedule, i.e.,
\begin{equation}
	\label{edm_noise_schedule}
	\sigma_{i< M}=\Big[\sigma_{\max}^{\frac{1}{\rho}}+\frac{i}{M-1}(\sigma_{\min}^{\frac{1}{\rho}}-\sigma_{\max}^{\frac{1}{\rho}})\Big]^{\rho}.
\end{equation}
In Guided-diffusion, two commonly used noise schedules are provided (the linear and the cosine noise schedules). For the linear noise schedule, we have
\begin{equation}
	\label{linear_noise_schedule}
	\alpha_t = e^{-\frac{1}{4}t^2(\beta_{\max} - \beta_{\min})-\frac{1}{2}t\beta_{\min}}, 
\end{equation}
where $\beta_{\min} = 0.1$ and $\beta_{\max} = 20$, following~\cite{song2021score}. For the cosine noise schedule, we have
\begin{equation}
	\label{cosine_noise_schedule}
	\alpha_t = \frac{f(t)}{f(0)},\ \ f(t) = \cos\Big(\frac{t/T + s}{1+s}\cdot\frac{\pi}{2}\Big)^2,
\end{equation}
where $s = 0.008$, following~\cite{nichol2021improved}. Since Guided-diffusion is a VP-type pretrained model, it satisfies $\alpha_t^2 + \sigma_t^2 = 1$.

\subsection{Relationship between EDM and VE,VP}
\label{EDM_VE_VP}
EDM describes the forward process as follows:
\begin{equation}
	\boldsymbol x(t) = s(t)\boldsymbol x_0 + s(t)\sigma(t)\boldsymbol z , \ \ \  \boldsymbol z \sim \mathcal{N}(\mathbf{0},\mathbf{I}).
\end{equation}
It is not difficult to find that $\alpha_t = s(t)$ and $\sigma_t = s(t)\sigma(t)$ in VP. In addition, the expression for $s(t)$ and $\sigma(t)$ w.r.t $t$ is
\begin{equation}
	\sigma(t) = \sqrt{e^{\frac{1}{2}\beta_d t^2 + \beta_{\min}t} - 1}
\end{equation}
and
\begin{equation}
s(t) = \frac{1}{\sqrt{e^{\frac{1}{2}\beta_d t^2 + \beta_{\min}t}}}.
\end{equation}
where $\beta_d=19.9$, $\beta_{\min}=0.1$ and $t$ follows the uniform time size in Eq.(\ref{uniform_time_step}). However, in order to match the noise schedule of EDM in Eq.(\ref{edm_noise_schedule}), we let 
\begin{equation}
	\sigma(\epsilon) = \sigma_{\min},\quad \sigma(1) = \sigma_{\max}.
\end{equation}
Thus, we can get the corresponding $\beta_d$ and $\beta_{\min}$. The corresponding time step can also be obtained by using the inverse function of $\sigma(t)$.

Similarly, $s(t) = 1$ and $\sigma_t = \sigma(t)$ in VE, so we can directly use the EDM as VE-type pretrained diffusion models to match the noise schedule. Through the inverse function of $\sigma(t) = \sqrt{t}$, we can get the corresponding VE-type time step, following~\cite{karras2022elucidating}.

\subsection{Ablating Numerical Integration Points $N$}
\begin{figure}[t]
\centering
\includegraphics[width=0.5\textwidth]{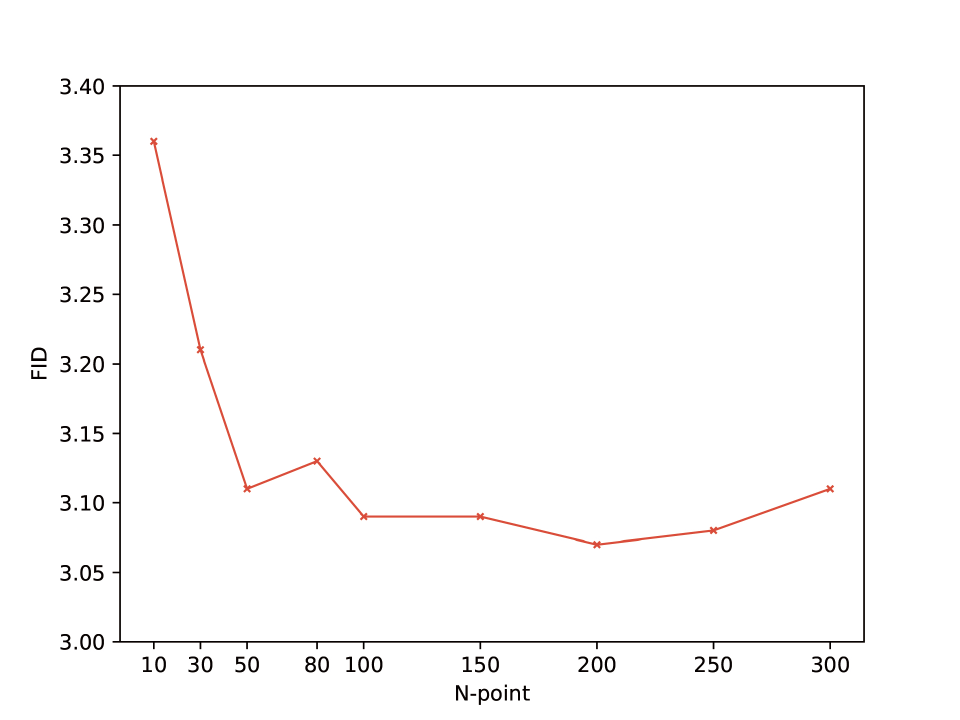}
\caption{FID (NFE=20) on CIFAR-10 with the pretrained EDM, varying with the number of integration points. As the number of integration points $N$ increases, FID scores initially show a decreasing trend, reaching a minimum at $N=200$. Subsequently, FID scores slowly increase, indicating a decrease in image generation quality.}
\label{N_points}
\end{figure}

As mentioned before, the terms $\int_{\sigma_{t_{i}}}^{\sigma_{t_{i-1}}} \frac{(\sigma - \sigma_{t_{i-1}})^{n-1}}{(n-1)!\phi(\sigma)} d \sigma$ in MT Eq.(11) and 
$\int_{\lambda_{t_{i}}}^{\lambda_{t_{i-1}}} \frac{(\lambda - \lambda_{t_{i-1}})^{n-1}}{(n-1)!\phi(\lambda)} d \lambda$ in MT Eq.(15) lack analytical expressions and need to be estimated using $N$-point numerical integration. In this section, we conduct ablation experiments to select an appropriate number of integration points. As Fig.\ref{N_points} shows, FID oscillations decrease with an increase in the number of points $N$, reaching a minimum at $N=200$. Subsequently, image quality deteriorates as the number of points increases further. Additionally, a higher number of integration points leads to slower inference speed. Therefore, to strike a balance between efficiency and performance, we choose $N=100$ for all experiments in this paper.

\subsection{Code Implementation}
Our code is available in the supplementary material, which is implemented with PyTorch. The implementation code for the pretrained model EDM is accessible at \href{https://github.com/NVlabs/edm}{https://github.com/NVlabs/edm}. The implementation code for the pretrained model Guided-diffusion is accessible at \href{https://github.com/openai/guided-diffusion}{https://github.com/openai/guided-diffusion} and the code license is MIT License.

\begin{figure*}[t]
    \begin{minipage}[t]{0.5\linewidth}
        \centering
        \includegraphics[width=1.0\textwidth]{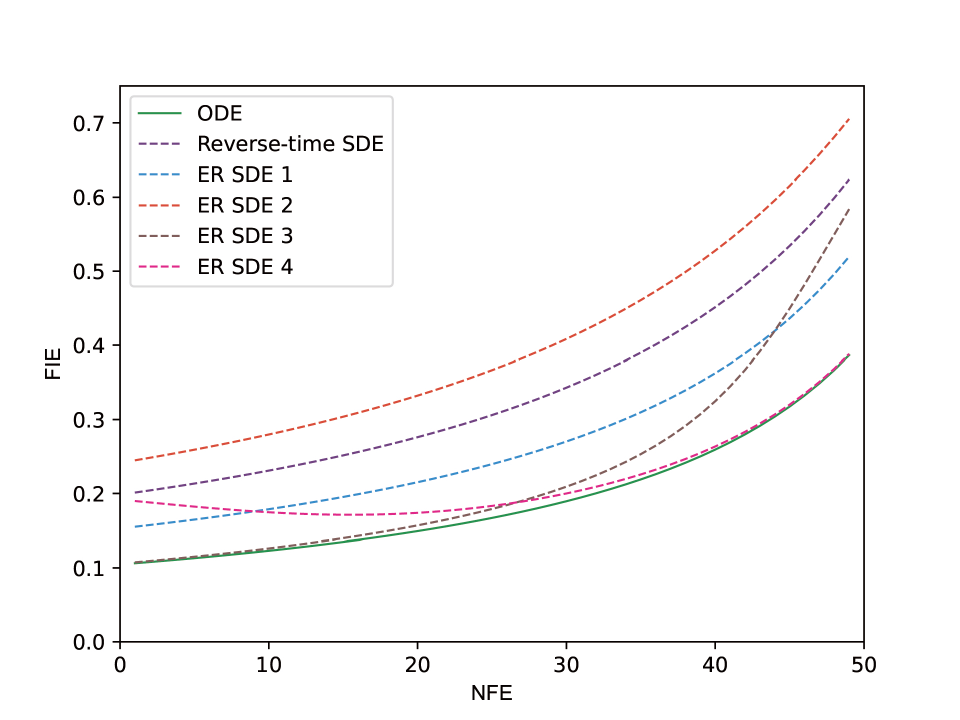}
        \centerline{(a) EDM as the pretrained model }
    \end{minipage}%
    \begin{minipage}[t]{0.5\linewidth}
        \centering
        \includegraphics[width=1.0\textwidth]{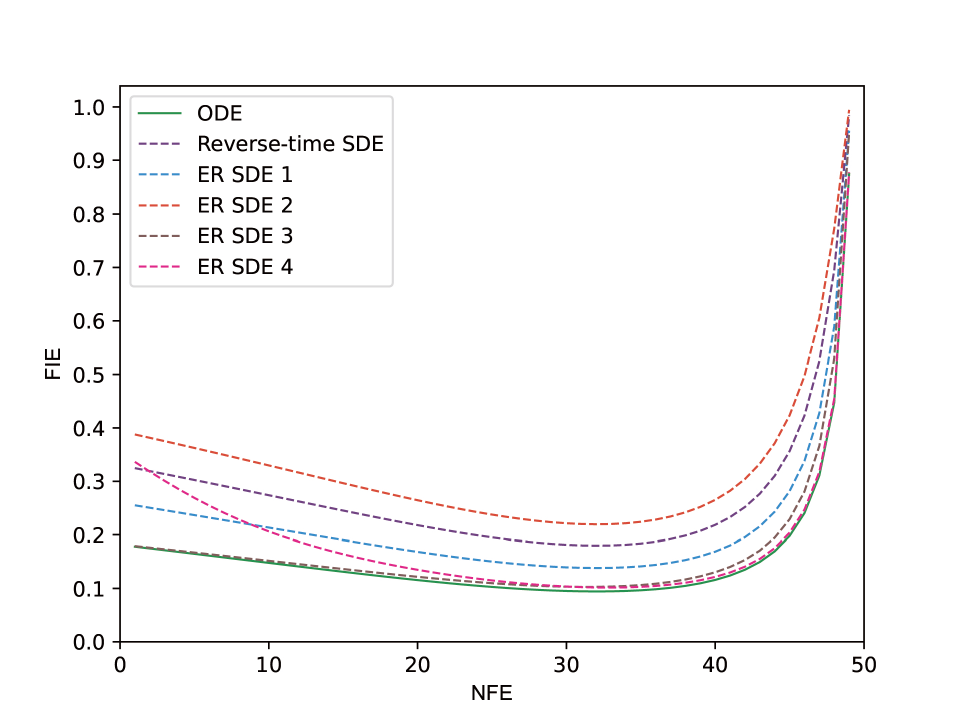}
        \centerline{(b) Guided-diffusion as the pretrained model}
    \end{minipage}
    \caption{FIE coefficients with the pretrained model EDM (a) and Guided-diffusion (b) (linear noise schedule), varying with NFE. Different step size schedules and noise schedules used in the pretrained models lead to variations in the shape of the FIE-NFE curves.}
\label{FEI_step}
\end{figure*}

\begin{table*}[h]
\centering
\caption{Sample quality measured by FID$\downarrow$ on CIFAR-10 for different stages of VE and VP ER-SDE-Solvers with the pretrained model EDM, varying the NFE. VE(P)-x denotes the x-th stage VE(P) ER-SDE-Solver.\\}
\label{vevpcifar10}
\begin{tabular}{@{}lllllllllll@{}}
\toprule
\multicolumn{2}{l}{Method\textbackslash{}NFE} & 10    & 20   & 30   & 50   & Method\textbackslash{}NFE & 10   & 20   & 30   & 50   \\ \midrule
\multicolumn{2}{l}{VE-2}                      & 10.33 & 3.33 & 2.39 & 2.03 & VE-3                      & 9.86 & 3.13 & 2.31 & 1.97 \\
\multicolumn{2}{l}{VP-2}                      & 10.12 & 3.28 & 2.41 & 2.08 & VP-3                      & 9.77 & 3.02 & 2.29 & 1.96 \\ \bottomrule
\end{tabular}
\end{table*}
\begin{table*}[t]
\centering
\caption{Sample quality measured by FID$\downarrow$ on unconditional LSUN Bedrooms $256\times256$ with the pretrained model Guided-diffusion (linear noise schedule), varying the NFE.\\}
\label{LSUN}
\begin{tabular}{@{}llllllll@{}}
\toprule
\multicolumn{3}{c}{Sampling method\textbackslash{}NFE}                                & \multicolumn{1}{l}{30} & \multicolumn{1}{l}{50} & \multicolumn{1}{l}{60} & \multicolumn{1}{l}{70} & \multicolumn{1}{l}{80} \\ \midrule
\multirow{3}{*}{\begin{tabular}[c]{@{}l@{}}Stochastic\\ Sampling\end{tabular}}  & \multicolumn{2}{l}{DDIM ($\eta=1$) \cite{song2021denoising}}              &13.76           & 8.68   &7.38   &6.51 &5.91\\
                                           & \multicolumn{2}{l}{SDE-DPM-Solver++(2M) \cite{lu2022dpm-solver}} & 4.53    & 3.31  &3.06  &2.93    &2.83     \\
                                           & \multicolumn{2}{l}{Ours (ER-SDE-Solver-3)}      & 3.55       & \bf{2.71}   & \bf{2.57}   &\bf{2.51}   &\bf{2.44}  \\ \midrule
\multirow{3}{*}{\begin{tabular}[c]{@{}l@{}}Deterministic\\ Sampling\end{tabular}} & \multicolumn{2}{l}{DDIM \cite{song2021denoising}}   &4.77  & 3.67 &3.31 &3.21     &3.09     \\
                                                                                  & \multicolumn{2}{l}{DPM-Solver-3 \cite{lu2022dpm}}  & \bf{3.45} & 2.71  &2.68 &2.63 &2.57\\
                                                                                  & \multicolumn{2}{l}{DPM-Solver++(2M) \cite{lu2022dpm-solver}}   & 4.02 & 3.15  &2.95 &2.88 &2.80 \\  \bottomrule
\end{tabular}
\end{table*}

\begin{table*}[h]
  \centering
  \scriptsize
\caption{Sample quality measured by FID$\downarrow$ on class-conditional ImageNet $256\times256$ with the pretrained Guided-diffusion (with classifier guidance scale, linear noise schedule), varying the NFE.}
\label{image256_g1}
\vskip 0.15in
\scalebox{1.2}{
      \begin{tabular}{@{}lcccccccccccccccc@{}}
        \toprule
        \multirow{2}{*}{\textbf{Sampling method\textbackslash{}NFE}} & \multicolumn{4}{c}{\textbf{classifier guidance scale = 1.0}} & \phantom{c} & \multicolumn{4}{c}{\textbf{classifier guidance scale = 2.0}} \\ \cmidrule{2-5} \cmidrule{7-10}
    &\bf{10} & \bf{20} & \bf{30} & \bf{50} & \phantom{c} & \bf{10} & \bf{20} & \bf{30} & \bf{50} \\ \midrule
        \textbf{Stochastic Sampling}      \\ \midrule
        DDIM($\eta=1$) \cite{song2021denoising}         &22.60  & 11.44   &8.62   &6.69  & \phantom{c} & 17.97 & 10.23   & 8.19   & 6.85\\
        SDE-DPM-Solver++(2M) \cite{lu2022dpm-solver}    &11.46   & 6.80   &5.94  &5.28  & \phantom{c} & 9.21   & 6.01   &5.47  &5.19 \\
        Ours(ER-SDE-Solver-3)  & \bf{8.17}  & \bf{5.62}     &\bf{5.24}  &\bf{5.07}  & \phantom{c}& \bf{6.24}  & \bf{4.76}     &\bf{4.62}  &\bf{4.57} \\
        \\[-2ex]
        \textbf{Deterministic Sampling}      \\ \midrule
        DDIM \cite{song2021denoising}   & 11.93   &7.37    &6.39    &5.93  & \phantom{c}  & 8.63   &5.60    &5.00    &4.59   \\
        DPM-Solver-3 \cite{lu2022dpm}   &9.00    &6.86    &6.62   & 6.05  & \phantom{c}  &6.45    &5.03    &4.94   & 4.92    \\
        DPM-Solver++(2M) \cite{lu2022dpm-solver}   &9.65  &7.73  &7.26  &6.90  & \phantom{c}  &7.19  & 5.54 & 5.32 & 5.16 \\ \bottomrule      
\end{tabular}}
\vspace{-3mm}
\end{table*}

\section{Experiment Details}
\label{appendixd}
We list all the experimental details and experimental results in this section.

Firstly, we have observed that the shape of the FIE-NFE curve is closely related to the choice of step size schedule and noise schedule within the pretrained diffusion models. Fig.\ref{FEI_step} illustrates two distinct FIE-NFE curves using EDM \cite{karras2022elucidating} and Guided-diffusion \cite{dhariwal2021diffusion} as pretrained models.

Table \ref{vevpcifar10} shows how FID scores change with NFE for different stages of VE ER-SDE-Solvers and VP ER-SDE-Solvers on CIFAR-10. Consistent with the findings in Main Text Table 4, VE ER-SDE-Solvers and VP ER-SDE-Solvers demonstrate similar image generation quality. Furthermore, as the stage increases, convergence becomes faster with the decreasing discretization errors.

With EDM pretrained model used in Main Text Table 2,4 and Fig.1,3, we evaluate all the methods using the same pretrained checkpoint provided by~\cite{karras2022elucidating}. For a fair comparison, all the methods in our evaluation follow the noise schedule in Eq.(\ref{edm_noise_schedule}). Although many techniques, such as \textit{thresholding methods} \cite{lu2022dpm-solver} and \textit{numerical clip alpha} \cite{dhariwal2021diffusion, nichol2021improved,lu2022dpm}, can help reduce FID scores, they are challenging to apply to all sampling methods equally. For example, \textit{thresholding methods} are usually used in the data prediction model or where it comes to prediction data, but they cannot directly be used in the noise prediction model. Therefore, for the purpose of fair comparison and evaluating the true efficacy of these sampling methods, none of the FID scores we report use these techniques. That is why they may appear slightly higher than the FID scores in the original paper. For experiments involving the ImageNet $64\times64$ dataset, we use the checkpoint \href{https://nvlabs-fi-cdn.nvidia.com/edm/pretrained/}{edm-imagenet-64x64-cond-adm.pkl}. For experiments involving the CIFAR-10 dataset, we use the checkpoint \href{https://nvlabs-fi-cdn.nvidia.com/edm/pretrained/}{edm-cifar10-32x32-cond-ve.pkl}. For the tow categories of experiments mentioned above, we calculate FID scores using random seeds 100000-149999, following ~\cite{karras2022elucidating}. For comparison method EDM-Stochastic, we use the optimal settings mentioned in~\cite{karras2022elucidating}, which are $S_{churn}=40,S_{min}=0.05, S_{max}=50,S_{noise}=1.003$ in Main Text Table 2.

Next, we apply ER-SDE-Solvers to high-resolution image generation. Table \ref{LSUN} provides comparative results on LSUN $256\times256$ \cite{yu2015lsun}, using Guided-diffusion \cite{dhariwal2021diffusion} as the pretrained model. The results demonstrate that ER-SDE-Solvers can also accelerate the generation of high-resolution images.

Main Text Table 3 demonstrates that ER-SDE-Solvers with classifier guidance exhibit high image generation quality even with very low NFE. We explore the impact of the classifier guidance scale on the efficiency of high-quality sampling, as illustrated in Table \ref{image256_g1}. The results indicate that a higher classifier guidance scale allows for the generation of higher-quality images with fewer NFE. The introduction of the classifier guidance scale further accentuates the efficient high-quality sampling capability of ER-SDE-Solvers.

With Guided-diffusion pretrained model used in Main Text Table 1, 3, and Table \ref{LSUN}, \ref{image256_g1}, we evaluate all the methods using the same pretrained checkpoint provided by~\cite{dhariwal2021diffusion}. Similarly, for the purpose of fair comparison and evaluating the true efficacy of these sampling methods, none of the FID scores we report use techniques like \textit{thresholding methods} or \textit{numerical clip alpha}. That is why they may appear slightly higher than the FID scores in the original paper. For experiments involving the ImageNet $128\times128$ dataset, we use the checkpoint \href{https://github.com/openai/guided-diffusion}{128x128\_diffusion.pt}. All the methods in our evaluation follow the linear schedule and uniform time steps in Eq.(\ref{uniform_time_step}). Although DPM-Solver \cite{lu2022dpm} introduces some types of discrete time steps, we do not use them in our experiment and just follow the initial settings in Guided-diffusion. We do not use the classifier guidance but evaluate all methods on class-conditional. For experiments involving the LSUN $256\times256$ dataset, we use the checkpoint \href{https://github.com/openai/guided-diffusion}{lsun\_bedroom.pt}. All the methods in our evaluation follow the linear schedule and uniform time steps in Eq.(\ref{uniform_time_step}). For experiments involving the ImageNet $256\times256$ dataset, we use the diffusion checkpoint \href{https://github.com/openai/guided-diffusion}{256x256\_diffusion.pt} and the classifier checkpoint \href{https://github.com/openai/guided-diffusion}{256x256\_classifier.pt}. All the methods in our evaluation follow the linear schedule and uniform time steps in Eq.(\ref{uniform_time_step}).

\begin{figure*}[htbp]
	\hspace{0.15\linewidth}
	\begin{minipage}{0.2\linewidth}
	\centering
	Ours\\
\centering
(ER-SDE-Solver-3)
	\end{minipage}
	\begin{minipage}{0.2\linewidth}
	\centering
	DDIM($\eta=1$)\\
\centering
~\cite{song2021denoising}
	\end{minipage}
	\begin{minipage}{0.2\linewidth}
	\centering
	DDIM\\
\centering
~\cite{song2021denoising}
	\end{minipage}
	\begin{minipage}{0.2\linewidth}
	\centering
	DPM-Solver++(2M)\\
\centering
~\cite{lu2022dpm-solver}
	\end{minipage}
	\vspace{0.2cm}
	\\
	\centering
	\begin{minipage}{0.15\linewidth}
	\centering
    NFE=10 
	\end{minipage}
	\begin{minipage}{0.2\linewidth}
		\centering
			\includegraphics[width=\linewidth]{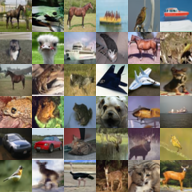}\\
	\end{minipage}
	\begin{minipage}{0.2\linewidth}
		\centering
			\includegraphics[width=\linewidth]{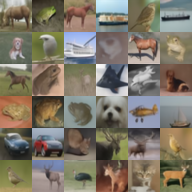}\\
	\end{minipage}
	\begin{minipage}{0.2\linewidth}
		\centering
			\includegraphics[width=\linewidth]{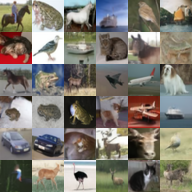}\\
	\end{minipage}
	\begin{minipage}{0.2\linewidth}
		\centering
			\includegraphics[width=\linewidth]{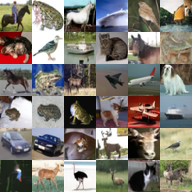}\\
	\end{minipage}
     \vspace{0.1cm}
     \\
	\centering
	\begin{minipage}{0.15\linewidth}
	\centering
    NFE=20 
	\end{minipage}
	\begin{minipage}{0.2\linewidth}
		\centering
			\includegraphics[width=\linewidth]{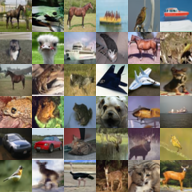}\\
	\end{minipage}
\begin{minipage}{0.2\linewidth}
		\centering
			\includegraphics[width=\linewidth]{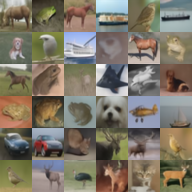}\\
	\end{minipage}
	\begin{minipage}{0.2\linewidth}
		\centering
			\includegraphics[width=\linewidth]{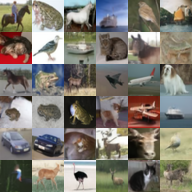}\\
	\end{minipage}
	\begin{minipage}{0.2\linewidth}
		\centering
			\includegraphics[width=\linewidth]{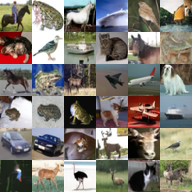}\\
	\end{minipage}
     \vspace{0.1cm}
     \\
	\centering
	\begin{minipage}{0.15\linewidth}
	\centering
    NFE=30 
	\end{minipage}
	\begin{minipage}{0.2\linewidth}
		\centering
			\includegraphics[width=\linewidth]{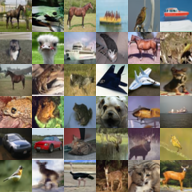}\\
	\end{minipage}
\begin{minipage}{0.2\linewidth}
		\centering
			\includegraphics[width=\linewidth]{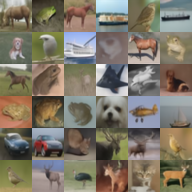}\\
	\end{minipage}
	\begin{minipage}{0.2\linewidth}
		\centering
			\includegraphics[width=\linewidth]{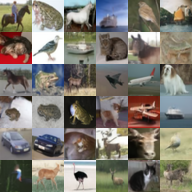}\\
	\end{minipage}
	\begin{minipage}{0.2\linewidth}
		\centering
			\includegraphics[width=\linewidth]{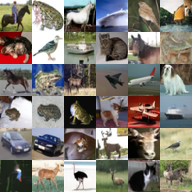}\\
	\end{minipage}
     \vspace{0.1cm}
     \\
	\centering
	\begin{minipage}{0.15\linewidth}
	\centering
    NFE=40 
	\end{minipage}
	\begin{minipage}{0.2\linewidth}
		\centering
			\includegraphics[width=\linewidth]{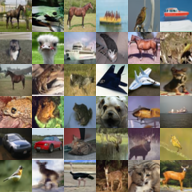}\\
	\end{minipage}
\begin{minipage}{0.2\linewidth}
		\centering
			\includegraphics[width=\linewidth]{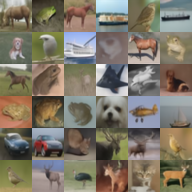}\\
	\end{minipage}
	\begin{minipage}{0.2\linewidth}
		\centering
			\includegraphics[width=\linewidth]{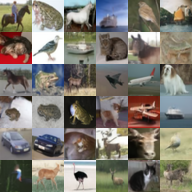}\\
	\end{minipage}
	\begin{minipage}{0.2\linewidth}
		\centering
			\includegraphics[width=\linewidth]{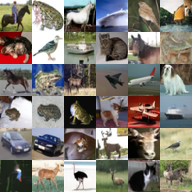}\\
	\end{minipage}
     \vspace{0.1cm}
     \\
	\centering
	\begin{minipage}{0.15\linewidth}
	\centering
    NFE=50 
	\end{minipage}
	\begin{minipage}{0.2\linewidth}
		\centering
			\includegraphics[width=\linewidth]{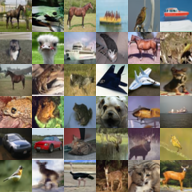}\\
	\end{minipage}
\begin{minipage}{0.2\linewidth}
		\centering
			\includegraphics[width=\linewidth]{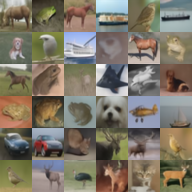}\\
	\end{minipage}
	\begin{minipage}{0.2\linewidth}
		\centering
			\includegraphics[width=\linewidth]{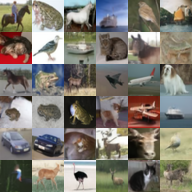}\\
	\end{minipage}
	\begin{minipage}{0.2\linewidth}
		\centering
			\includegraphics[width=\linewidth]{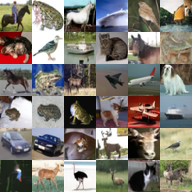}\\
	\end{minipage}
\caption{\label{fig:cifar10}
Samples by stochastic samplers (DDIM($\eta=1$), ER-SDE-Solver-3 (ours)) and deterministic samplers (DDIM, DPM-Solver++(2M)) with 10, 20, 30, 40, 50  number of function evaluations (NFE) with the same random seed (666), using the pretrained EDM \cite{karras2022elucidating} on CIFAR-10 $32\times32$.
}
\end{figure*}

\begin{figure*}[htbp]
	\hspace{0.15\linewidth}
	\begin{minipage}{0.2\linewidth}
	\centering
	Ours\\
\centering
(ER-SDE-Solver-3)
	\end{minipage}
	\begin{minipage}{0.2\linewidth}
	\centering
	DDIM($\eta=1$)\\
\centering
~\cite{song2021denoising}
	\end{minipage}
	\begin{minipage}{0.2\linewidth}
	\centering
	DDIM\\
\centering
~\cite{song2021denoising}
	\end{minipage}
	\begin{minipage}{0.2\linewidth}
	\centering
	DPM-Solver++(2M)\\
\centering
~\cite{lu2022dpm-solver}
	\end{minipage}
	\vspace{0.2cm}
	\\
	\centering
	\begin{minipage}{0.15\linewidth}
	\centering
    NFE=10 
	\end{minipage}
	\begin{minipage}{0.2\linewidth}
		\centering
			\includegraphics[width=\linewidth]{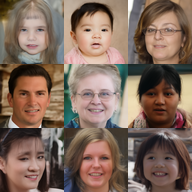}\\
	\end{minipage}
	\begin{minipage}{0.2\linewidth}
		\centering
			\includegraphics[width=\linewidth]{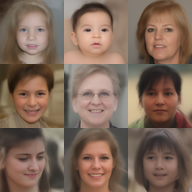}\\
	\end{minipage}
	\begin{minipage}{0.2\linewidth}
		\centering
			\includegraphics[width=\linewidth]{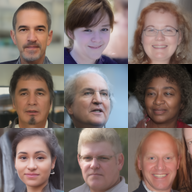}\\
	\end{minipage}
	\begin{minipage}{0.2\linewidth}
		\centering
			\includegraphics[width=\linewidth]{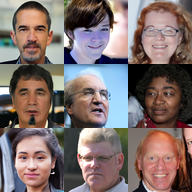}\\
	\end{minipage}
     \vspace{0.1cm}
     \\
	\centering
	\begin{minipage}{0.15\linewidth}
	\centering
    NFE=20 
	\end{minipage}
	\begin{minipage}{0.2\linewidth}
		\centering
			\includegraphics[width=\linewidth]{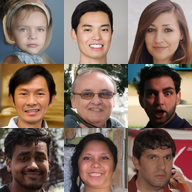}\\
	\end{minipage}
\begin{minipage}{0.2\linewidth}
		\centering
			\includegraphics[width=\linewidth]{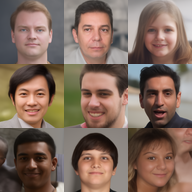}\\
	\end{minipage}
	\begin{minipage}{0.2\linewidth}
		\centering
			\includegraphics[width=\linewidth]{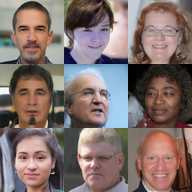}\\
	\end{minipage}
	\begin{minipage}{0.2\linewidth}
		\centering
			\includegraphics[width=\linewidth]{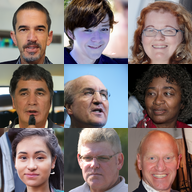}\\
	\end{minipage}
     \vspace{0.1cm}
     \\
	\centering
	\begin{minipage}{0.15\linewidth}
	\centering
    NFE=30 
	\end{minipage}
	\begin{minipage}{0.2\linewidth}
		\centering
			\includegraphics[width=\linewidth]{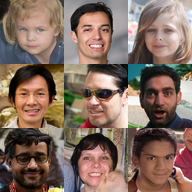}\\
	\end{minipage}
\begin{minipage}{0.2\linewidth}
		\centering
			\includegraphics[width=\linewidth]{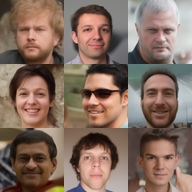}\\
	\end{minipage}
	\begin{minipage}{0.2\linewidth}
		\centering
			\includegraphics[width=\linewidth]{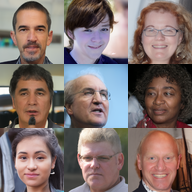}\\
	\end{minipage}
	\begin{minipage}{0.2\linewidth}
		\centering
			\includegraphics[width=\linewidth]{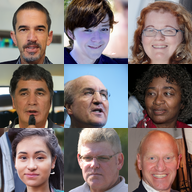}\\
	\end{minipage}
     \vspace{0.1cm}
     \\
	\centering
	\begin{minipage}{0.15\linewidth}
	\centering
    NFE=40 
	\end{minipage}
	\begin{minipage}{0.2\linewidth}
		\centering
			\includegraphics[width=\linewidth]{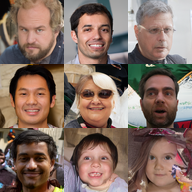}\\
	\end{minipage}
\begin{minipage}{0.2\linewidth}
		\centering
			\includegraphics[width=\linewidth]{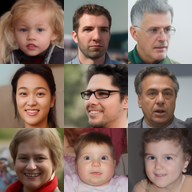}\\
	\end{minipage}
	\begin{minipage}{0.2\linewidth}
		\centering
			\includegraphics[width=\linewidth]{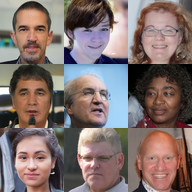}\\
	\end{minipage}
	\begin{minipage}{0.2\linewidth}
		\centering
			\includegraphics[width=\linewidth]{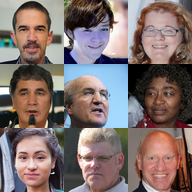}\\
	\end{minipage}
     \vspace{0.1cm}
     \\
	\centering
	\begin{minipage}{0.15\linewidth}
	\centering
    NFE=50 
	\end{minipage}
	\begin{minipage}{0.2\linewidth}
		\centering
			\includegraphics[width=\linewidth]{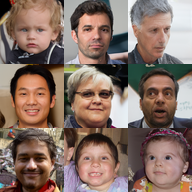}\\
	\end{minipage}
\begin{minipage}{0.2\linewidth}
		\centering
			\includegraphics[width=\linewidth]{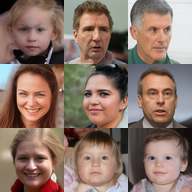}\\
	\end{minipage}
	\begin{minipage}{0.2\linewidth}
		\centering
			\includegraphics[width=\linewidth]{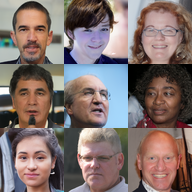}\\
	\end{minipage}
	\begin{minipage}{0.2\linewidth}
		\centering
			\includegraphics[width=\linewidth]{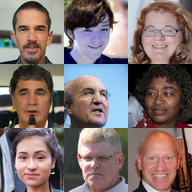}\\
	\end{minipage}
\caption{\label{fig:ffhq64}
Samples by stochastic samplers (DDIM($\eta=1$), ER-SDE-Solver-3 (ours)) and deterministic samplers (DDIM, DPM-Solver++(2M)) with 10, 20, 30, 40, 50  number of function evaluations (NFE) with the same random seed (666), using the pretrained EDM \cite{karras2022elucidating} on FFHQ $64\times64$.
}
\end{figure*}

\begin{figure*}[htbp]
	\hspace{0.08\linewidth}
	\begin{minipage}{0.45\linewidth}
	\centering
	\quad \quad \quad\quad DPM-Solver-3~\cite{lu2022dpm}
	\end{minipage}
	\begin{minipage}{0.45\linewidth}
	\centering
	Ours (ER-SDE-Solver-3)
	\end{minipage}
	\vspace{0.2cm}
	\\
	\centering
	\begin{minipage}{0.15\linewidth}
	\centering
    NFE=10 
	\end{minipage}
	\begin{minipage}{0.4\linewidth}
		\centering
			\includegraphics[width=\linewidth]{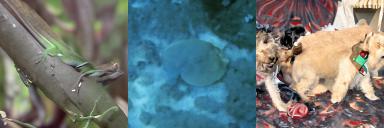}\\
	\end{minipage}
	\begin{minipage}{0.4\linewidth}
		\centering
			\includegraphics[width=\linewidth]{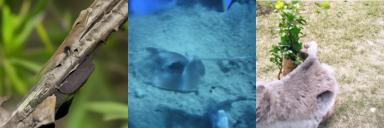}\\
	\end{minipage}
     \vspace{0.1cm}
     \\
	\centering
	\begin{minipage}{0.15\linewidth}
	\centering
    NFE=20 
	\end{minipage}
	\begin{minipage}{0.4\linewidth}
		\centering
			\includegraphics[width=\linewidth]{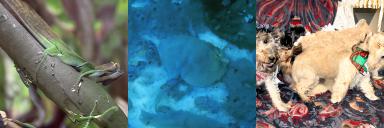}\\
	\end{minipage}
	\begin{minipage}{0.4\linewidth}
		\centering
			\includegraphics[width=\linewidth]{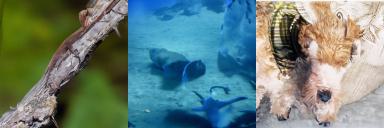}\\
	\end{minipage}
     \vspace{0.1cm}
     \\
	\centering
	\begin{minipage}{0.15\linewidth}
	\centering
    NFE=30 
	\end{minipage}
	\begin{minipage}{0.4\linewidth}
		\centering
			\includegraphics[width=\linewidth]{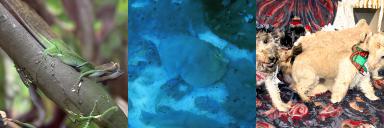}\\
	\end{minipage}
	\begin{minipage}{0.4\linewidth}
		\centering
			\includegraphics[width=\linewidth]{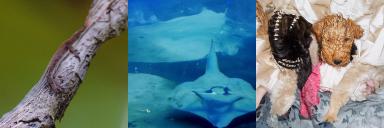}\\
	\end{minipage}
     \vspace{0.1cm}
     \\
	\centering
	\begin{minipage}{0.15\linewidth}
	\centering
    NFE=40 
	\end{minipage}
	\begin{minipage}{0.4\linewidth}
		\centering
			\includegraphics[width=\linewidth]{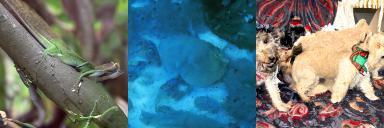}\\
	\end{minipage}
	\begin{minipage}{0.4\linewidth}
		\centering
			\includegraphics[width=\linewidth]{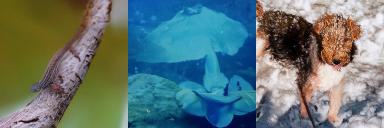}\\
	\end{minipage}
     \vspace{0.1cm}
     \\
	\centering
	\begin{minipage}{0.15\linewidth}
	\centering
    NFE=50 
	\end{minipage}
	\begin{minipage}{0.4\linewidth}
		\centering
			\includegraphics[width=\linewidth]{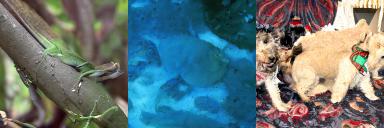}\\
	\end{minipage}
	\begin{minipage}{0.4\linewidth}
		\centering
			\includegraphics[width=\linewidth]{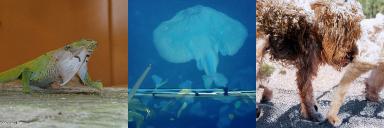}\\
	\end{minipage}
\caption{\label{fig:img128}
Samples by stochastic sampler (ER-SDE-Solver-3 (ours)) and deterministic sampler (DPM-Solver-3) with 10, 20, 30, 40, 50  number of function evaluations (NFE) with the same random seed (999), using the pretrained Guided-diffusion \cite{dhariwal2021diffusion} on ImageNet $128\times128$ without classifier guidance.
}
\end{figure*}

\begin{figure*}[htbp]
	\hspace{0.08\linewidth}
	\begin{minipage}{0.45\linewidth}
	\centering
	\quad \quad \quad\quad DPM-Solver-3~\cite{lu2022dpm}
	\end{minipage}
	\begin{minipage}{0.45\linewidth}
	\centering
	Ours (ER-SDE-Solver-3)
	\end{minipage}
	\vspace{0.2cm}
	\\
	\centering
	\begin{minipage}{0.15\linewidth}
	\centering
    NFE=10 
	\end{minipage}
	\begin{minipage}{0.4\linewidth}
		\centering
			\includegraphics[width=\linewidth]{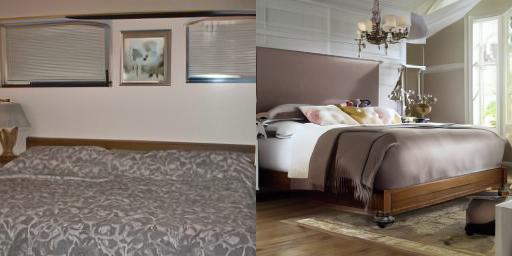}\\
	\end{minipage}
	\begin{minipage}{0.4\linewidth}
		\centering
			\includegraphics[width=\linewidth]{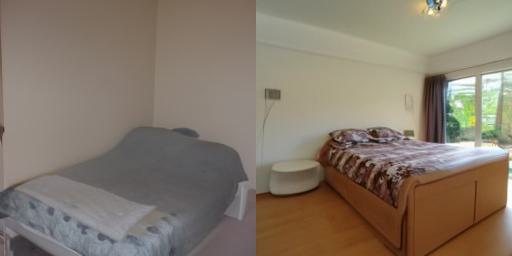}\\
	\end{minipage}
     \vspace{0.1cm}
     \\
	\centering
	\begin{minipage}{0.15\linewidth}
	\centering
    NFE=20 
	\end{minipage}
	\begin{minipage}{0.4\linewidth}
		\centering
			\includegraphics[width=\linewidth]{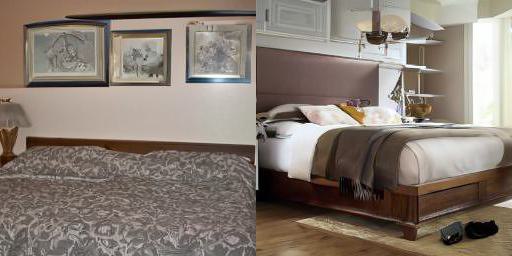}\\
	\end{minipage}
	\begin{minipage}{0.4\linewidth}
		\centering
			\includegraphics[width=\linewidth]{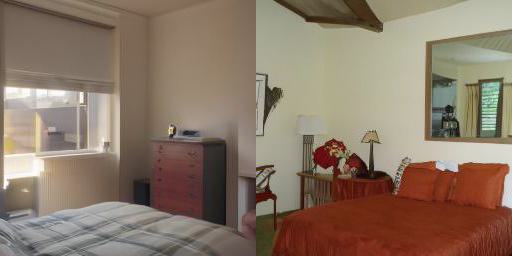}\\
	\end{minipage}
     \vspace{0.1cm}
     \\
	\centering
	\begin{minipage}{0.15\linewidth}
	\centering
    NFE=30 
	\end{minipage}
	\begin{minipage}{0.4\linewidth}
		\centering
			\includegraphics[width=\linewidth]{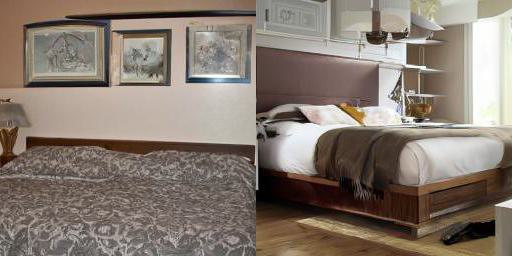}\\
	\end{minipage}
	\begin{minipage}{0.4\linewidth}
		\centering
			\includegraphics[width=\linewidth]{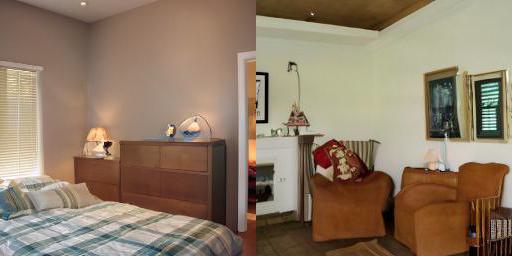}\\
	\end{minipage}
     \vspace{0.1cm}
     \\
	\centering
	\begin{minipage}{0.15\linewidth}
	\centering
    NFE=40 
	\end{minipage}
	\begin{minipage}{0.4\linewidth}
		\centering
			\includegraphics[width=\linewidth]{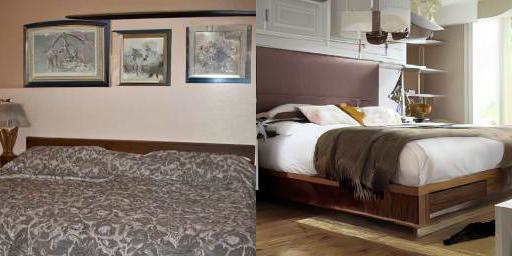}\\
	\end{minipage}
	\begin{minipage}{0.4\linewidth}
		\centering
			\includegraphics[width=\linewidth]{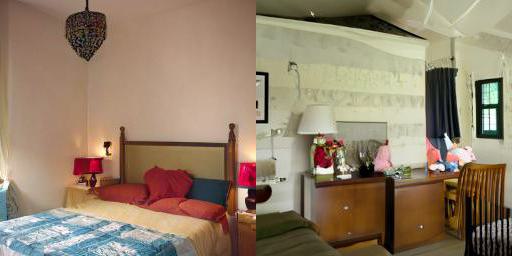}\\
	\end{minipage}
     \vspace{0.1cm}
     \\
	\centering
	\begin{minipage}{0.15\linewidth}
	\centering
    NFE=50 
	\end{minipage}
	\begin{minipage}{0.4\linewidth}
		\centering
			\includegraphics[width=\linewidth]{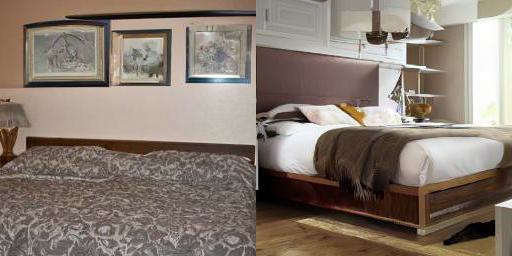}\\
	\end{minipage}
	\begin{minipage}{0.4\linewidth}
		\centering
			\includegraphics[width=\linewidth]{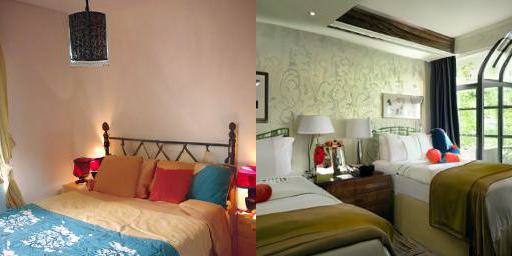}\\
	\end{minipage}
\caption{\label{fig:lsun256}
Samples by stochastic sampler (ER-SDE-Solver-3 (ours)) and deterministic sampler (DPM-Solver-3) with 10, 20, 30, 40, 50  number of function evaluations (NFE) with the same random seed (666), using the pretrained Guided-diffusion \cite{dhariwal2021diffusion} on LSUN Bedrooms $256\times256$.
}
\end{figure*}

\begin{figure*}[htbp]
	\hspace{0.08\linewidth}
	\begin{minipage}{0.45\linewidth}
	\centering
	\quad \quad \quad\quad DPM-Solver-3~\cite{lu2022dpm}
	\end{minipage}
	\begin{minipage}{0.45\linewidth}
	\centering
	Ours (ER-SDE-Solver-3)
	\end{minipage}
	\vspace{0.2cm}
	\\
	\centering
	\begin{minipage}{0.15\linewidth}
	\centering
    NFE=10 
	\end{minipage}
	\begin{minipage}{0.4\linewidth}
		\centering
			\includegraphics[width=\linewidth]{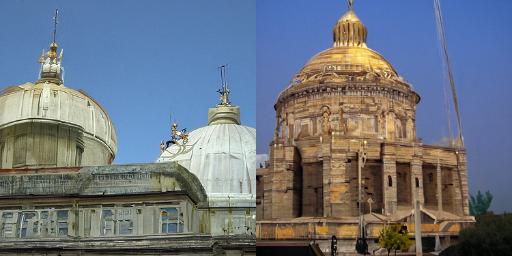}\\
	\end{minipage}
	\begin{minipage}{0.4\linewidth}
		\centering
			\includegraphics[width=\linewidth]{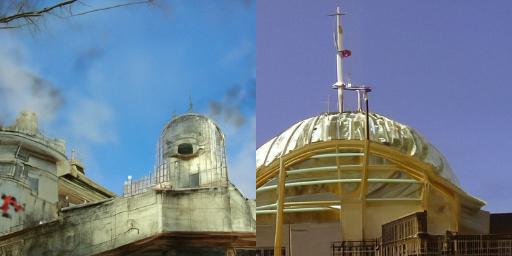}\\
	\end{minipage}
     \vspace{0.1cm}
     \\
	\centering
	\begin{minipage}{0.15\linewidth}
	\centering
    NFE=20 
	\end{minipage}
	\begin{minipage}{0.4\linewidth}
		\centering
			\includegraphics[width=\linewidth]{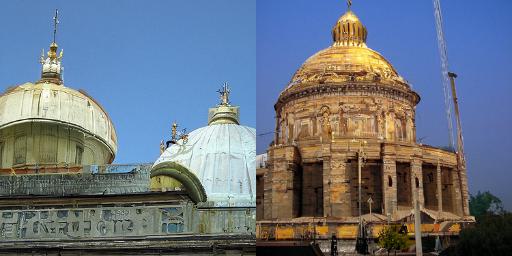}\\
	\end{minipage}
	\begin{minipage}{0.4\linewidth}
		\centering
			\includegraphics[width=\linewidth]{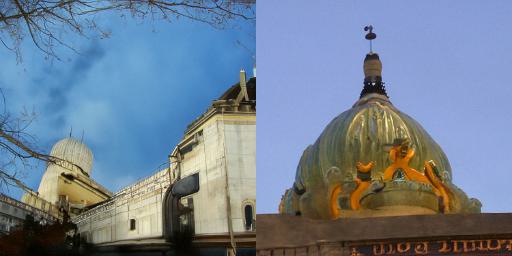}\\
	\end{minipage}
     \vspace{0.1cm}
     \\
	\centering
	\begin{minipage}{0.15\linewidth}
	\centering
    NFE=30 
	\end{minipage}
	\begin{minipage}{0.4\linewidth}
		\centering
			\includegraphics[width=\linewidth]{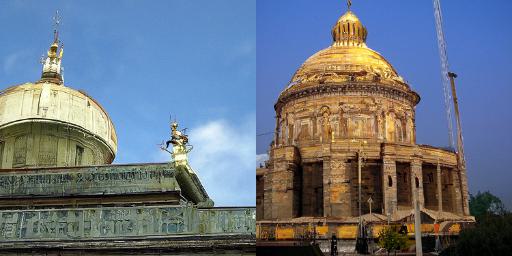}\\
	\end{minipage}
	\begin{minipage}{0.4\linewidth}
		\centering
			\includegraphics[width=\linewidth]{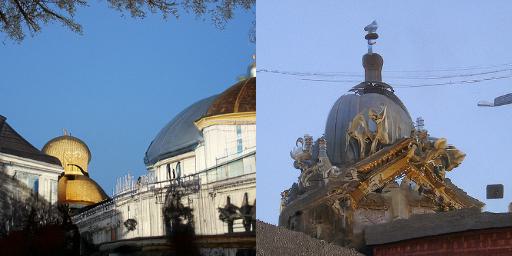}\\
	\end{minipage}
     \vspace{0.1cm}
     \\
	\centering
	\begin{minipage}{0.15\linewidth}
	\centering
    NFE=40 
	\end{minipage}
	\begin{minipage}{0.4\linewidth}
		\centering
			\includegraphics[width=\linewidth]{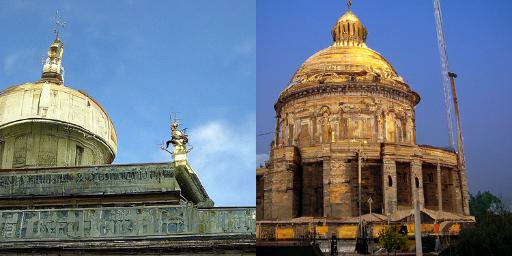}\\
	\end{minipage}
	\begin{minipage}{0.4\linewidth}
		\centering
			\includegraphics[width=\linewidth]{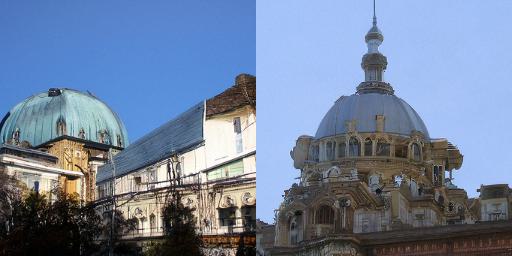}\\
	\end{minipage}
     \vspace{0.1cm}
     \\
	\centering
	\begin{minipage}{0.15\linewidth}
	\centering
    NFE=50 
	\end{minipage}
	\begin{minipage}{0.4\linewidth}
		\centering
			\includegraphics[width=\linewidth]{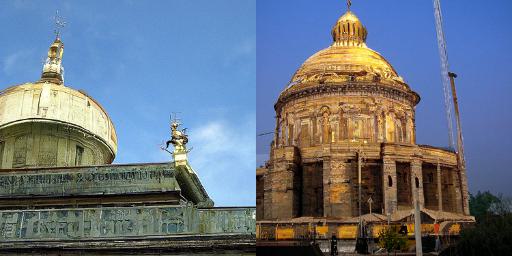}\\
	\end{minipage}
	\begin{minipage}{0.4\linewidth}
		\centering
			\includegraphics[width=\linewidth]{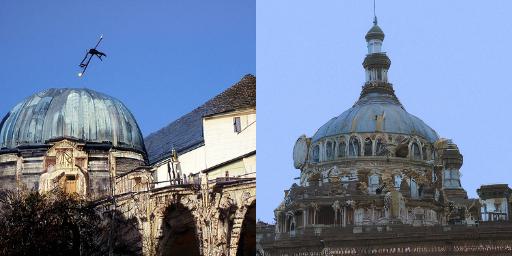}\\
	\end{minipage}
\caption{\label{fig:img256}
Samples by stochastic sampler (ER-SDE-Solver-3 (ours)) and deterministic sampler (DPM-Solver-3) with 10, 20, 30, 40, 50  number of function evaluations (NFE) with the same random seed (999), using the pretrained Guided-diffusion \cite{dhariwal2021diffusion} on ImageNet $256\times256$. The class is fixed as dome and classifier guidance scale is 2.0.
}
\end{figure*}

When the random seed is fixed, Fig.\ref{fig:cifar10} -  Fig.\ref{fig:img256} compare the sampling results between stochastic samplers and deterministic samplers on CIFAR-10 $32\times 32$ \cite{krizhevsky2009learning}, FFHQ $64\times 64$ \cite{karras2019style}, ImageNet $128\times 128$, ImageNet $256\times 256$ \cite{deng2009imagenet} and LSUN $256\times 256$~\cite{yu2015lsun} datasets. As stochastic samplers introduce stochastic noise at each step of the sampling process, the generated images exhibit greater variability, which becomes more pronounced in higher-resolution images. For lower-resolution images, such as CIFAR-10 $32\times 32$, stochastic samplers may not introduce significant variations within a limited number of steps. However, this does not diminish the value of stochastic samplers, as low-resolution images are becoming less common with the advancements in imaging and display technologies. Furthermore, we also observe that stochastic samplers and deterministic samplers diverge towards different trajectories early in the sampling process, while samplers belonging to the same category exhibit similar patterns of change. Further exploration of stochastic samplers and deterministic samplers is left for future work.

\begin{figure*}[htbp]
\centering
\includegraphics[width=0.85\textwidth]{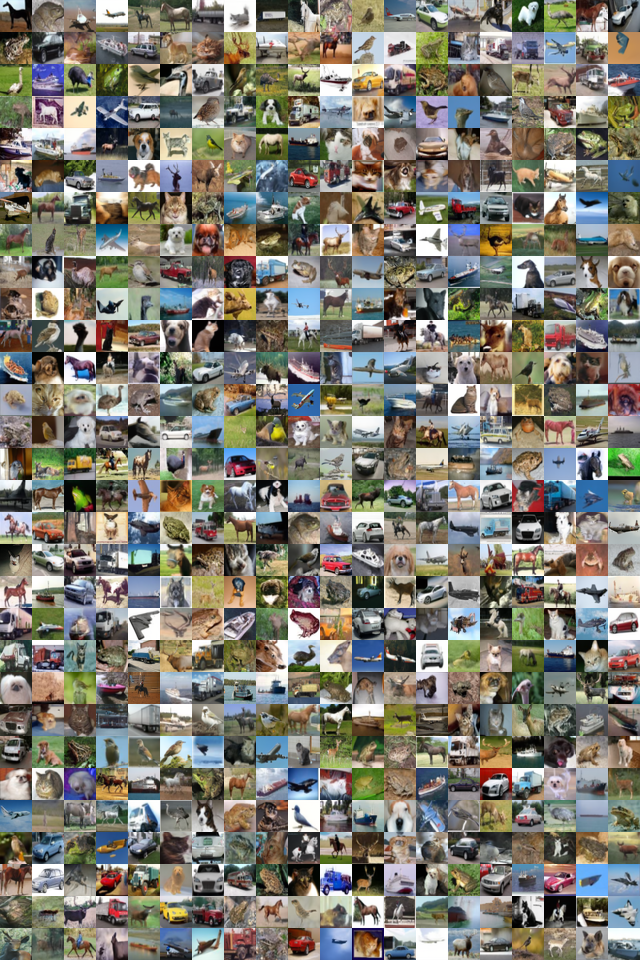}
\caption{Generated images with ER-SDE-Slover-3 (ours) on CIFAR-10 (NFE=20). The pretrained model is EDM \cite{karras2022elucidating}.}
\label{our_cifar}
\end{figure*}

\begin{figure*}[htbp]
\centering
\includegraphics[width=0.85\textwidth]{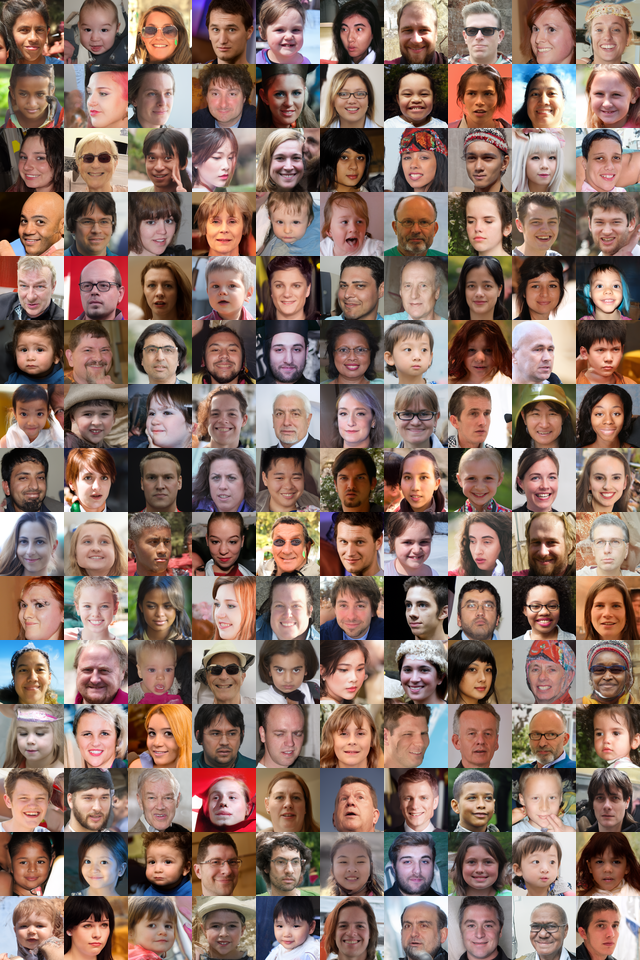}
\caption{Generated images with ER-SDE-Slover-3 (ours) on FFHQ $64\times64$ (NFE=20). The pretrained model is EDM \cite{karras2022elucidating}.}
\label{our_ffhq}
\end{figure*}

\begin{figure*}[htbp]
\centering
\includegraphics[width=0.85\textwidth]{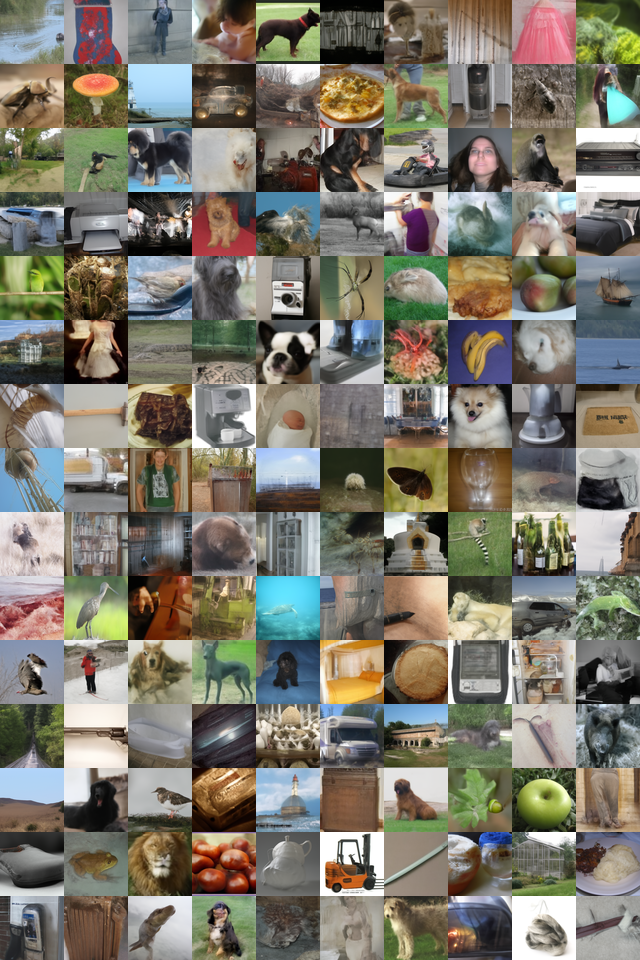}
\caption{Generated images with ER-SDE-Slover-3 (ours) on ImageNet $64\times64$ (NFE=20). The pretrained model is EDM \cite{karras2022elucidating}.}
\label{our_img64}
\end{figure*}

\begin{figure*}[htbp]
\centering
\includegraphics[width=0.8\textwidth]{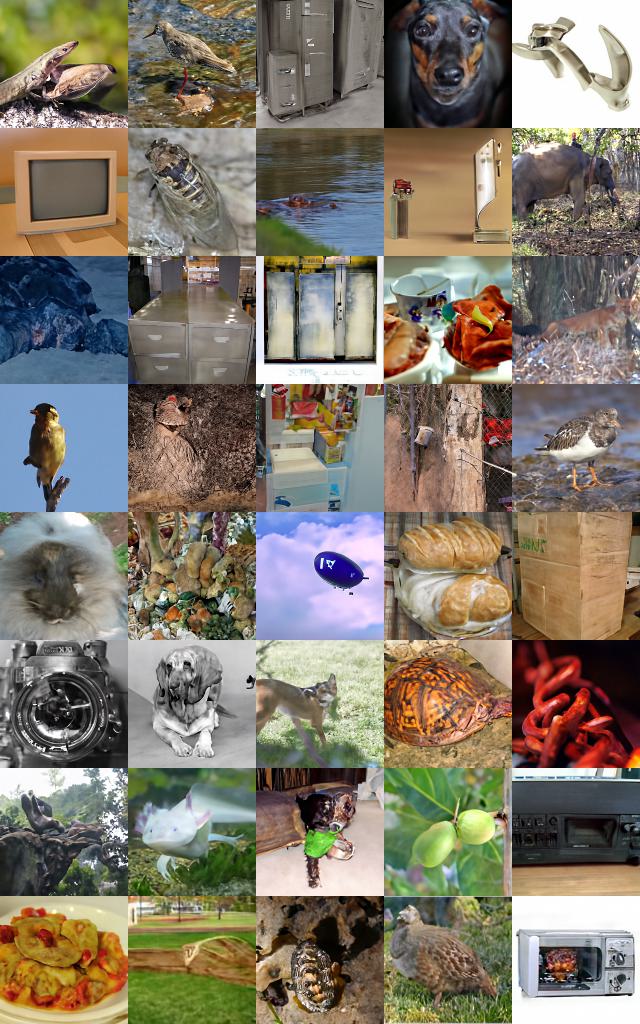}
\caption{Generated images with ER-SDE-Slover-3 (ours) on ImageNet $128\times128$ (NFE=20). The pretrained model is Guided-diffusion \cite{dhariwal2021diffusion} (without classifier guidance).}
\label{our_img128}
\end{figure*}

\begin{figure*}[htbp]
\centering
\includegraphics[width=0.9\textwidth]{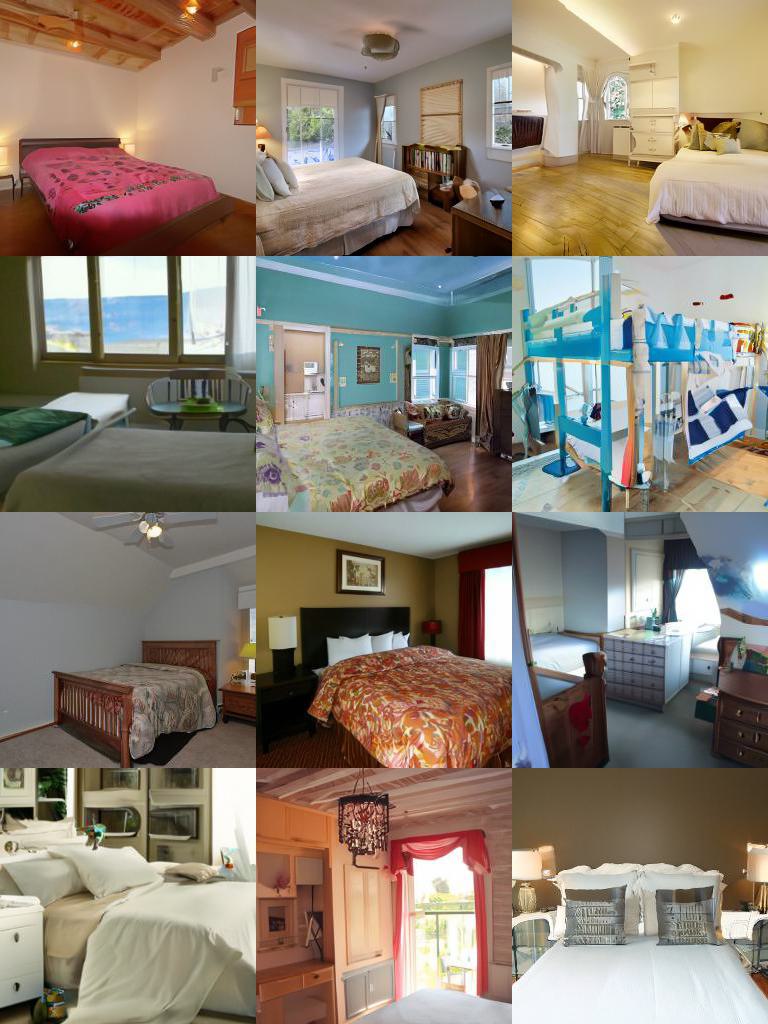}
\caption{Generated images with ER-SDE-Slover-3 (ours) on LSUN Bedrooms $256\times256$ (NFE=20). The pretrained model is Guided-diffusion \cite{dhariwal2021diffusion}.}
\label{our_lusn}
\end{figure*}

\begin{figure*}[htbp]
\centering
\includegraphics[width=0.9\textwidth]{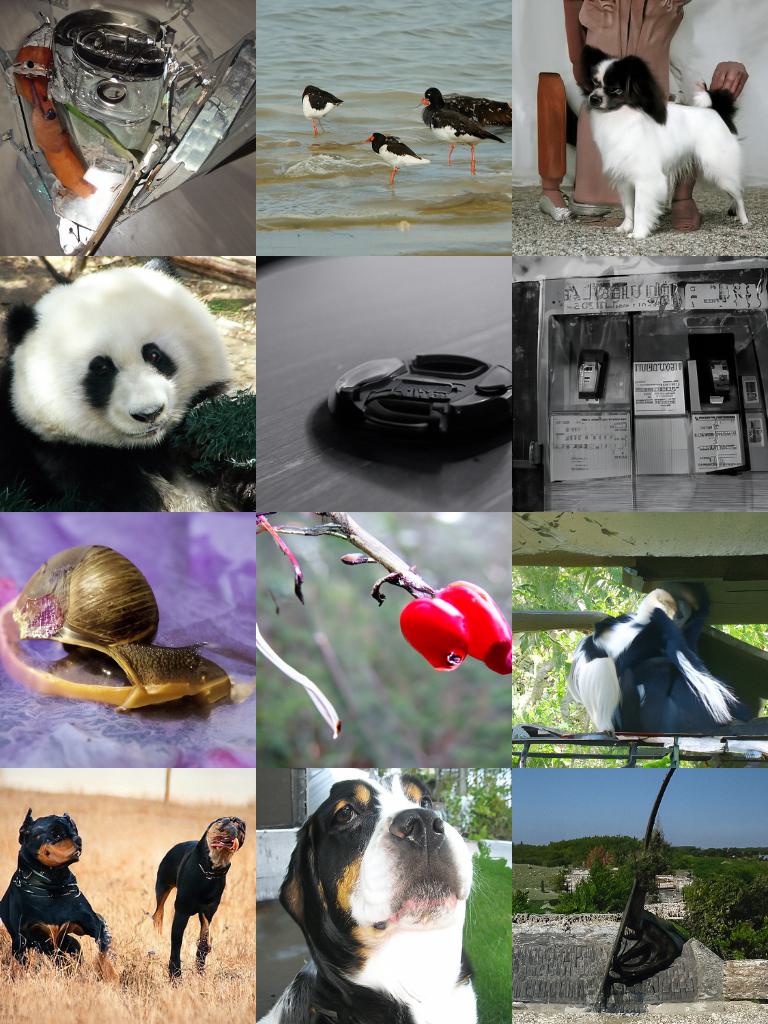}
\caption{Generated images with ER-SDE-Slover-3 (ours) on ImageNet $256\times256$ (NFE=20). The pretrained model is Guided-diffusion \cite{dhariwal2021diffusion} (classifier guidance scale=2.0).}
\label{our_img256}
\end{figure*}

Finally, we also provide samples generated by ER-SDE-Solvers using different pretrained models on various datasets, as illustrated in Fig.\ref{our_cifar} - Fig.\ref{our_img256}.
\section{Additional Discussion}
\label{appendixe}
\textbf{Limitations.} Despite the promising acceleration capabilities, ER-SDE-Solvers are designed for efficient high-quality sampling, which may not be suitable for accelerating the likelihood evaluation of DMs. Furthermore, compared to commonly used GANs \cite{goodfellow2014generative}, flow-based generative models \cite{kingma2018glow}, and techniques like distillation for speeding up sampling \cite{salimans2022progressive,song2023consistency}, DMs with ER-SDE-Solvers are still not fast enough for real-time applications.

\textbf{Future work.} This paper introduces a unified framework for DMs, and there are several aspects that merit further investigation in future work. For instance, this paper maintains consistency between the time step mentioned in Proposition 4.3, 4.5 and the pretrained model (see Sec.\ref{stepsize}). In fact, many works \cite{lu2022dpm,jolicoeur2021gotta} have carefully designed time steps tailored to their solvers and achieved good performance. Although our experimental results demonstrate that ER-SDE-Solvers can achieve outstanding performance even without any tricks, further exploration of time step adjustments may potentially enhance the performance. Additionally, this paper only explores some of the noise scale functions for the reverse process, providing examples of excellent performance (such as ER SDE 5). Whether an optimal choice exists for the noise scale function is worth further investigation. Lastly, applying ER-SDE-Solvers to other data modalities, such as speech data \cite{kong2020diffwave}, would be an interesting avenue for future research.

\end{document}